\pdfoutput=1
\documentclass{article}
\usepackage{style/iclr2027_conference,times}

\usepackage{amsmath,amsfonts,bm}

\def\eqref#1{equation~\ref{#1}}

\def\1{\bm{1}}

\DeclareMathAlphabet{\mathsfit}{\encodingdefault}{\sfdefault}{m}{sl}
\SetMathAlphabet{\mathsfit}{bold}{\encodingdefault}{\sfdefault}{bx}{n}

\usepackage{hyperref}
\usepackage{url}
\usepackage{xurl}
\usepackage{graphicx}
\usepackage{booktabs}
\usepackage{array}

\usepackage[boxed]{amazontitle}
\amazontitlesetup{
  author-layout={2,1,0},
  title-break-after={:},
  logo={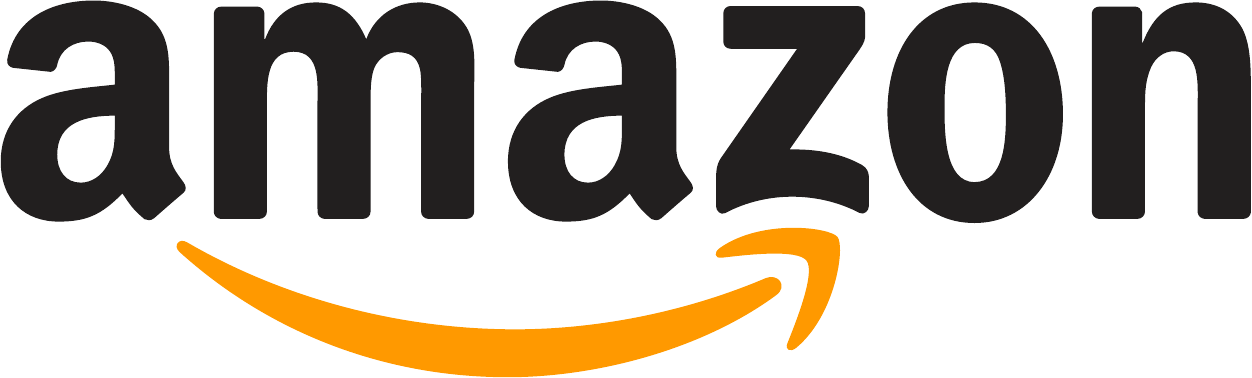},
  footer-label={Project page},
  correspondence={\url{https://guanghanwang.com/dlm-scaling}}
}

\usepackage{multirow}
\usepackage{amsmath,amssymb}
\usepackage{xcolor}
\usepackage{caption}
\definecolor{bdcite}{rgb}{0.368,0.507,0.710}
\definecolor{bdlink}{rgb}{0.923,0.386,0.209}
\definecolor{bdurl}{rgb}{0.851,0.322,0.145}
\hypersetup{colorlinks=true,citecolor=bdcite,linkcolor=bdlink,urlcolor=bdurl}

\usepackage{subcaption}
\usepackage{wrapfig}
\usepackage{needspace}

\newcommand{\Lstar}{\mathcal{L}^{\ast}}

\newcommand{\mn}{m\text{-}nats}

\title{\raggedright\hyphenpenalty=10000\exhyphenpenalty=10000 Revisiting the 16x Pretraining Compute Gap between Diffusion and Autoregressive Language Models}

\author{%
Guanghan Wang$^{1,\dagger,\ddagger}$\quad
Bing He$^{2}$\quad
Zhan Shi$^{2}$\quad
Benoit Dumoulin$^{2}$\\
\textbf{Hanqing Lu$^{2,\ast}$\quad
Volodymyr Kuleshov$^{1,\ast}$}\\
{\normalfont $^{1}$Cornell University, Cornell Tech\qquad $^{2}$Amazon}
}

\iclrfinalcopy

\begin{document}
\raggedbottom
\renewcommand{\topfraction}{0.9}\renewcommand{\bottomfraction}{0.9}\renewcommand{\textfraction}{0.07}\renewcommand{\floatpagefraction}{0.8}
\makeatletter\setlength{\@fptop}{0pt}\setlength{\@fpsep}{12pt plus 2pt}\setlength{\@fpbot}{0pt plus 1fil}\makeatother
\maketitle
\lhead{Preprint}
{\renewcommand{\thefootnote}{}\footnotetext{$^{\dagger}$Work done during internship at Amazon.}\footnotetext{$^{\ast}$Equal supervision.}\footnotetext{$^{\ddagger}$Correspondence to: \texttt{guanghan@cs.cornell.edu}.}}

\vspace{-10pt}
\begin{abstract}
Diffusion language models (DLMs) have attracted wide attention for their fast, parallel decoding, but they are believed to be far less efficient to train than autoregressive language models (ARMs): a widely cited study reports that a DLM needs roughly $16\times$ the training compute of an ARM to match its likelihood, a figure that has made pretraining DLMs from scratch look prohibitive at scale. We find that the methodology behind this figure has fallen out of step with current DLM practice: it scores the model with its evidence lower bound (ELBO) and trains a full-sequence, decoder-only model. We therefore revisit the scaling laws of DLMs, adopting the exact likelihood under the model's own decoding order, block diffusion training, an encoder--decoder architecture, and a sweep over the block sizes used in practice. Under this new protocol, we study across ten model variants, about $1{,}700$ models in total, and find that pretraining DLMs is far cheaper than previously thought. Concretely, scoring with the exact likelihood reduces the compute multiple to $12.8\times$; block diffusion training and an encoder--decoder architecture further reduce it to $7.8\times$ at block size $32$, and smaller block sizes narrow it to $4.6\times$ at block size $8$ and $3.0\times$ at block size $4$. We complement these scaling analyses with an empirical rule for the optimal decoder depth of encoder--decoder block diffusion, a held-out compute budget beyond the fitting range at which the gap between variants persists, and downstream benchmark results where likelihood improvements carry over to accuracy. Overall, the $16\times$ figure substantially overstates the pretraining cost gap between DLMs and ARMs, and pretraining DLMs from scratch is far more viable than previously assumed.
\end{abstract}



\noindent\begin{minipage}{\linewidth}
\centering
\includegraphics[width=0.85\linewidth]{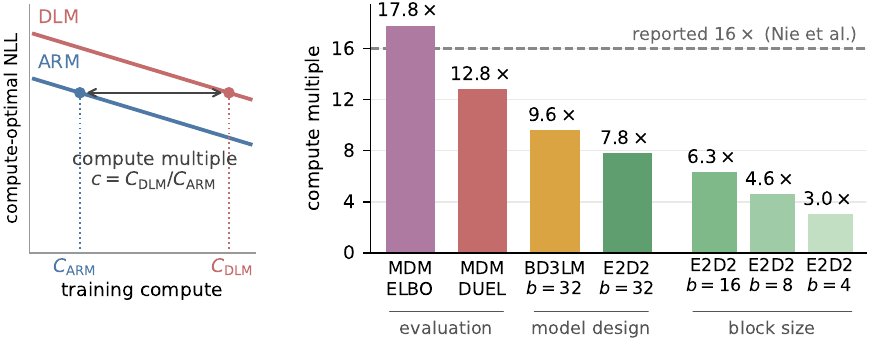}
\vspace{-4pt}
\captionof{figure}{Revisiting the $16\times$ compute multiple between diffusion and autoregressive language models. Left: illustration of the compute multiple $c=C_{\rm DLM}/C_{\rm ARM}$ on the compute-optimal negative log-likelihood (NLL) scaling law plot. Right: $c$ at $10^{20}$ FLOPs as the evaluation (ELBO $\to$ DUEL), the model design (MDM $\to$ BD3LM $\to$ E2D2) and the block size are revised.}
\label{fig:staircase}
\end{minipage}
\vspace{2pt}

\section{Introduction}
\label{sec:intro}

Masked diffusion models (MDMs) have become the leading non-autoregressive alternative for language modeling \citep{austin2021structured,lou2023discrete,sahoo2024simple,shi2024simplified,ou2025your,nie2026large}. Their appeal is fast, parallel generation: open-weight masked diffusion language models report about $2.5\times$ the generation throughput of autoregressive language models (ARMs) at comparable accuracy \citep{wu2026fastv2,reda2026nemotron}, with commercial systems reporting up to $10\times$ \citep{labs2025mercury}. Yet, their training efficiency is widely believed to be far worse than that of ARMs. The most cited quantification is the IsoFLOP study of \citet{nie2025scaling} (SMDM), which concludes that an MDM needs about $16\times$ the training FLOPs of an ARM to reach the same validation perplexity, and the number is now routinely quoted as a property of the MDM family. This $16\times$ figure has made pretraining MDMs from scratch look prohibitive at scale.

Taking a closer look at how this $16\times$ figure was achieved, however, we find that the methodology behind it has fallen out of step with current practice of diffusion language modeling. SMDM measured it on a full-sequence, decoder-only MDM scored with its negative evidence lower bound (ELBO), all three of which now have alternatives. First, the ELBO is an upper bound on the negative log-likelihood (NLL), and it scores the model under uniformly random token orders it never uses at inference; in contrast, DUEL \citep{turok2026duel} computes the exact NLL of a diffusion model under the custom decoding order it is deployed with. Second, nearly all of today's large-scale diffusion language models (DLMs) use block-wise factorization \citep[BD3LM;][]{arriola2025block} with block sizes between $4$ and $32$, rather than operating over the full sequence. Third, encoder--decoder architectures that separate the processing of clean and noised tokens \citep[E2D2;][]{arriola2026encoder} offer computationally cheaper training than the decoder-only network. While the papers that introduce these techniques compare a handful of models at fixed sizes, none of these has been measured under a rigorous compute-matched IsoFLOP analysis.

We therefore revisit the scaling laws of masked diffusion models with a protocol that implements the ideas above: we score every diffusion model using the exact likelihood induced by its top-1 confidence decoding rule, utilize block-aware training and an encoder--decoder architecture, and sweep the block sizes $b\in\{4,8,16,32\}$ which are commonly used in practice. In addition, we make two corrections to the SMDM protocol: we count training FLOPs exactly, including the output head and attention computation, and we read compute multiples within the measured range of the scaling laws rather than by extrapolation. Under this new protocol we study ten model variants and train about $1{,}700$ models in total; to our knowledge, this is the largest compute-matched scaling analysis of diffusion language models to date.

With this analysis, we find that pretraining diffusion language models is far cheaper than previously thought (Figure~\ref{fig:staircase}). Scored with the ELBO, our full-sequence MDM lands close to the published figure, at $17.8\times$. Scoring the very same checkpoints with the exact likelihood lowers the multiple to $12.8\times$. Block diffusion training and the encoder--decoder architecture then reduce it further to $7.8\times$ at block size $32$, and smaller block sizes narrow it to $4.6\times$ at block size $8$ and $3.0\times$ at block size $4$. In addition, we find that the optimal decoder depth of encoder--decoder block diffusion follows a simple rule that depends on the block size alone, that the gap between E2D2 and the ARM holds at a held-out budget three times larger than our largest fitted FLOPs, and that likelihood improvements carry over to downstream accuracy.

Our contributions are the following. (i) We revisit the scaling laws of diffusion language models under a revised protocol that reflects recent advances: exact likelihood evaluation, block diffusion, encoder--decoder models, and a sweep over practically used block sizes. (ii) We show that, under this protocol, pretraining diffusion language models is far cheaper than the $16\times$ suggests: exact likelihood evaluation reduces the compute multiple to $12.8\times$, block diffusion training with an encoder--decoder architecture reduces it to $7.8\times$ at $b{=}32$, and smaller block sizes further narrow it to $4.6\times$ at $b{=}8$ and $3.0\times$ at $b{=}4$. (iii) We complement these scaling analyses with an empirical rule for the optimal decoder depth of encoder--decoder block diffusion, which depends on the block size alone; a held-out budget at $3{\times}10^{20}$ FLOPs, at which the gap between E2D2 and the ARM persists; and downstream benchmark results where likelihood improvements carry over to downstream accuracy.

\section{Background}
\label{sec:background}

\paragraph{Masked diffusion models.}
Let $\mathbf{x}^{1:n}=(x^1,\dots,x^n)$ be a sequence of $n$ tokens from a vocabulary $\mathcal V$ augmented with a mask token $\mathtt{m}$. The forward process of a masked diffusion model \citep{austin2021structured,sahoo2024simple,shi2024simplified,ou2025your} is indexed by $t\in[0,1]$ and masks each token independently with probability $1-\alpha_t$, where the schedule $\alpha_t$ decreases from $\alpha_0=1$ to $\alpha_1=0$; we write $\mathbf{z}_t^{1:n}$ for the partially masked sequence. A network $p_\theta(\cdot\mid \mathbf{z}_t^{1:n})$ predicts every masked position given the partially masked token sequence. Under the linear schedule $\alpha_t=1-t$, an MDM is trained with a negative ELBO,
\begin{equation}
\label{eq:elbo}
\mathcal L_{\rm ELBO}(\mathbf{x}^{1:n})\;=\;\mathbb E_{t\sim\mathcal U(0,1)}\,\mathbb E_{\mathbf{z}_t^{1:n}\sim q_t(\cdot\mid \mathbf{x}^{1:n})}\Big[\tfrac{1}{t}\sum_{\ell:\,z_t^\ell=\mathtt m}-\log p_\theta(x^\ell\mid \mathbf{z}_t^{1:n})\Big]\;\ge\;-\log p_\theta(\mathbf{x}^{1:n}).
\end{equation}
Following \citet{sahoo2026scaling}, we train all diffusion models on the low-variance version of this loss that drops the weight $1/t$.

Prior scaling studies \citep{nie2025scaling,von2026scaling,ni2025training} evaluate a trained MDM by a Monte-Carlo estimate of Eq.~\ref{eq:elbo} and compare it with the exact NLL of an ARM. This comparison is problematic in two respects. First, Eq.~\ref{eq:elbo} is only an upper bound on the NLL; how far it lies from the NLL, and how this varies across models, is poorly understood. Second, Eq.~\ref{eq:elbo} equals the expected NLL of an autoregressive factorization under a uniformly random token order \citep{hoogeboom2021autoregressive,shi2024simplified,ou2025your,kim2025train}, whereas an MDM is never decoded in a random order in practice: its sampler chooses the order adaptively, most commonly by the model's own confidence \citep{chang2022maskgit,nie2026large}. Scoring a model under orders it never uses thus evaluates a distribution that differs from the one induced by its decoding rule at deployment.

\paragraph{Exact likelihood under a deterministic unmasking rule (DUEL).}
\citet{turok2026duel} address both points. Let $\pi$ be a deterministic unmasking rule that, given the tokens unmasked so far, selects the next position to reveal. Writing $\sigma_i$ for the position revealed at step $i$, scoring $\mathbf{x}^{1:n}$ under $\pi$ amounts to teacher-forcing the true tokens in the order $\sigma$, and $p^{\pi}_\theta(\mathbf{x}^{1:n})=\prod_{i=1}^{n} p_\theta(x^{\sigma_i}\mid x^{\sigma_{<i}})$ is a properly normalized distribution, since $\sigma_i$ depends only on $x^{\sigma_{<i}}$. Throughout the paper, we fix $\pi$ to the top-1 confidence rule \citep{nie2026large}, the default sampler of diffusion language models, and report $-\log p^{\pi}_\theta(\mathbf{x}^{1:n})$ (DUEL) in nats per token like the ARM's NLL.

\paragraph{Block diffusion.}
Full-sequence MDMs cannot generate sequences of arbitrary length and cannot reuse a KV cache. Block diffusion (BD3LM; \citealp{arriola2025block}) partitions $\mathbf{x}^{1:n}$ into contiguous blocks of $b$ tokens, factorizes the sequence autoregressively over blocks, and models each block with a masked diffusion model conditioned on the clean prefix. Nearly all large-scale diffusion language models utilize block decomposition now. BD3LM keeps the decoder-only architecture: to score all blocks in one forward pass it concatenates the noised and clean sequences, so a training step runs over $2n$ tokens and costs roughly twice the FLOPs of an ARM or MDM step at the same size.

\paragraph{Encoder--decoder block diffusion.}
E2D2 \citep{arriola2026encoder} avoids this overhead by splitting the $L$-layer transformer into an $L_{\rm enc}$-layer encoder that represents the clean prefix and an $L_{\rm dec}=L-L_{\rm enc}$-layer decoder that denoises the noisy block, attending to its own block by self-attention and to the encoder outputs of the prefix by cross-attention. Each training step is therefore cheaper than a BD3LM step because clean and noised tokens do not go through the same big network.

\paragraph{IsoFLOP scaling laws and the compute multiple.}
Prior scaling studies of diffusion language models \citep{nie2025scaling,von2026scaling,sahoo2026scaling,yang2026continuous} adopt one IsoFLOP protocol from \citet{hoffmann2022training}: for a fixed training budget $C$, a ladder of model sizes $N$ is trained on $D=C/c(N)$ tokens each, where $c(N)$ is the training cost per token. Then, a parabola fitted to the loss against $\log N$ gives the compute-optimal loss $\Lstar(C)$ at that budget. Across budgets they fit the power law $\Lstar(C)=a\,C^{\alpha}$ by least squares in log--log space. The compute gap between two model families is measured at fixed loss: if family $X$ reaches loss $\mathcal{L}$ at budget $C_X(\mathcal{L})$ and family $Y$ at $C_Y(\mathcal{L})$, the compute multiple is $C_X(\mathcal{L})/C_Y(\mathcal{L})$. It is in this sense that \citet{nie2025scaling} report a multiple of $16$ between MDM and ARM.

\section{Scaling law protocol}
\label{sec:protocol}

In this section, we introduce our revised scaling law protocol for masked diffusion language models. It inherits the protocol of SMDM \citep{nie2025scaling} with two kinds of changes. First, we replace three of SMDM's choices with more recent alternatives: exact likelihood evaluation, block diffusion at commonly used block sizes, and an encoder--decoder architecture. Second, we correct two aspects of the original SMDM protocol, namely the FLOP count that sets the length of each run and the readout of the compute multiple from the fitted laws. Section~\ref{sec:updates} describes the first kind of changes, and Sections~\ref{sec:recipe} and~\ref{sec:fitting} state the resulting training and fitting procedure with both corrections included.

\subsection{Incorporating recent advances}
\label{sec:updates}

\paragraph{Exact likelihood evaluation.}
\label{sec:eval}
We score every diffusion model with DUEL under the top-1 confidence unmasking rule, yielding the exact likelihood induced by this decoding order. To keep the comparison with SMDM, we also score every diffusion model with the ELBO (Eq.~\ref{eq:elbo}, estimated with $128$-particle Monte-Carlo), and the ARM with its exact NLL. All models are scored on one frozen validation set of $49{,}152$ sequences, and unless stated otherwise the diffusion models are reported with DUEL.

\paragraph{Block diffusion.}
\label{sec:variants}
In our scaling law analysis protocol, we train block diffusion models at block sizes $b\in\{4,8,16,32\}$. These configurations are chosen from the perspective of language model serving. By the roofline model \citep{williams2009roofline}, a GPU becomes compute-bound only above the ratio of its peak throughput to its memory bandwidth; taking BF16 serving as an example, this ratio is about $300$ FLOPs per byte for an H100 GPU \citep{nvidia2023h100}. A block diffusion model decodes up to $b$ tokens per sequence in one forward pass, which raises the arithmetic intensity to about $Bb$, where $B$ is the batch size. For mid-size serving with $B$ between $8$ and $32$, block sizes from $32$ down to $8$ thus bring $Bb$ to about $256$ and put the compute units left idle by autoregressive decoding to use, while a larger block adds little throughput. We therefore take $b\in\{8,16,32\}$ as the main block sizes, and use $b{=}4$ as a reference for how much training compute a still smaller block can save. These block sizes also match those used to train large diffusion language models: Nemotron-Labs-Diffusion \citep{fu2026nemotron} is trained with $b{=}32$, Nemotron-TwoTower \citep{reda2026nemotron} with $b{=}16$, and SDAR \citep{cheng2026sdar} and FLARE \citep{zhu2026flare} with $b{=}4$ (see Table~\ref{tab:blocksizes} in Appendix~\ref{app:blocksizes} for more models).

\paragraph{Encoder--decoder architecture.}
Besides decoder-only block diffusion models (BD3LM), we also train encoder--decoder block diffusion models (E2D2) at the same four block sizes in our scaling law analysis protocol. Overall, this gives us ten model variants: ARM, MDM, BD3LM with $b\in\{4,8,16,32\}$ and E2D2 with $b\in\{4,8,16,32\}$.

\paragraph{Optimal decoder depth of E2D2.}
\label{sec:sweep}
E2D2 has one degree of freedom that the other variants do not have: how the $L$ layers of a given ladder size are split between the encoder and the decoder, i.e., the decoder fraction $L_{\rm dec}/L$. A wrong split would penalize E2D2 for a reason unrelated to the architecture design itself, so at every grid point of E2D2, i.e., every combination of block size $b$, training budget $C$ and model size $N$, we hold the total depth $L$ fixed and sweep $L_{\rm dec}$ over a wide range. We then report the best DUEL value from the sweep. As a side product, we also analyze how the optimal decoder depth $L^\ast_{\rm dec}$ is related to $b$, $C$ and $N$. We believe our results (Section~\ref{sec:ndec}) can give practical guidance on the pretraining of encoder--decoder block diffusion models.

\subsection{Training recipe and IsoFLOP sweep}
\label{sec:recipe}

\paragraph{Training recipe.}
To keep our results consistent with the $16\times$ scaling analysis of SMDM, we largely inherit its choices of network architecture, data, tokenizer and hyperparameters. Concretely, all models are LLaMA-style transformers \citep{touvron2023llama} trained on SlimPajama \citep{cerebras2023slimpajama} with a sequence length of $2048$ and a batch of $256\times2048$ tokens. Following SMDM, we use standard parameterization (SP, as opposed to $\mu$P; \citealp{yang2022tensor}) and AdamW ($\beta=(0.9,0.95)$, weight decay $0.1$, gradient clipping at $1.0$) with a peak learning rate of $4{\times}10^{-4}$, a linear warm-up over $1\%$ of the steps (at least $100$) and cosine decay to $1/10$ of the peak. All diffusion variants use the linear noise schedule $\alpha_t=1-t$ and, following \citet{sahoo2026scaling}, the low-variance training loss; following \citet{arriola2026encoder}, BD3LM and E2D2 draw an independent noise level for every block and always mask at least one token in each block.

\paragraph{IsoFLOP grid.}
We reuse the size ladder of SMDM (19M--2.1B non-embedding parameters) for all variants and, for each variant, train $7$--$10$ sizes around the compute-optimal size on each of five budgets $\{6{\times}10^{18},10^{19},3{\times}10^{19},6{\times}10^{19},10^{20}\}$ FLOPs. Across the ten variants, five budgets, sizes per budget and the decoder-depth sweep of E2D2, the grids comprise about $1{,}700$ trained models (approximately $136$k B200 GPU hours), more than any prior compute-matched study of diffusion language model scaling that we are aware of.

\paragraph{Compute accounting.}
\label{sec:flops}
At a FLOP budget $C$, the length of each run is set by the variant's training FLOPs per token. SMDM takes this to be $6N$ \citep{kaplan2020scaling}, which is too coarse for a comparison across variants. The $6N$ rule ignores the attention score computation entirely, although at our sequence length of $2048$ it accounts for a non-negligible fraction of the training FLOPs. More importantly, this fraction differs across variants. We therefore follow \citet{hoffmann2022training} and count the actual training FLOPs, including the output head and the score computation of every attention sublayer (Appendix~\ref{app:flops} gives the per-variant formulas).

\subsection{Fitting scaling laws}
\label{sec:fitting}

\paragraph{Fitting.}
For each variant we have five budgets $C$, each contributing $7$--$10$ sizes $N$ with their measured NLL (DUEL for the diffusion variants). For each $(\text{variant},C)$ we fit a centered parabola in $\log_{10}N$ to these values and take its vertex, which gives the compute-optimal NLL $\Lstar(C)$; for E2D2, each size contributes only its best $L_{\rm dec}$. Across budgets we fit the power law $\Lstar(C)=a\,C^{\alpha}$ in log--log space, keeping the pure power-law form of prior work so that our numbers are comparable (see further discussion on the law's functional form in Appendix~\ref{app:eform}).

\paragraph{Reading the compute multiple.}
For each diffusion variant we take its compute-optimal NLL $\Lstar$ at $10^{20}$ FLOPs, find the compute at which the ARM's fitted law reaches that same $\Lstar$, and take the ratio. Unlike SMDM, we do not extrapolate to do so: given the compute gap between the ARM and the diffusion variants, extrapolating a diffusion law to reach the ARM's loss lands several times beyond the largest budget we ran, where a small level of noise in the fitted exponent would swing the multiple substantially. Reading the multiple inside the ARM's measured range instead introduces little noise (see Appendix~\ref{app:isoflop} for a comparison between the two readouts).

\section{Results}
\label{sec:results}

\subsection{The $16\times$ overstates the cost of pretraining DLMs}
\label{sec:staircase}

\begin{figure}[t]
\centering
\includegraphics[width=\linewidth]{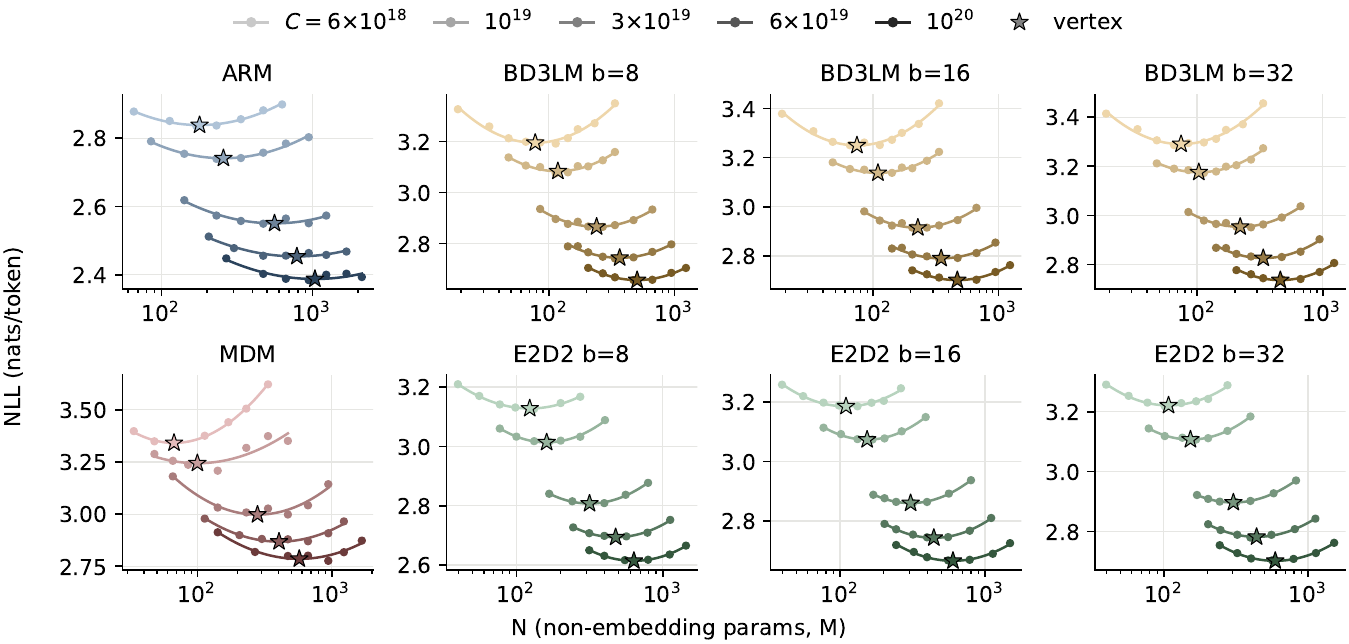}
\caption{IsoFLOP curves for eight main model variants (DUEL for diffusion variants, exact NLL for the ARM; for E2D2, each point is the best $L_{\rm dec}$ at that size). Darker shades mark larger budgets, and stars mark fitted vertices.}
\label{fig:isoflop}
\end{figure}

In Figure~\ref{fig:isoflop} we show the IsoFLOP curves of eight main model variants (those of the two $b{=}4$ variants are in Appendix~\ref{app:isoflopcurves}), and in Figure~\ref{fig:laws} the fitted scaling laws, whose parameters we list in Table~\ref{tab:mainlaws}. These laws fit the data well, with a maximum relative residual of only $0.5\%$. Next, we analyze one by one how each technique introduced in our revised protocol affects the scaling and the compute multiple of masked diffusion language models.

\begin{figure}[t]
\begin{minipage}[t]{0.45\linewidth}
\centering
\footnotesize
\captionof{table}{Fitted laws $\Lstar(C)=a\,C^{\alpha}$ (DUEL for diffusion unless stated) and the compute multiple against ARM.}
\label{tab:mainlaws}
\vspace{1pt}
\renewcommand{\arraystretch}{0.95}\setlength{\tabcolsep}{5pt}
\begin{tabular}{lrrr}
\toprule
variant & $a$ & $\alpha$ & multiple \\
\midrule
ARM & 40.95 & $-0.0618$ & --- \\
MDM (ELBO) & 49.93 & $-0.0623$ & 17.8 \\
MDM & 57.87 & $-0.0659$ & 12.8 \\
BD3LM $b{=}32$ & 55.54 & $-0.0654$ & 9.6 \\
E2D2 $b{=}32$ & 47.79 & $-0.0624$ & 7.8 \\
\midrule
BD3LM $b{=}16$ & 55.56 & $-0.0657$ & 7.8 \\
E2D2 $b{=}16$ & 49.26 & $-0.0634$ & 6.3 \\
BD3LM $b{=}8$ & 54.34 & $-0.0656$ & 5.9 \\
E2D2 $b{=}8$ & 48.52 & $-0.0635$ & 4.6 \\
BD3LM $b{=}4$ & 52.99 & $-0.0655$ & 4.1 \\
E2D2 $b{=}4$ & 45.50 & $-0.0627$ & 3.0 \\
\bottomrule
\end{tabular}
\end{minipage}\hfill
\begin{minipage}[t]{0.52\linewidth}\vspace{-20pt}
\centering
\includegraphics[width=0.9\linewidth]{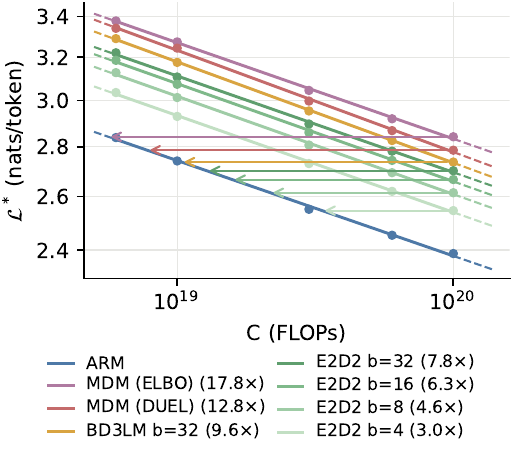}
\vspace{-9pt}
\captionof{figure}{Compute-optimal NLL of different model variants, solid inside the fitted range and dashed outside.}
\label{fig:laws}
\end{minipage}

\vspace{4pt}
\centering
\footnotesize
\captionof{table}{Compute multiples at $10^{20}$ FLOPs of every diffusion variant, measured with the negative ELBO and with DUEL. DUEL yields a smaller multiple than the negative ELBO for every variant.}
\label{tab:readout}
\setlength{\tabcolsep}{5.5pt}\renewcommand{\arraystretch}{1.05}
\resizebox{\linewidth}{!}{\begin{tabular}{l ccccc cccc}
\toprule
 & MDM & BD3LM\,32 & BD3LM\,16 & BD3LM\,8 & BD3LM\,4 & E2D2\,32 & E2D2\,16 & E2D2\,8 & E2D2\,4 \\
\midrule
DUEL & 12.8 & 9.6 & 7.8 & 5.9 & 4.1 & 7.8 & 6.3 & 4.6 & 3.0 \\
ELBO & 17.8 & 16.5 & 13.6 & 10.1 & 6.3 & 12.0 & 9.8 & 7.1 & 4.2 \\
\bottomrule
\end{tabular}}
\end{figure}

\paragraph{An ELBO baseline in line with $16\times$.}
With the ELBO readout, our MDM needs $17.8\times$ the ARM's compute at $10^{20}$ FLOPs, comparable to the $16\times$ of \citet{nie2025scaling}. Although our protocol differs from theirs (in the low-variance training loss, real FLOPs in place of $6ND$, and the readout), our compute multiple still agrees roughly with their $16\times$. On the same corpus and tokenizer, \citet{sahoo2026scaling} and \citet{yang2026continuous} report a smaller $\approx14\times$ for the low-variance MDM, but with a DiT backbone \citep{peebles2023scalable} in place of the Llama-style transformer of SMDM that we adopt. This baseline anchors the decomposition that follows.

\paragraph{Exact likelihood readout: the evaluation penalty.}
Scoring the very same MDM checkpoints with DUEL instead of the ELBO lowers the multiple to $12.8\times$. Nothing about the model or its training has changed; about a third of the reported gap is thus an artifact of the evaluation. A reduction is expected: the negative ELBO is an upper bound on the marginal NLL of the MDM whose tightness is poorly understood, whereas DUEL is the exact NLL of the distribution induced by the model’s top-1 confidence decoding rule, and is therefore directly comparable with the ARM's NLL. This is further supported by extra empirical evidence: in Table~\ref{tab:readout}, where every diffusion variant is scored both ways, DUEL gives a smaller multiple than the negative ELBO in every case. Beyond likelihood, we also find that DUEL scores higher than the ELBO on multiple-choice benchmarks (Section~\ref{sec:bench}), further evidence in favor of DUEL than the ELBO.

\paragraph{Block factorization: the first modeling change.}
The remaining $12.8\times$ is a genuine training-compute gap, and the next two replacements reduce it by changing how the model is built. Moving from the full-sequence MDM to BD3LM lowers the multiple to $9.6\times$ at $b{=}32$, $7.8\times$ at $b{=}16$ and $5.9\times$ at $b{=}8$. Although block-wise training has become common practice \citep{bie2025llada2,wu2026fastv2,fu2026nemotron,reda2026nemotron,cheng2026sdar,zhu2026flare}, this is, to our knowledge, the first controlled, compute-matched demonstration of the benefit of this method in pretraining diffusion language models. We take this as evidence that introducing some degree of left-to-right bias helps language model learning, and that the likelihood gain from the left-to-right block factorization outweighs the penalty of BD3LM's more expensive training steps, which at matched compute roughly halve the number of steps relative to the MDM.

\begin{figure}[t]
\centering
\includegraphics[width=0.97\linewidth]{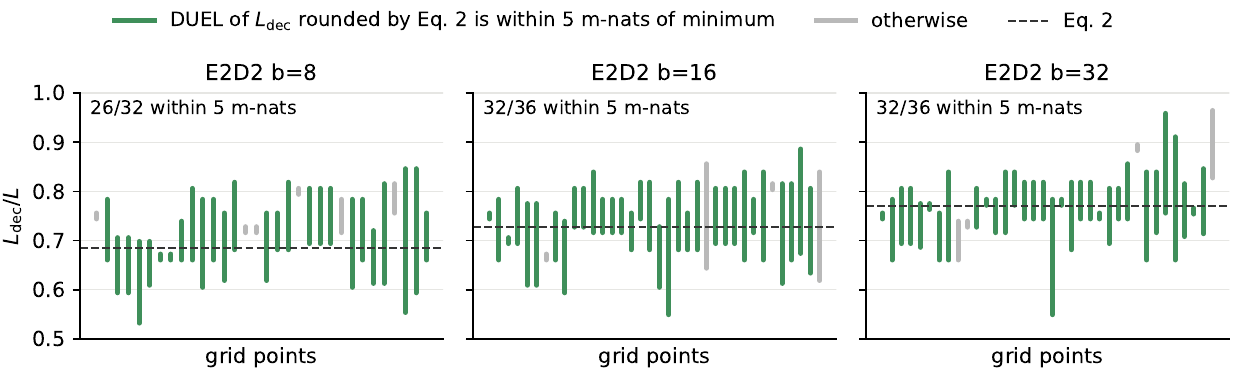}
\caption{The decoder depth given by Eq.~\ref{eq:ndec} is near-optimal for E2D2 at $b{=}8$, $16$ and $32$. Each bar represents one grid point $(C,N)$ and spans the decoder fractions $L_{\rm dec}/L$ whose DUEL lies within $5$ \mn{} of the minimum of that grid point's sweep. A bar is green if the DUEL of the decoder depth given by Eq.~\ref{eq:ndec}, i.e., its prediction multiplied by $L$ and rounded to an integer, is within $5$ \mn{} of the minimum.}
\label{fig:ndec}
\end{figure}

\paragraph{Encoder--decoder architecture: the second modeling change.}
Moving from BD3LM to E2D2 at the same block size lowers the multiple once more, from $9.6\times$ to $7.8\times$ at $b{=}32$, from $7.8\times$ to $6.3\times$ at $b{=}16$ and from $5.9\times$ to $4.6\times$ at $b{=}8$; at $b{=}4$ it reaches $3.0\times$. Note that the decoder-only architecture, inherited from ARMs, remains the convention for diffusion language models: nearly all large-scale DLMs to date, including those trained with block diffusion, use a single decoder-only network \citep{nie2026large,ye2025dream,wu2026fastv2,fu2026nemotron}. Our experiments provide strong evidence that this convention is not the most training-efficient choice for masked diffusion: at matched compute, the encoder--decoder architecture is consistently better, which suggests that it is better suited to the masked diffusion training task. This shows that, for diffusion language models, a conceptual split between encoding and decoding is beneficial: although the encoder and the decoder each have less capacity than the original unified model, letting each concentrate on a different job allows them to use the pretraining compute more efficiently.

\subsection{Where the layers go: optimal decoder depth in E2D2}
\label{sec:ndec}

The decoder-depth sweeps of Section~\ref{sec:sweep} give the optimal decoder fraction $L^\ast_{\rm dec}/L$ at every one of the $142$ $(b,C,N)$ grid points of E2D2. They lead to a simple conclusion: \textbf{the optimal decoder fraction depends on the block size alone, and it is decoder-heavy.} A regression of $L^\ast_{\rm dec}/L$ on $\log_{10}N$, $\log_{10}C$ and $\log_2 b$ finds $\log_2 b$ to be the dominant predictor, with the dependence on $N$ and $C$ at most weak (Appendix~\ref{app:ndec}); regressing on $\log_2 b$ alone gives the empirical rule
\begin{equation}
\label{eq:ndec}
L^\ast_{\rm dec}/L \;\approx\; 0.556 + 0.043\,\log_2 b.
\end{equation}

The rule is convenient because the DUEL basin around $L^\ast_{\rm dec}$ is flat, so a slightly wrong choice costs little. To apply the rule, one multiplies it by the total depth $L$ and rounds to a whole number of decoder layers. Figure~\ref{fig:ndec} shows that this choice is nearly optimal: at $90$ of the $104$ grid points with $b\in\{8,16,32\}$, its DUEL lies within $5$ \mn{} of the best split found by the sweep, and within $10$ \mn{} at $99$ of them, including every compute-optimal one (see Appendix~\ref{app:ndec} for $b{=}4$).

\begin{figure}[t]
\centering
\begin{minipage}[t]{0.48\linewidth}\vspace{0pt}
\centering
\includegraphics[width=\linewidth]{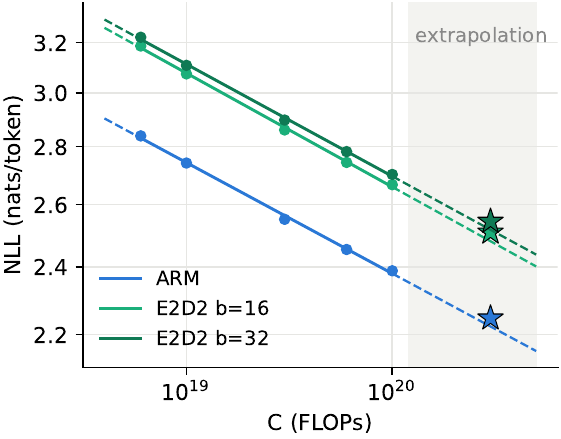}
\vspace{-14pt}
\caption{The held-out budget at $3{\times}10^{20}$ FLOPs. Lines are the power laws fitted on the five budgets $\le10^{20}$ (dots), dashed outside the fitted range; stars are the measured values at the held-out budget.}
\label{fig:extrap}
\end{minipage}\hfill
\begin{minipage}[t]{0.48\linewidth}\vspace{0pt}
\centering
\includegraphics[width=\linewidth]{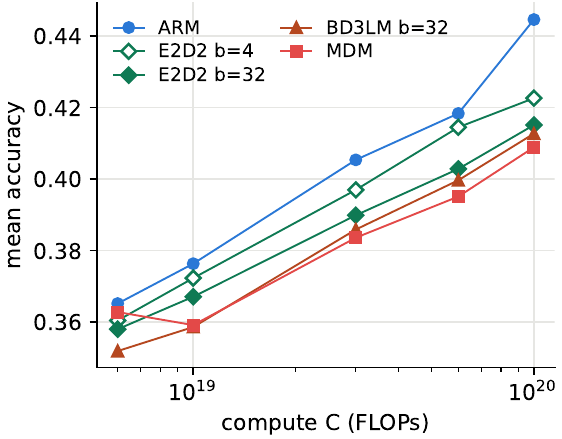}
\vspace{-14pt}
\caption{Mean accuracy over the eight retained benchmarks against compute, for the compute-optimal cell of the ARM and of four diffusion variants. The order in benchmark accuracy broadly matches the order in NLL.}
\label{fig:bench}
\end{minipage}
\end{figure}

\subsection{The gap between variants persists beyond the fitting range}
\label{sec:extrap}

The laws of Section~\ref{sec:staircase} are fitted on five budgets spanning $1.2$ decades, and the natural question is whether the training efficiency gain they describe holds beyond that range. Given the cost of a full sweep at a larger budget, we proceeded as follows: at $3{\times}10^{20}$ FLOPs we trained the ARM and E2D2 at $b\in\{16,32\}$. Each variant was trained at the three ladder sizes that bracket the optimal model size predicted by its law, and E2D2 additionally at three values of $L_{\rm dec}$ centered on the prediction of Eq.~\ref{eq:ndec}. We compare the measured NLL of each variant at the held-out budget with the prediction of the law fitted on the five smaller budgets (Figure~\ref{fig:extrap}).

\textbf{The gap between variants is reproduced; the absolute laws are not exact.} All three variants exhibit a common upward deviation of about $1.1\%$ ($24$--$28$ \mn{}) from the pure power-law extrapolation, suggesting a mild misspecification of the functional form, which Appendix~\ref{app:extrap} traces to a small irreducible term that the five budgets cannot pin down (we therefore conservatively keep the pure power-law form of SMDM and leave fits with an irreducible term to future work). Importantly, the deviation is shared across variants, and the ARM--E2D2 gap is reproduced almost exactly: $261.6$ \mn{} measured against $257$ predicted at $b{=}16$, and $296.8$ against $294$ at $b{=}32$. The quantity this paper is about, the gap between variants, is thus insensitive to the misspecification.

\subsection{Downstream performance broadly tracks likelihood}
\label{sec:bench}

\begin{wrapfigure}{r}{0.46\linewidth}
\vspace{-14pt}
\centering
\small
\captionof{table}{Difference between DUEL and the ELBO in the eight-benchmark mean accuracy, as mean $\pm$ s.d.\ over the five budgets of each variant. DUEL outperforms the ELBO with statistical significance.}
\label{tab:duelmc}
\setlength{\tabcolsep}{4pt}\resizebox{\linewidth}{!}{\begin{tabular}{lcccc}
\toprule
 & $b{=}4$ & $b{=}8$ & $b{=}16$ & $b{=}32$ \\
\midrule
MDM & \multicolumn{4}{c}{$+0.4\pm0.1$} \\
BD3LM & $+1.0\pm0.4$ & $+0.9\pm0.2$ & $+0.9\pm0.2$ & $+0.8\pm0.2$ \\
E2D2 & $+1.0\pm0.5$ & $+1.1\pm0.4$ & $+1.0\pm0.3$ & $+1.0\pm0.2$ \\
\bottomrule
\end{tabular}}
\vspace{-8pt}
\end{wrapfigure}
We hereby test whether the improvements in likelihood generalize to downstream benchmarks. Concretely, we take the union of the benchmarks of \citet{nie2025scaling} and \citet{prabhudesai2026diffusion}, eleven in all, and score the compute-optimal cell of every variant at every budget zero-shot with lm-evaluation-harness \citep{gao2021framework}. We rank the candidate answers by the exact NLL for the ARM, and by both DUEL and the $128$-sample Monte-Carlo ELBO for the diffusion variants. Three benchmarks (WinoGrande, BoolQ and COPA) are excluded because, at these scales, the models' performance on them is indistinguishable from chance or from noise (Appendix~\ref{app:bench}). For each variant and budget we record the mean accuracy over the remaining eight---HellaSwag, ARC-easy, PIQA, SciQ, RACE, SIQA, OpenBookQA and LAMBADA (Table~\ref{tab:accbench} reports every benchmark)---and find two things.

\textbf{DUEL is the better scoring rule.} As Table~\ref{tab:duelmc} shows, DUEL yields a statistically significantly higher accuracy than the ELBO for every diffusion variant. The exact readout is thus the better one, which also justifies fitting the scaling laws on DUEL rather than on the ELBO.

\textbf{Likelihood improvements carry over to downstream accuracy.} Figure~\ref{fig:bench} plots the eight-benchmark mean against compute for ARM, MDM, BD3LM ($b{=}32$), E2D2 ($b{=}32$), and E2D2 ($b{=}4$). Across variants, the order in accuracy tracks the order in NLL---the ARM, then E2D2 at $b{=}4$, E2D2 at $b{=}32$, BD3LM at $b{=}32$ and the MDM. The two orders coincide exactly at the three largest budgets and differ only within noise at the two smallest. Likelihood improvements in pretraining thus carry over to downstream accuracy. Full results on each of the eight benchmarks are reported in Table~\ref{tab:accbench} in Appendix~\ref{app:bench}.

\section{Related work}
\label{sec:related}

As diffusion language models have attracted attention for their fast, parallel inference, a number of studies have examined how they scale. \citet{nie2025scaling} were the first to study the scaling of diffusion language models: with IsoFLOP sweeps of MDMs and ARMs on SlimPajama with Llama-style transformers, they found that MDM needs $16\times$ the compute of the ARM to reach the same likelihood. Since then, a number of scaling studies of diffusion language models have followed. \citet{sahoo2026scaling} and \citet{yang2026continuous} extend the IsoFLOP methodology to other types of diffusion language model: \citet{sahoo2026scaling} find that uniform-state diffusion needs $23\times$ the compute of the ARM, and \citet{yang2026continuous} find that continuous diffusion needs $20\times$. \citet{von2026scaling} measure scaling laws for masked, uniform and hybrid-noise diffusion language models on Nemotron-CC with a CompleteP-parameterized transformer. Interestingly, although they criticize the pure power-law form of \citet{nie2025scaling} for implicitly assuming a zero irreducible loss, their own fits return an approximately zero irreducible term within their compute range, and they adopt a pure power law in the end. \citet{ni2025training} refit a Chinchilla-form scaling law for the MDM. Beyond the compute-optimal setting, \citet{prabhudesai2026diffusion} and \citet{ni2025diffusion} study the scaling of masked diffusion language models in the data-constrained regime and find that, under multi-epoch training, the MDM eventually overtakes the ARM. Our study differs from all of these. To our knowledge, ours is the first to substantially lower the $16\times$ figure, revealing that pretraining a diffusion language model from scratch is far cheaper than previously assumed.

\section{Conclusion}
In this paper we revisited the scaling laws of diffusion language models, whose widely cited $16\times$ compute gap to ARMs \citep{nie2025scaling} was measured with a methodology that has fallen out of step with current practice. We revisited and substantially revised their scaling law protocol, proposing to score the model with its exact likelihood, utilize block-aware training and the encoder--decoder architecture, and sweep the commonly used block sizes. In total, we trained about $1{,}700$ models across ten model variants, to our knowledge the largest diffusion language model scaling analysis to date, and found that pretraining diffusion language models is far cheaper than previously thought. Exact likelihood evaluation reduces the compute multiple to $12.8\times$; block diffusion training and an encoder--decoder architecture reduce it further to $7.8\times$ at block size $32$; and smaller block sizes narrow it to $4.6\times$ at block size $8$ and $3.0\times$ at block size $4$. We complemented these scaling analyses with a simple empirical rule for the optimal decoder depth of encoder--decoder block diffusion models, a held-out budget at which the gap between variants persists, and downstream benchmarks where likelihood improvements carry over to accuracy. In conclusion, the $16\times$ figure substantially overstates the compute gap between diffusion and autoregressive language models, and pretraining diffusion language models from scratch is far more viable than previously assumed.

\paragraph{Limitations.}
We hereby list the limitations of this work. Owing to the cost of the grid, all experiments use a single fixed random seed. The held-out budget was not run for $b{=}8$. When measuring DUEL we use only the top-1 confidence sampler, the default sampler of diffusion language models. The exact likelihood under other unmasking rules may differ. All experiments share one learning-rate and batch-size configuration; although \citet{porian2024resolving} show that not tuning the hyperparameters at every scale can distort a scaling law, we keep the recipe of \citet{nie2025scaling} for consistency with the protocol under which the $16\times$ was measured. Likewise, we fit pure power laws without an irreducible term for consistency with \citet{nie2025scaling}; the consistent miss at the held-out budget suggests that a small irreducible term is present (Appendix~\ref{app:extrap}), and we treat the fitted exponents as local to our compute range. It should be noted that, rather than the exact form of the scaling laws themselves, this paper is more concerned with the comparison between variants.

\subsubsection*{Reproducibility statement}
In Section~\ref{sec:protocol}, we state the training recipe, the FLOP accounting, the IsoFLOP grids and the readout and fitting procedure in full. Appendix~\ref{app:details} gives the FLOP formulas and the benchmark evaluation protocol. We also list the per-cell ledgers with the NLL of every model in Appendix~\ref{app:cells}, so that the scaling law results can be regenerated without running any code.

\subsubsection*{AI use statement}
Generative AI tools were used to help develop the code to conduct experiments and to assist with writing, including drafting certain sections of this paper. The research ideas, the experimental design, and the interpretation of the results are the authors' own. The authors take full responsibility for the content of the paper.

\subsubsection*{Acknowledgments}
We thank Gilad Turok for helpful discussions on DUEL, and Youngsuk Park for helpful discussions and feedback on the manuscript.

\bibliography{refs}
\bibliographystyle{style/iclr2027_conference}

\newpage
\appendix
\setcounter{tocdepth}{2}
\tableofcontents
\bigskip
\section{Additional background}
\label{app:background}

\subsection{Existing large-scale diffusion language models}
\label{app:blocksizes}
Table~\ref{tab:blocksizes} lists the block sizes adopted by recent large-scale masked diffusion language models.

\begin{table}[htbp]
\centering\small
\caption{Block sizes adopted by large-scale masked diffusion language models. $^\dagger$Trained with full-sequence masking; the block size is the semi-autoregressive block length used at inference (for Dream, the setting chosen by Fast-dLLM \citep{wu2026fast}).}
\label{tab:blocksizes}
\begin{tabular}{lll}
\toprule
Model & Size & Block size \\
\midrule
LLaDA$^\dagger$ \citep{nie2026large} & 8B & 32 \\
Dream$^\dagger$ \citep{ye2025dream} & 7B & 32 \\
LLaDA2.0 \citep{bie2025llada2} & 16B, 100B & 32 \\
Fast-dLLM v2 \citep{wu2026fastv2} & 1.5B, 7B & 32 \\
Nemotron-Labs-Diffusion \citep{fu2026nemotron} & 3B, 8B, 14B & 32 \\
Nemotron-TwoTower \citep{reda2026nemotron} & 30B-A3B & 16 \\
SDAR \citep{cheng2026sdar} & 1.7B, 4B, 8B, 30B-A3B & 4 (4--64 released) \\
FLARE \citep{zhu2026flare} & 2B, 4B, 9B & 4 \\
\bottomrule
\end{tabular}
\end{table}

\subsection{Attention patterns of the model variants}
\label{app:masks}
\begin{figure}[t]
\centering
\includegraphics[width=\linewidth]{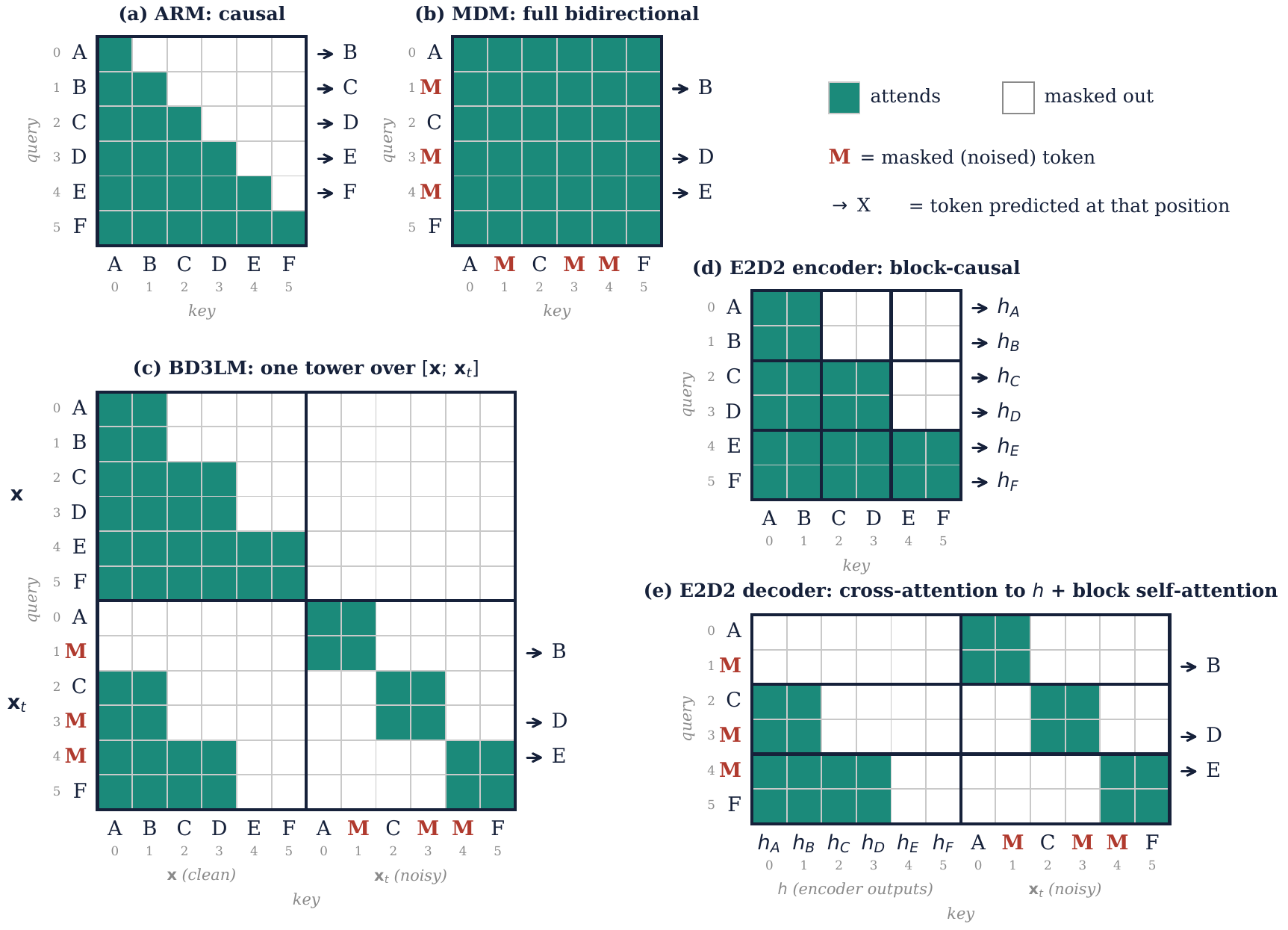}
\caption{Attention masks of the four model variants on a six-token sequence with block size $b{=}2$. Rows are queries and columns are keys; filled cells attend. (a) ARM: causal self-attention. (b) MDM: full bidirectional self-attention over the partially masked sequence. (c) BD3LM: a single tower over the concatenation of the clean sequence $\mathbf{x}$ and the noised sequence $\mathbf{x}_t$; the clean half attends block-causally, and each noised block attends to itself and to the clean tokens of all earlier blocks. (d, e) E2D2: a block-causal encoder over the clean sequence, and a decoder with block-diagonal self-attention within each noised block plus cross-attention to the encoder outputs $h$ of the strictly preceding blocks. \textcolor{red!70!black}{M} marks a masked (noised) token; arrows mark the token predicted at that position.}
\label{fig:masks}
\end{figure}

Figure~\ref{fig:masks} illustrates the attention masks of the four model variants on a short sequence, which is what the (query, key) pair counts of Appendix~\ref{app:flops} are derived from.
\begin{itemize}
\item \textbf{ARM} uses a causal mask: each position attends to itself and to all earlier positions, giving $\tfrac{n(n+1)}{2}$ pairs per layer.
\item \textbf{MDM} runs full bidirectional attention over the partially masked sequence: every position, masked or not, attends to every other position, giving $n^2$ pairs.
\item \textbf{BD3LM} keeps the decoder-only architecture but runs a single tower over the length-$2n$ concatenation of the clean sequence $\mathbf{x}$ and the noised sequence $\mathbf{x}_t$: the clean half attends block-causally to itself, and each noised block attends to itself and to the clean tokens of all earlier blocks, so that a token can be denoised given its own block and an exact clean prefix; logits are scored on the noised half only.
\item \textbf{E2D2} separates the two roles into an encoder and a decoder: the encoder runs block-causal self-attention over the clean sequence, and the decoder runs block-diagonal self-attention within each noised block together with cross-attention to the encoder outputs of the strictly preceding blocks.
\end{itemize}

\section{Additional experimental details}
\label{app:details}

\subsection{FLOP accounting}
\label{app:flops}
We count real training FLOPs identically for every variant. Unlike $6ND$, our count includes the output head and the attention score computation. Let $n$ be the sequence length, $d$ the width, $L$ the number of layers, $V$ the vocabulary size, $P$ the non-embedding parameter count, and $b$ the block size; the number of (query, key) pairs per layer depends on the attention mask: causal $\tfrac{n(n+1)}{2}$, full $n^2$, block-causal $\tfrac{n^2}{2}+\tfrac{nb}{2}$, block-diagonal $nb$, and block strict-prefix $\tfrac{n^2}{2}-\tfrac{nb}{2}$. For E2D2, the $L$ layers split into $L_{\rm enc}$ encoder and $L_{\rm dec}$ decoder layers with non-embedding parameter counts $P_{\rm enc}$ and $P_{\rm dec}$, the latter including the cross-attention projections. For each variant, the per-sample training FLOPs are computed as follows:
\begin{align*}
\text{ARM} &= 6Pn + 6dL\,n(n+1) + 6dVn,\\
\text{MDM} &= 6Pn + 12dL\,n^2 + 6dVn,\\
\text{BD3LM} &= 12Pn + 12dL\,(n^2+nb) + 6dVn,\\
\text{E2D2} &= 6P_{\rm enc}n + 6dL_{\rm enc}\,(n^2+nb) + 6P_{\rm dec}n + 12dL_{\rm dec}\,nb + 6dL_{\rm dec}\,(n^2-nb) + 6dVn.
\end{align*}
Table~\ref{tab:attnfrac} reports the attention score term as a fraction of the per-token training FLOPs for every variant across the size ladder. For BD3LM and E2D2 we take $b{=}32$ as the example, and for E2D2 we additionally set the decoder depth to half the total depth. Two patterns motivate counting real FLOPs rather than using $6ND$. First, the fraction is sizeable and grows toward smaller models, reaching about $30\%$ for the MDM, so $6ND$ distorts the shape of $\Lstar(C)$ along the ladder. Second, at matched size it differs markedly across variants: the MDM's full bidirectional attention ($n^2$) costs about twice the ARM's causal attention ($n^2/2$), while E2D2's decoder attention is the cheapest. An attention-agnostic count therefore mis-charges the variants relative to one another, and this variant-specific difference, unlike terms common to all variants, does not cancel in the compute multiple.

\begin{table}[htbp]
\centering\small
\caption{Attention score FLOPs as a percentage of per-token training FLOPs, by variant and model size ($b{=}32$; E2D2 decoder $=L/2$; sequence length $2048$).}
\label{tab:attnfrac}
\begin{tabular}{rrr cccc}
\toprule
Size (M) & $d$ & $L$ & ARM & MDM & BD3LM & E2D2 \\
\midrule
19   & 384  & 8  & 16.8 & 28.8 & 20.3 & 16.0 \\
34   & 512  & 8  & 14.4 & 25.1 & 16.9 & 13.6 \\
66   & 640  & 10 & 13.2 & 23.4 & 14.9 & 12.4 \\
113  & 768  & 12 & 12.0 & 21.5 & 13.2 & 11.2 \\
170  & 768  & 18 & 12.7 & 22.6 & 13.6 & 11.8 \\
268  & 1024 & 16 & 10.0 & 18.2 & 10.7 & 9.2  \\
472  & 1280 & 18 & 8.4  & 15.5 & 8.9  & 7.7  \\
761  & 1408 & 24 & 7.9  & 14.6 & 8.2  & 7.2  \\
1233 & 1792 & 24 & 6.4  & 12.0 & 6.6  & 5.8  \\
1678 & 2048 & 25 & 5.7  & 10.7 & 5.9  & 5.2  \\
2121 & 2176 & 28 & 5.4  & 10.2 & 5.6  & 4.9  \\
\bottomrule
\end{tabular}
\end{table}

\subsection{Scaling-law runs}
\label{app:runs}
Tables~\ref{tab:runs} and~\ref{tab:runs_e2d2} list every cell of the scaling-law grids: for each variant and budget, the ladder sizes $N$ that were trained (non-embedding parameters in millions, as named in the SMDM ladder), and for E2D2 the decoder depths $L_{\rm dec}$ swept at each size. BD3LM at $b\in\{8,16,32\}$ share one grid; BD3LM at $b{=}4$ and E2D2 at $b{=}4$ use a slightly shifted one. The held-out budget of $3{\times}10^{20}$ FLOPs (Section~\ref{sec:extrap}) is listed in its own rows.

\begin{table}[htbp]
\centering
\footnotesize
\caption{Ladder sizes $N$ (millions of non-embedding parameters) trained at each budget for the ARM, the MDM and BD3LM.}
\label{tab:runs}
\setlength{\tabcolsep}{6pt}
\begin{tabular}{lll}
\toprule
variant & $C$ (FLOPs) & trained sizes $N$ \\
\midrule
\multirow{6}{*}{ARM} & $6{\times}10^{18}$ & $\{66, 113, 170, 231, 336, 472, 629\}$ \\
 & $10^{19}$ & $\{85, 142, 231, 336, 472, 666, 944\}$ \\
 & $3{\times}10^{19}$ & $\{142, 231, 336, 472, 666, 944, 1233\}$ \\
 & $6{\times}10^{19}$ & $\{206, 302, 472, 666, 944, 1233, 1678\}$ \\
 & $10^{20}$ & $\{268, 472, 666, 944, 1233, 1678, 2121\}$ \\
 & $3{\times}10^{20}$ (held-out) & $\{1678, 2121, 2718\}$ \\
\midrule
\multirow{5}{*}{MDM} & $6{\times}10^{18}$ & $\{34, 48, 66, 113, 170, 231, 336\}$ \\
 & $10^{19}$ & $\{48, 66, 85, 142, 231, 336, 472\}$ \\
 & $3{\times}10^{19}$ & $\{66, 142, 231, 336, 472, 666, 944\}$ \\
 & $6{\times}10^{19}$ & $\{113, 206, 302, 472, 666, 944, 1233\}$ \\
 & $10^{20}$ & $\{142, 268, 472, 666, 944, 1233, 1678\}$ \\
\midrule
\multirow{5}{*}{BD3LM $b\in\{8,16,32\}$} & $6{\times}10^{18}$ & $\{19, 34, 48, 66, 85, 113, 142, 170, 231, 336\}$ \\
 & $10^{19}$ & $\{48, 66, 85, 113, 142, 170, 206, 268, 336\}$ \\
 & $3{\times}10^{19}$ & $\{85, 113, 142, 170, 206, 268, 336, 472, 666\}$ \\
 & $6{\times}10^{19}$ & $\{142, 170, 206, 268, 336, 472, 666, 944\}$ \\
 & $10^{20}$ & $\{206, 268, 336, 472, 666, 944, 1233\}$ \\
\midrule
\multirow{5}{*}{BD3LM $b{=}4$} & $6{\times}10^{18}$ & $\{34, 48, 66, 85, 113, 142, 170, 231, 336\}$ \\
 & $10^{19}$ & $\{48, 66, 85, 113, 142, 170, 206, 268, 336\}$ \\
 & $3{\times}10^{19}$ & $\{142, 170, 206, 268, 336, 472, 666\}$ \\
 & $6{\times}10^{19}$ & $\{206, 268, 336, 472, 666, 944, 1233\}$ \\
 & $10^{20}$ & $\{268, 336, 472, 666, 944, 1233, 1678\}$ \\
\bottomrule
\end{tabular}
\end{table}

\begin{table}[htbp]
\centering
\footnotesize
\caption{E2D2 grid: for each budget and ladder size $N$ (with total depth $L$), the range of decoder depths $L_{\rm dec}$ trained at each block size. ``---'' means the size was not trained at that block size.}
\label{tab:runs_e2d2}
\setlength{\tabcolsep}{5pt}
\renewcommand{\arraystretch}{0.95}
\begin{tabular}{lrrllll}
\toprule
$C$ (FLOPs) & $N$ (M) & $L$ & $b{=}4$ & $b{=}8$ & $b{=}16$ & $b{=}32$ \\
\midrule
$6{\times}10^{18}$ & 34 & 8 & 2--7 & 2--7 & 2--7 & 2--7 \\
 & 48 & 9 & 2--8 & 2--8 & 2--8 & 2--8 \\
 & 66 & 10 & 2--9 & 2--9 & 2--9 & 3--9 \\
 & 85 & 13 & 2--12 & 4--12 & 4--12 & 5--12 \\
 & 113 & 12 & 2--11 & 4--11 & 3--11 & 3--11 \\
 & 142 & 15 & --- & --- & 6--14 & 5--14 \\
 & 170 & 18 & 2--17 & 8--16 & 8--17 & 8--17 \\
 & 231 & 18 & 2--17 & 7--15 & 8--17 & 7--17 \\
 & 336 & 20 & 2--19 & --- & --- & --- \\
 & 472 & 18 & 2--17 & --- & --- & --- \\
\midrule
$10^{19}$ & 66 & 10 & 4--9 & 3--9 & 2--9 & 3--9 \\
 & 85 & 13 & 4--12 & 5--12 & 4--11 & 6--12 \\
 & 113 & 12 & 4--11 & 4--11 & 3--11 & 5--11 \\
 & 142 & 15 & 3--14 & 6--14 & 6--14 & 8--14 \\
 & 170 & 18 & 4--17 & 9--17 & 8--17 & 11--17 \\
 & 231 & 18 & 4--17 & 8--16 & 6--16 & 10--16 \\
 & 336 & 20 & 4--19 & 8--16 & 8--18 & 10--16 \\
 & 472 & 18 & 4--17 & --- & --- & --- \\
\midrule
$3{\times}10^{19}$ & 142 & 15 & 4--14 & 8--14 & 6--14 & 8--14 \\
 & 170 & 18 & --- & --- & 8--17 & 11--17 \\
 & 206 & 16 & 4--15 & 8--15 & 7--15 & 9--15 \\
 & 268 & 16 & 3--15 & 8--15 & 7--15 & 9--15 \\
 & 336 & 20 & 4--19 & 10--18 & 9--19 & 12--18 \\
 & 472 & 18 & 4--17 & 8--16 & 8--17 & 9--17 \\
 & 666 & 21 & 4--20 & 11--19 & 10--20 & 11--20 \\
 & 944 & 25 & 4--24 & --- & --- & --- \\
\midrule
$6{\times}10^{19}$ & 170 & 18 & --- & --- & 7--17 & 11--17 \\
 & 206 & 16 & 4--14 & 8--15 & 7--15 & 9--15 \\
 & 268 & 16 & 4--14 & 8--15 & 6--15 & 9--15 \\
 & 336 & 20 & 4--19 & 10--18 & 10--19 & 13--19 \\
 & 472 & 18 & 3--17 & 8--16 & 8--17 & 10--16 \\
 & 666 & 21 & 4--20 & 10--18 & 10--20 & 12--20 \\
 & 944 & 25 & 4--24 & 14--22 & 14--24 & 15--21 \\
 & 1233 & 24 & 4--23 & --- & --- & --- \\
\midrule
$10^{20}$ & 206 & 16 & --- & --- & 7--15 & 9--15 \\
 & 268 & 16 & 5--14 & 8--15 & 7--15 & 9--15 \\
 & 336 & 20 & 4--19 & 11--19 & 11--19 & 13--19 \\
 & 472 & 18 & 4--17 & 8--15 & 8--17 & 11--17 \\
 & 666 & 21 & 3--20 & 10--18 & 11--20 & 13--19 \\
 & 944 & 25 & 4--24 & 14--22 & 14--24 & 16--22 \\
 & 1233 & 24 & 4--23 & 12--20 & 14--23 & 14--23 \\
 & 1678 & 25 & 4--24 & --- & --- & --- \\
\midrule
$3{\times}10^{20}$ (held-out) & 831 & 22 & --- & --- & 15--17 & 16--18 \\
 & 944 & 25 & --- & --- & 16--20 & 18--20 \\
 & 1233 & 24 & --- & --- & 16--18 & 17--21 \\
\bottomrule
\end{tabular}
\end{table}

\subsection{Compute cost}
\label{app:cost}
Table~\ref{tab:gpuhours} lists the cost of the scaling law runs in B200 GPU hours, split into training and DUEL evaluation, for a total of approximately $136$k GPU hours over $1{,}741$ training runs (downstream benchmark evaluation not included). The E2D2 rows include the decoder-depth sweep, which is why they dominate the total. DUEL evaluation is by far the most expensive for the MDM, whose unmasking sweep spans the full sequence rather than one block at a time.

\begin{table}[htbp]
\centering
\footnotesize
\caption{Compute cost in B200 GPU hours by variant, split into training and DUEL evaluation. E2D2 rows include the decoder-depth sweep.}
\label{tab:gpuhours}
\setlength{\tabcolsep}{7pt}
\begin{tabular}{lrrrr}
\toprule
variant & runs & training & DUEL evaluation & total \\
\midrule
ARM & 43 & 1{,}507 & --- & 1{,}507 \\
MDM & 80 & 2{,}393 & 8{,}384 & 10{,}777 \\
BD3LM $b{=}32$ & 43 & 3{,}901 & 292 & 4{,}193 \\
BD3LM $b{=}16$ & 43 & 3{,}903 & 147 & 4{,}050 \\
BD3LM $b{=}8$ & 43 & 3{,}903 & 74 & 3{,}977 \\
BD3LM $b{=}4$ & 40 & 3{,}504 & 39 & 3{,}543 \\
E2D2 $b{=}32$ & 285 & 22{,}551 & 1{,}155 & 23{,}706 \\
E2D2 $b{=}16$ & 355 & 28{,}082 & 718 & 28{,}800 \\
E2D2 $b{=}8$ & 270 & 18{,}349 & 269 & 18{,}618 \\
E2D2 $b{=}4$ & 539 & 36{,}960 & 327 & 37{,}287 \\
\midrule
total & 1{,}741 & 125{,}053 & 11{,}405 & 136{,}458 \\
\bottomrule
\end{tabular}
\end{table}

\subsection{Benchmark evaluation protocol}
\label{app:benchprotocol}
We obtain benchmark accuracies with lm-evaluation-harness \citep{gao2021framework}. For HellaSwag \citep{zellers2019hellaswag}, ARC-easy \citep{clark2018think}, PIQA \citep{bisk2020piqa}, SciQ \citep{welbl2017crowdsourcing}, RACE \citep{lai2017race} and OpenBookQA \citep{mihaylov2018can} we report the \texttt{acc\_norm} metric returned by the harness, and for SIQA \citep{sap2019social}, LAMBADA \citep{paperno2016lambada}, WinoGrande \citep{sakaguchi2021winogrande}, BoolQ \citep{clark2019boolq} and COPA \citep{roemmele2011choice} we report \texttt{acc}. The diffusion variants are scored with DUEL (top-1 confidence) and with the $128$-sample ELBO. For LAMBADA the two readouts coincide, since the task is a single-word completion.

\section{Additional experimental results}
\label{app:results}

\subsection{IsoFLOP curves}
\label{app:isoflopcurves}
\begin{figure}[h]
\centering
\includegraphics[width=0.62\linewidth]{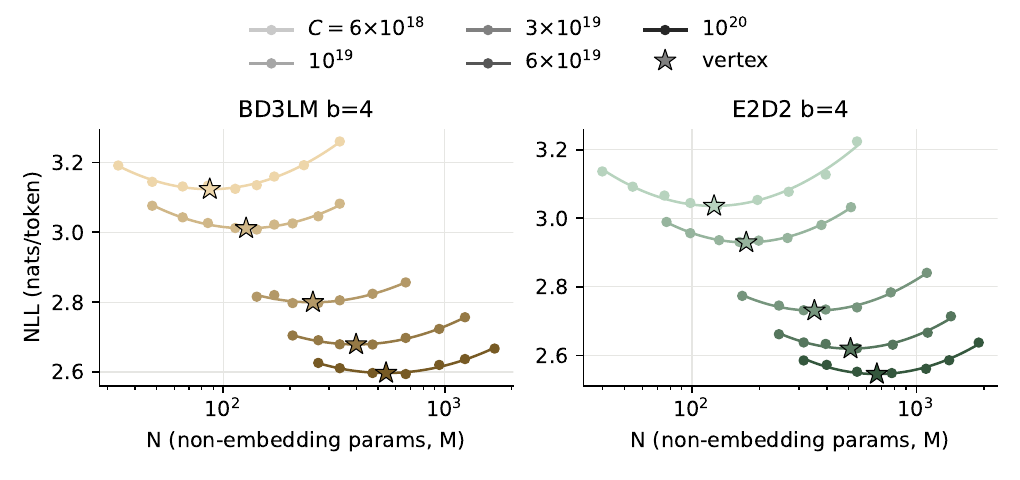}
\caption{IsoFLOP curves for BD3LM and E2D2 at $b{=}4$ (DUEL; for E2D2, each point is the best $L_{\rm dec}$ at that size). Darker shades mark larger budgets, and stars mark fitted vertices.}
\label{fig:isoflop_b4}
\end{figure}

Figure~\ref{fig:isoflop_b4} shows the IsoFLOP curves of the two $b{=}4$ variants, which complement the eight variants in Figure~\ref{fig:isoflop}.

\subsection{Compute multiple}
\label{app:isoflop}
Table~\ref{tab:mult_protocol} reports the compute multiple of every diffusion variant at every budget under the protocol of Section~\ref{sec:protocol}: the variant's measured compute-optimal NLL at budget $C$ is located on the ARM's fitted law, and the multiple is $C$ divided by the ARM compute that reaches the same NLL. Below $6{\times}10^{19}$ FLOPs, the NLL of most diffusion variants lies above the ARM's measured range ($2.39$--$2.84$ nats), so the ARM law has to be extrapolated below $6{\times}10^{18}$ FLOPs; these entries are shown in parentheses for reference only. Table~\ref{tab:mult_extrap} reports the alternative, extrapolating readout of SMDM \citep{nie2025scaling}: the diffusion variant's fitted law is extrapolated to the compute at which it reaches the ARM's compute-optimal NLL at budget $C$, and the multiple is that compute divided by $C$. Since the fitted exponents of the diffusion variants are slightly steeper than the ARM's, this readout decreases with $C$; at $10^{20}$ FLOPs it differs from Table~\ref{tab:mult_protocol} by at most $22\%$ (the MDM under DUEL), and the ordering of the variants is the same under both readouts.

\begin{table}[htbp]
\scriptsize\setlength{\tabcolsep}{2.8pt}
\begin{minipage}[t]{0.49\linewidth}\centering
\caption{Compute multiple against the ARM under the protocol of the main text (DUEL for diffusion unless stated), at each of the five budgets. Values in parentheses require extrapolating the ARM law below $6{\times}10^{18}$ FLOPs and are shown for reference only; the last column is the value reported in Table~\ref{tab:readout}.}
\label{tab:mult_protocol}
\begin{tabular}{lrrrrr}
\toprule
variant & $6{\times}10^{18}$ & $10^{19}$ & $3{\times}10^{19}$ & $6{\times}10^{19}$ & $10^{20}$ \\
\midrule
MDM (ELBO)     & (17.5) & (17.6) & (16.3) & (16.5) & 17.8 \\
MDM            & (14.6) & (15.1) & (12.6) & (12.4) & 12.8 \\
BD3LM $b{=}32$ & (11.4) & (10.7) & (9.9)  & 9.7    & 9.6 \\
BD3LM $b{=}16$ & (9.3)  & (8.8)  & (8.0)  & 7.9    & 7.8 \\
BD3LM $b{=}8$  & (7.1)  & (6.6)  & (6.1)  & 6.0    & 5.9 \\
BD3LM $b{=}4$  & (4.9)  & (4.5)  & 4.2    & 4.1    & 4.1 \\
E2D2 $b{=}32$  & (8.1)  & (7.5)  & (7.3)  & 7.5    & 7.8 \\
E2D2 $b{=}16$  & (6.7)  & (6.3)  & (5.9)  & 6.0    & 6.3 \\
E2D2 $b{=}8$   & (5.0)  & (4.6)  & 4.4    & 4.5    & 4.6 \\
E2D2 $b{=}4$   & (3.1)  & (2.9)  & 2.8    & 2.9    & 3.0 \\
\bottomrule
\end{tabular}
\end{minipage}\hfill
\begin{minipage}[t]{0.49\linewidth}\centering
\caption{Compute multiple against the ARM under the extrapolating readout of SMDM (DUEL for diffusion unless stated): the fitted law of the diffusion variant is extrapolated to the compute at which it reaches the ARM's compute-optimal NLL at each budget.}
\label{tab:mult_extrap}
\begin{tabular}{lrrrrr}
\toprule
variant & $6{\times}10^{18}$ & $10^{19}$ & $3{\times}10^{19}$ & $6{\times}10^{19}$ & $10^{20}$ \\
\midrule
MDM (ELBO)     & 16.9 & 16.8 & 16.7 & 16.6 & 16.5 \\
MDM            & 12.5 & 12.1 & 11.3 & 10.8 & 10.5 \\
BD3LM $b{=}32$ & 9.8  & 9.5  & 8.9  & 8.6  & 8.4 \\
BD3LM $b{=}16$ & 8.0  & 7.8  & 7.3  & 7.0  & 6.8 \\
BD3LM $b{=}8$  & 6.2  & 6.0  & 5.6  & 5.4  & 5.3 \\
BD3LM $b{=}4$  & 4.3  & 4.2  & 4.0  & 3.8  & 3.7 \\
E2D2 $b{=}32$  & 7.6  & 7.6  & 7.5  & 7.4  & 7.4 \\
E2D2 $b{=}16$  & 6.2  & 6.1  & 5.9  & 5.8  & 5.8 \\
E2D2 $b{=}8$   & 4.6  & 4.5  & 4.4  & 4.3  & 4.3 \\
E2D2 $b{=}4$   & 2.9  & 2.9  & 2.9  & 2.8  & 2.8 \\
\bottomrule
\end{tabular}
\end{minipage}
\end{table}

\subsection{Decoder depth}
\label{app:ndec}

\begin{table}[htbp]
\centering
\small
\caption{Optimal decoder fraction $L^\ast_{\rm dec}/L$ of E2D2 at each block size, read off in three ways: the discrete argmin of DUEL over all columns, the vertex of a parabola fitted near the minimum over all columns, and the argmin restricted to the compute-optimal $N$ of each budget. Entries are mean $\pm$ s.d.}
\label{tab:ndec}
\begin{tabular}{lcccc}
\toprule
$b$ & 4 & 8 & 16 & 32 \\
\midrule
argmin, all columns & $0.645\pm0.085$ & $0.716\pm0.040$ & $0.728\pm0.062$ & $0.780\pm0.055$ \\
vertex, all columns & $0.643\pm0.064$ & $0.682\pm0.029$ & $0.723\pm0.030$ & $0.771\pm0.049$ \\
at compute-optimal $N$ & $0.630\pm0.045$ & $0.706\pm0.048$ & $0.724\pm0.034$ & $0.756\pm0.018$ \\
\bottomrule
\end{tabular}
\end{table}

\begin{figure}[htbp]
\centering
\includegraphics[width=0.62\linewidth]{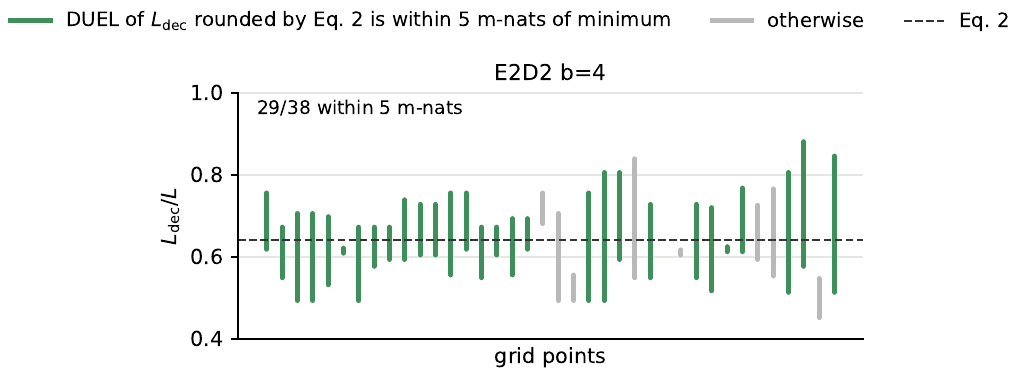}
\caption{Same as Figure~\ref{fig:ndec}, for E2D2 at $b{=}4$.}
\label{fig:ndec_b4}
\end{figure}

\paragraph{Three readouts of the optimal split.}
Table~\ref{tab:ndec} reads $L^\ast_{\rm dec}/L$ off the decoder-depth sweeps in three ways: the discrete argmin of DUEL, the vertex of a parabola fitted near the minimum (the readout used in the main text), and the argmin restricted to the compute-optimal $N$ of each budget. All three agree closely at every block size, so the optimal split is insensitive to how it is read.

\paragraph{The smallest block size.}
Figure~\ref{fig:ndec_b4} repeats the analysis of Section~\ref{sec:ndec} for $b{=}4$ ($38$ columns). The picture is the same as at the larger block sizes: the optimal split is decoder-heavy, and the three readouts agree ($0.63$--$0.65$, against $0.64$ predicted by Eq.~\ref{eq:ndec}). Rounding Eq.~\ref{eq:ndec} to the nearest integer number of decoder layers lands within $10$ \mn{} of the column minimum in $87\%$ of columns and in all five compute-optimal columns (Table~\ref{tab:rounding}).

\paragraph{Regressions.}
We regress the parabola-vertex readout of $L^\ast_{\rm dec}/L$ over the 142 columns on $\log_{10}N$, $\log_{10}C$ and $\log_2 b$ jointly. The block size dominates: its coefficient is $+0.043$ per doubling of $b$ ($t{=}12.7$). The model size has no detectable effect, with a coefficient of $-0.007$ per decade of $N$ ($t{=}-0.5$). The budget has a small one, $+0.029$ per decade of $C$ ($t{=}2.4$); across the $1.2$ decades of our budgets this amounts to a shift of $0.035$ in $L^\ast_{\rm dec}/L$, less than one layer at every depth in the grid, whereas going from $b{=}4$ to $b{=}32$ shifts it by $0.13$. We therefore regress on $\log_2 b$ alone, which gives Eq.~\ref{eq:ndec}. The rule describes the data well: it reproduces the mean vertex at each block size to within $0.005$ ($0.643$, $0.682$, $0.723$ and $0.771$ measured at $b=4$, $8$, $16$ and $32$, against $0.642$, $0.685$, $0.728$ and $0.771$ predicted), and its residuals over the individual columns have an rms of $0.045$ in $L^\ast_{\rm dec}/L$, or $0.8$ layers.

\begin{table}[htbp]
\centering
\footnotesize
\caption{Rounded Eq.~\ref{eq:ndec} against the column minimum, by block size.}
\label{tab:rounding}
\setlength{\tabcolsep}{7pt}
\begin{tabular}{lcccc}
\toprule
$b$ & columns & within $5$ \mn{} & within $10$ \mn{} & compute-optimal columns within $10$ \mn{} \\
\midrule
4 & 38 & $76\%$ & $87\%$ & 5/5 \\
8 & 32 & $81\%$ & $88\%$ & 5/5 \\
16 & 36 & $89\%$ & $97\%$ & 5/5 \\
32 & 36 & $89\%$ & $100\%$ & 5/5 \\
\midrule
all & 142 & $84\%$ & $93\%$ & 20/20 \\
\bottomrule
\end{tabular}
\end{table}

\subsection{Held-out budget}
\label{app:extrap}
\label{app:eform}
Table~\ref{tab:extrap} compares, at the held-out budget of $3{\times}10^{20}$ FLOPs, the NLL predicted by the power law fitted on the five budgets $\le10^{20}$ with the best measured cell of the ARM and of E2D2 at $b\in\{16,32\}$. As reported in Section~\ref{sec:extrap}, the gap between E2D2 and the ARM is reproduced: $261.6$ \mn{} measured against $257$ predicted at $b{=}16$, and $296.8$ against $294$ at $b{=}32$. At the same time, all three variants land $24$--$28$ \mn{} ($1.07$--$1.15\%$) above their own law, a miss that is consistent in sign and size across the three.

\begin{table}[htbp]
\centering
\small
\caption{The held-out budget at $3{\times}10^{20}$ FLOPs. Power laws fitted on the five budgets $\le10^{20}$ are compared with the best measured cell.}
\label{tab:extrap}
\begin{tabular}{lrrr}
\toprule
variant & law $\Lstar(3{\times}10^{20})$ & measured best cell & miss \\
\midrule
ARM & 2.2231 & 2.2468 & $+23.7$ \mn{} ($+1.07\%$) \\
E2D2 $b{=}16$ & 2.4800 & 2.5084 & $+28.4$ \mn{} ($+1.15\%$) \\
E2D2 $b{=}32$ & 2.5167 & 2.5436 & $+26.9$ \mn{} ($+1.07\%$) \\
\bottomrule
\end{tabular}
\end{table}

\begin{figure}[htbp]
\centering
\includegraphics[width=0.6\linewidth]{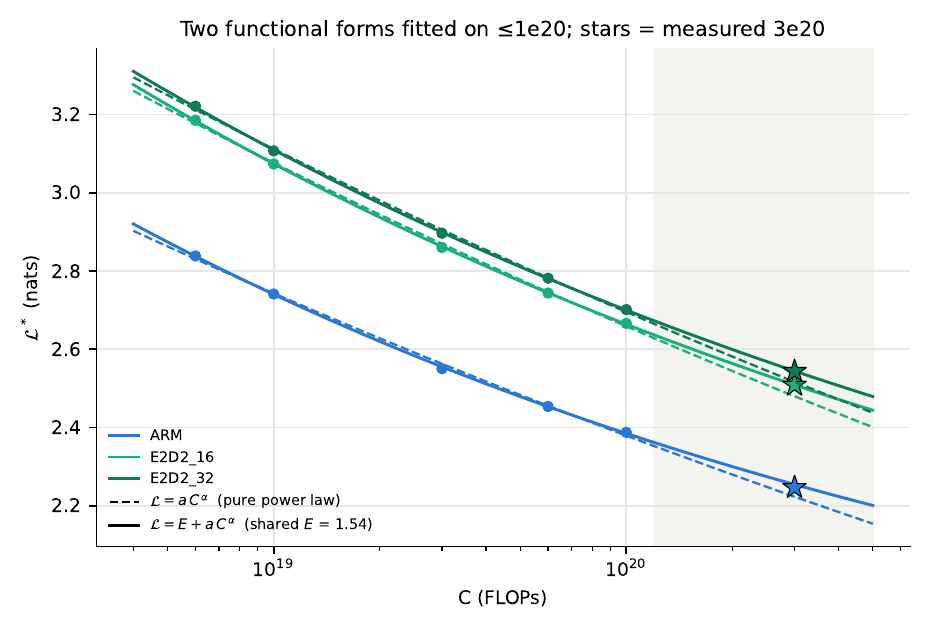}
\caption{Pure power laws (dashed) and shared-$E$ laws $\Lstar=E+a\,C^\alpha$ (solid), both fitted on $\le10^{20}$, against the held-out budget at $3{\times}10^{20}$ FLOPs (stars).}
\label{fig:eform}
\end{figure}

\paragraph{Fits with an irreducible term.}
A pure power law has no irreducible term, whereas the loss is bounded below by the entropy of the data \citep{hoffmann2022training}, a known weakness of the pure power law \citep{von2026scaling,ni2025training}: a curve that bends towards a floor is under-predicted by a straight line in log--log space, by more the further one extrapolates, which is the signature observed. We therefore refit $\Lstar(C)=E+a\,C^\alpha$ jointly over the eight DUEL variants (ARM, MDM, BD3LM and E2D2 at $b\in\{4,16,32\}$) with a single shared $E$ and per-variant $(a,\alpha)$, on the five budgets $\le10^{20}$, and profile the fit over fixed values of $E$ (Table~\ref{tab:eprofile}). Two observations follow. First, the in-sample rms decreases from $5.9$ \mn{} at $E=0$ to $3.1$ at $E\approx1.5$, so the five budgets favor $E>0$; but the minimum is shallow ($3.1$--$3.9$ \mn{} for $E\in[1.2,1.8]$), and the exponents move along a ridge with $E$, so the five budgets cannot determine the value of $E$. Second, adding $E\approx1.5$ removes the miss at the held-out budget: at $E=1.5$ the three $3{\times}10^{20}$ cells are predicted to $-7$, $+0.4$ and $0.0$ \mn{} (Figure~\ref{fig:eform}). However, the held-out cells come from three sizes bracketing the predicted optimal model size rather than from a full IsoFLOP search, so we do not use them to fix $E$, and we conservatively keep the pure power law in the body of the paper, treating its exponents as local to the fitted range. We leave a scaling law with an irreducible term, which would require more budgets to pin down $E$, to future work.

\begin{table}[htbp]
\centering
\small
\caption{Profile of the shared-$E$ fit on the five budgets $\le10^{20}$: in-sample rms, exponents, and the miss at the held-out budget (measured minus predicted, \mn{}).}
\label{tab:eprofile}
\begin{tabular}{rrrrrrr}
\toprule
$E$ & rms (\mn) & $\alpha_{\rm ARM}$ & $\alpha_{\text{E2D2 }16}$ & miss ARM & miss E2D2 16 & miss E2D2 32 \\
\midrule
0.0 & 5.9 & $-0.062$ & $-0.063$ & $+23.7$ & $+28.4$ & $+26.9$ \\
1.0 & 4.1 & $-0.101$ & $-0.097$ & $+9.2$ & $+14.2$ & $+13.2$ \\
1.2 & 3.7 & $-0.115$ & $-0.108$ & $+4.1$ & $+9.5$ & $+8.7$ \\
1.4 & 3.2 & $-0.134$ & $-0.122$ & $-2.5$ & $+3.8$ & $+3.2$ \\
1.5 & 3.1 & $-0.147$ & $-0.131$ & $-6.7$ & $+0.4$ & $-0.0$ \\
1.6 & 3.2 & $-0.161$ & $-0.141$ & $-11.5$ & $-3.5$ & $-3.7$ \\
1.8 & 3.9 & $-0.203$ & $-0.167$ & $-24.1$ & $-13.0$ & $-12.7$ \\
\bottomrule
\end{tabular}
\end{table}

\subsection{Downstream benchmarks}
\label{app:bench}

\begin{figure}[htbp]
\centering
\includegraphics[width=0.62\linewidth]{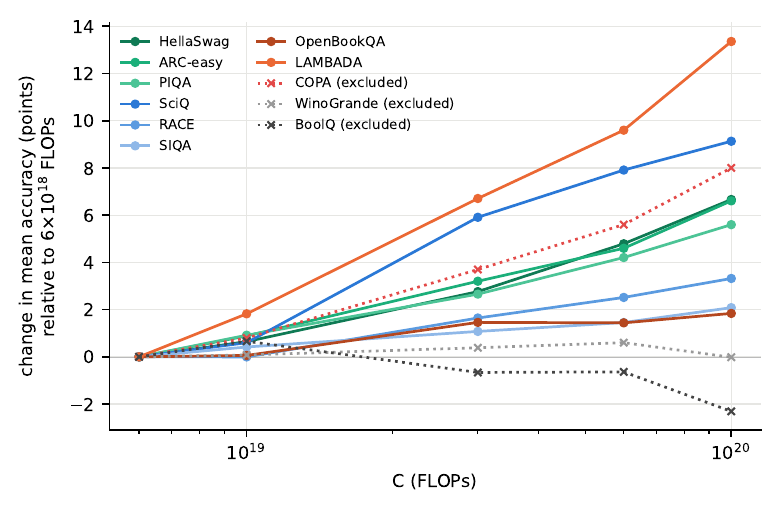}
\caption{Benchmark screening. For each benchmark, the accuracy averaged over the ten variants at their compute-optimal cells, relative to its value at $6{\times}10^{18}$ FLOPs. Dotted lines mark the three benchmarks we exclude.}
\label{fig:bench_full}
\end{figure}

\paragraph{Benchmark screening.}
We score the compute-optimal cell of every variant at every budget on all eleven benchmarks. Figure~\ref{fig:bench_full} shows, for each benchmark, how the accuracy averaged over the ten variants moves with compute. Eight benchmarks rise with compute by well more than their standard error. WinoGrande and BoolQ do not: on WinoGrande accuracy stays at chance at every budget ($50.8\%$ at $10^{20}$ FLOPs against a chance level of $50\%$), and on BoolQ it decreases with compute, so neither can distinguish a better model from a worse one at these scales. COPA does rise with compute, but it has only $100$ questions, so the standard error of a single COPA accuracy is $4.8$ points, against at most $2$ points for every other benchmark; in a mean over nine benchmarks, COPA would account for about two-thirds of the variance. We therefore exclude these three benchmarks and use the remaining eight.

\paragraph{Per-benchmark accuracy.}
Table~\ref{tab:accbench} lists the accuracy on each of the eight retained benchmarks, and their mean, for the compute-optimal cell of every variant at every budget; the means are the values plotted in Figure~\ref{fig:bench}.

\begin{table}[htbp]
\centering
\scriptsize\setlength{\tabcolsep}{3.5pt}\renewcommand{\arraystretch}{0.85}
\caption{Zero-shot accuracy (\%) on the eight retained benchmarks at the compute-optimal cell of every variant at every budget (DUEL scoring for the diffusion variants; exact NLL for the ARM), and their mean. This is the data plotted in Figure~\ref{fig:bench}.}
\label{tab:accbench}
\begin{tabular}{llrrrrrrrrr}
\toprule
variant & $C$ & HellaSwag & ARC-e & PIQA & SciQ & RACE & SIQA & OBQA & LAMBADA & mean \\
\midrule
ARM & $6{\times}10^{18}$ & 28.9 & 35.1 & 61.1 & 60.1 & 27.0 & 37.0 & 26.2 & 16.6 & 36.5 \\
 & $10^{19}$ & 30.2 & 37.7 & 60.8 & 60.6 & 28.3 & 37.0 & 27.4 & 19.0 & 37.6 \\
 & $3{\times}10^{19}$ & 34.1 & 40.4 & 63.2 & 65.9 & 29.8 & 37.4 & 27.2 & 26.3 & 40.5 \\
 & $6{\times}10^{19}$ & 37.1 & 41.8 & 65.6 & 67.3 & 29.1 & 37.1 & 29.4 & 27.2 & 41.8 \\
 & $10^{20}$ & 40.1 & 42.9 & 67.7 & 72.4 & 33.3 & 38.3 & 30.2 & 30.8 & 44.5 \\
\addlinespace[1pt]
MDM & $6{\times}10^{18}$ & 28.5 & 32.1 & 56.2 & 63.1 & 27.0 & 36.7 & 25.4 & 21.1 & 36.3 \\
 & $10^{19}$ & 28.7 & 34.2 & 56.6 & 61.0 & 27.8 & 36.3 & 24.8 & 17.9 & 35.9 \\
 & $3{\times}10^{19}$ & 30.5 & 36.3 & 58.8 & 64.7 & 30.1 & 37.3 & 25.0 & 24.2 & 38.4 \\
 & $6{\times}10^{19}$ & 31.9 & 36.6 & 59.6 & 66.0 & 31.8 & 36.5 & 26.8 & 26.9 & 39.5 \\
 & $10^{20}$ & 33.4 & 39.2 & 61.0 & 68.5 & 32.0 & 36.7 & 26.8 & 29.6 & 40.9 \\
\addlinespace[1pt]
BD3LM $b{=}32$ & $6{\times}10^{18}$ & 28.6 & 32.7 & 56.1 & 60.1 & 27.6 & 34.9 & 25.4 & 16.1 & 35.2 \\
 & $10^{19}$ & 29.3 & 33.4 & 56.8 & 62.0 & 26.8 & 36.0 & 25.2 & 17.5 & 35.9 \\
 & $3{\times}10^{19}$ & 30.2 & 36.1 & 58.5 & 67.5 & 29.1 & 36.1 & 27.2 & 23.9 & 38.6 \\
 & $6{\times}10^{19}$ & 32.3 & 37.2 & 60.2 & 70.0 & 29.9 & 38.0 & 26.4 & 25.8 & 40.0 \\
 & $10^{20}$ & 32.9 & 40.3 & 61.5 & 72.5 & 30.2 & 38.0 & 24.8 & 29.9 & 41.3 \\
\addlinespace[1pt]
BD3LM $b{=}16$ & $6{\times}10^{18}$ & 28.3 & 33.5 & 55.8 & 60.0 & 27.8 & 35.6 & 23.6 & 15.5 & 35.0 \\
 & $10^{19}$ & 28.6 & 34.3 & 57.9 & 60.7 & 28.1 & 35.5 & 25.8 & 19.4 & 36.3 \\
 & $3{\times}10^{19}$ & 30.6 & 36.2 & 58.9 & 67.5 & 27.8 & 36.2 & 28.0 & 24.3 & 38.7 \\
 & $6{\times}10^{19}$ & 32.4 & 38.4 & 60.3 & 70.5 & 30.0 & 37.3 & 26.2 & 28.0 & 40.4 \\
 & $10^{20}$ & 34.3 & 40.2 & 61.6 & 71.2 & 30.2 & 37.0 & 27.4 & 30.8 & 41.6 \\
\addlinespace[1pt]
BD3LM $b{=}8$ & $6{\times}10^{18}$ & 28.3 & 34.1 & 57.2 & 60.7 & 27.1 & 34.9 & 26.4 & 16.6 & 35.7 \\
 & $10^{19}$ & 29.0 & 34.4 & 58.3 & 61.4 & 28.3 & 36.4 & 24.6 & 18.6 & 36.4 \\
 & $3{\times}10^{19}$ & 30.6 & 35.8 & 58.9 & 68.6 & 28.7 & 37.7 & 27.2 & 24.2 & 39.0 \\
 & $6{\times}10^{19}$ & 32.8 & 37.2 & 60.9 & 70.3 & 30.5 & 37.0 & 25.2 & 26.3 & 40.1 \\
 & $10^{20}$ & 34.4 & 39.9 & 60.6 & 71.2 & 30.5 & 38.0 & 27.4 & 32.4 & 41.8 \\
\addlinespace[1pt]
BD3LM $b{=}4$ & $6{\times}10^{18}$ & 28.4 & 32.7 & 55.7 & 61.1 & 28.3 & 36.2 & 27.0 & 17.4 & 35.9 \\
 & $10^{19}$ & 28.7 & 34.6 & 58.4 & 60.8 & 26.1 & 36.9 & 25.2 & 19.2 & 36.2 \\
 & $3{\times}10^{19}$ & 31.1 & 36.4 & 59.7 & 64.9 & 29.7 & 37.1 & 25.8 & 23.2 & 38.5 \\
 & $6{\times}10^{19}$ & 33.3 & 37.7 & 60.6 & 67.8 & 29.4 & 36.1 & 27.8 & 25.7 & 39.8 \\
 & $10^{20}$ & 35.4 & 40.7 & 63.4 & 66.8 & 30.6 & 37.8 & 27.8 & 31.5 & 41.8 \\
\addlinespace[1pt]
E2D2 $b{=}32$ & $6{\times}10^{18}$ & 28.2 & 33.5 & 56.3 & 64.3 & 27.8 & 35.3 & 26.6 & 14.4 & 35.8 \\
 & $10^{19}$ & 28.7 & 34.6 & 58.1 & 65.6 & 27.8 & 35.2 & 26.2 & 17.5 & 36.7 \\
 & $3{\times}10^{19}$ & 30.5 & 37.0 & 59.1 & 70.3 & 28.7 & 36.8 & 28.4 & 21.1 & 39.0 \\
 & $6{\times}10^{19}$ & 32.3 & 37.7 & 60.5 & 72.4 & 31.4 & 36.6 & 26.8 & 24.6 & 40.3 \\
 & $10^{20}$ & 34.0 & 38.4 & 61.4 & 72.2 & 30.6 & 38.4 & 29.0 & 28.1 & 41.5 \\
\addlinespace[1pt]
E2D2 $b{=}16$ & $6{\times}10^{18}$ & 28.6 & 33.7 & 56.6 & 64.0 & 28.0 & 34.8 & 26.0 & 17.1 & 36.1 \\
 & $10^{19}$ & 29.6 & 32.9 & 57.2 & 63.6 & 29.9 & 35.5 & 26.4 & 17.6 & 36.6 \\
 & $3{\times}10^{19}$ & 31.2 & 36.3 & 58.9 & 68.2 & 29.8 & 36.1 & 26.8 & 20.8 & 38.5 \\
 & $6{\times}10^{19}$ & 33.0 & 37.4 & 60.0 & 71.8 & 28.9 & 37.8 & 27.0 & 26.6 & 40.3 \\
 & $10^{20}$ & 34.9 & 39.5 & 61.1 & 71.0 & 30.7 & 37.6 & 27.2 & 28.5 & 41.3 \\
\addlinespace[1pt]
E2D2 $b{=}8$ & $6{\times}10^{18}$ & 29.0 & 34.4 & 57.5 & 61.5 & 29.2 & 36.0 & 26.2 & 16.7 & 36.3 \\
 & $10^{19}$ & 29.6 & 33.7 & 56.9 & 62.2 & 26.4 & 36.4 & 27.4 & 20.4 & 36.6 \\
 & $3{\times}10^{19}$ & 31.9 & 36.9 & 59.4 & 69.5 & 30.2 & 37.4 & 28.8 & 23.9 & 39.7 \\
 & $6{\times}10^{19}$ & 33.4 & 38.9 & 61.9 & 69.0 & 30.8 & 38.0 & 28.8 & 25.9 & 40.8 \\
 & $10^{20}$ & 36.0 & 39.6 & 63.3 & 71.1 & 32.0 & 38.8 & 28.2 & 32.1 & 42.6 \\
\addlinespace[1pt]
E2D2 $b{=}4$ & $6{\times}10^{18}$ & 28.9 & 34.9 & 57.1 & 60.2 & 28.2 & 36.2 & 24.6 & 18.1 & 36.0 \\
 & $10^{19}$ & 29.9 & 35.6 & 57.9 & 63.2 & 28.4 & 36.9 & 25.0 & 20.8 & 37.2 \\
 & $3{\times}10^{19}$ & 32.8 & 37.3 & 60.9 & 67.1 & 30.5 & 36.5 & 27.6 & 24.8 & 39.7 \\
 & $6{\times}10^{19}$ & 35.2 & 39.8 & 62.2 & 69.1 & 31.3 & 37.9 & 27.4 & 28.7 & 41.4 \\
 & $10^{20}$ & 37.1 & 41.9 & 64.0 & 69.5 & 30.9 & 38.1 & 27.0 & 29.6 & 42.3 \\
\bottomrule
\end{tabular}
\end{table}

\section{Assets}
\label{app:assets}
Table~\ref{tab:assets} lists the third-party code, data and benchmarks used in this study, with their licenses.

\begin{table}[htbp]
\centering
\scriptsize
\caption{Third-party assets used in this study and their licenses.}
\label{tab:assets}
\setlength{\tabcolsep}{4pt}
\resizebox{\linewidth}{!}{\begin{tabular}{llcll}
\toprule
asset & license & & asset & license \\
\midrule
\multicolumn{2}{l}{\emph{Code}} & & \multicolumn{2}{l}{\emph{Data}} \\
SMDM \citep{nie2025scaling} & Apache-2.0 & & SlimPajama-627B \citep{cerebras2023slimpajama} & Apache-2.0 \\
TinyLlama \citep{zhang2024tinyllama} / lit-gpt & Apache-2.0 & & \multicolumn{2}{l}{\emph{Benchmarks}} \\
nanoGPT \citep{karpathy2022nanogpt} & MIT & & HellaSwag \citep{zellers2019hellaswag} & MIT \\
GPT-NeoX \citep{black2022gpt} & Apache-2.0 & & ARC-easy \citep{clark2018think} & CC BY-SA 4.0 \\
FlashAttention 2.6.3 \citep{dao2024flashattention} & BSD-3-Clause & & PIQA \citep{bisk2020piqa} & AFL-3.0 \\
xFormers 0.0.28 \citep{lefaudeux2022xformers} & BSD-3-Clause & & SciQ \citep{welbl2017crowdsourcing} & CC BY-NC 3.0 \\
PyTorch 2.4.1 \citep{paszke2019pytorch} & BSD-3-Clause & & RACE \citep{lai2017race} & Apache-2.0 \\
Lightning 2.1.2 \citep{falcon2019lightning} & Apache-2.0 & & SIQA \citep{sap2019social} & CC BY 4.0 \\
Transformers 4.31.0 \citep{wolf2020transformers} & Apache-2.0 & & OpenBookQA \citep{mihaylov2018can} & Apache-2.0 \\
Datasets 3.0.1 \citep{lhoest2021datasets} & Apache-2.0 & & LAMBADA \citep{paperno2016lambada} & CC BY 4.0 \\
SentencePiece \citep{kudo2018sentencepiece} & Apache-2.0 & & WinoGrande \citep{sakaguchi2021winogrande} & Apache-2.0 \\
lm-evaluation-harness 0.4.4 \citep{gao2021framework} & MIT & & BoolQ \citep{clark2019boolq} & CC BY-SA 3.0 \\
 &  & & COPA \citep{roemmele2011choice} & BSD-2-Clause \\
\bottomrule
\end{tabular}}
\end{table}

\section{Per-cell results of the IsoFLOP sweeps}
\label{app:cells}
Table~\ref{tab:cells1} lists every cell of the ARM, MDM and BD3LM sweeps, with the size that minimizes each IsoFLOP curve in bold, and Tables~\ref{tab:cells3}--\ref{tab:cells6} every trained model of the E2D2 decoder-depth sweeps, with the best decoder depth of each (budget, size) group in bold.

\begin{table}[p]
\centering
\scriptsize\setlength{\tabcolsep}{3.2pt}
\caption{Per-cell results of the IsoFLOP sweeps for the ARM, the MDM and BD3LM. $N$ is the nominal model size in millions of non-embedding parameters, as named in Table~\ref{tab:runs}; entries are validation NLLs in nats per token (exact for the ARM; DUEL and ELBO for the diffusion variants), and ``---'' marks a size that was not trained for that variant. The last three ARM rows are the held-out budget of Section~\ref{sec:extrap}. Within each budget, the smallest value in each column, i.e.\ the minimum of each IsoFLOP curve, is in bold.}
\label{tab:cells1}
\begin{tabular}{lrrrrrrrrrrrr}
\toprule
 & & ARM & \multicolumn{2}{c}{MDM} & \multicolumn{2}{c}{BD3LM $b{=}32$} & \multicolumn{2}{c}{BD3LM $b{=}16$} & \multicolumn{2}{c}{BD3LM $b{=}8$} & \multicolumn{2}{c}{BD3LM $b{=}4$} \\
\cmidrule(lr){3-3}\cmidrule(lr){4-5}\cmidrule(lr){6-7}\cmidrule(lr){8-9}\cmidrule(lr){10-11}\cmidrule(lr){12-13}
$C$ & $N$ & NLL & DUEL & ELBO & DUEL & ELBO & DUEL & ELBO & DUEL & ELBO & DUEL & ELBO \\
\midrule
$6{\times}10^{18}$ & 19 & --- & --- & --- & 3.4143 & 3.5110 & 3.3783 & 3.4803 & 3.3263 & 3.4305 & --- & --- \\
 & 34 & --- & 3.3979 & 3.4465 & 3.3510 & 3.4422 & 3.3074 & 3.4069 & 3.2588 & 3.3601 & 3.1909 & 3.2756 \\
 & 48 & --- & 3.3491 & 3.4045 & 3.3057 & 3.4000 & 3.2640 & 3.3654 & 3.2125 & 3.3146 & 3.1442 & 3.2268 \\
 & 66 & 2.8782 & \textbf{3.3355} & \textbf{3.3806} & \textbf{3.2863} & \textbf{3.3809} & \textbf{3.2504} & 3.3511 & 3.1987 & 3.2975 & 3.1310 & 3.2150 \\
 & 85 & --- & --- & --- & 3.2943 & 3.3836 & 3.2611 & 3.3577 & 3.1953 & 3.2954 & 3.1331 & 3.2176 \\
 & 113 & 2.8507 & 3.3758 & 3.4006 & 3.2971 & 3.3883 & 3.2510 & \textbf{3.3505} & \textbf{3.1914} & \textbf{3.2895} & \textbf{3.1242} & \textbf{3.2081} \\
 & 142 & --- & --- & --- & 3.3115 & 3.4035 & 3.2722 & 3.3680 & 3.2130 & 3.3090 & 3.1347 & 3.2168 \\
 & 170 & \textbf{2.8365} & 3.4406 & 3.4447 & 3.3480 & 3.4346 & 3.3004 & 3.3931 & 3.2480 & 3.3422 & 3.1593 & 3.2393 \\
 & 231 & 2.8373 & 3.5058 & 3.4980 & 3.3709 & 3.4586 & 3.3369 & 3.4312 & 3.2711 & 3.3655 & 3.1918 & 3.2730 \\
 & 336 & 2.8558 & 3.6221 & 3.6161 & 3.4562 & 3.5405 & 3.4207 & 3.5132 & 3.3495 & 3.4449 & 3.2598 & 3.3437 \\
 & 472 & 2.8818 & --- & --- & --- & --- & --- & --- & --- & --- & --- & --- \\
 & 629 & 2.8991 & --- & --- & --- & --- & --- & --- & --- & --- & --- & --- \\
\addlinespace[2pt]
$10^{19}$ & 48 & --- & 3.2884 & 3.3358 & 3.2118 & 3.3067 & 3.1806 & 3.2793 & 3.1372 & 3.2371 & 3.0758 & 3.1581 \\
 & 66 & --- & 3.2556 & 3.2969 & 3.1903 & 3.2867 & 3.1536 & 3.2548 & 3.1053 & 3.2049 & 3.0426 & 3.1240 \\
 & 85 & 2.7910 & 3.2367 & 3.2701 & 3.1844 & 3.2755 & 3.1508 & 3.2476 & 3.0994 & 3.1950 & 3.0264 & 3.1066 \\
 & 113 & --- & --- & --- & \textbf{3.1728} & \textbf{3.2663} & \textbf{3.1323} & \textbf{3.2310} & \textbf{3.0750} & \textbf{3.1724} & 3.0121 & 3.0946 \\
 & 142 & 2.7544 & \textbf{3.2083} & \textbf{3.2588} & 3.1787 & 3.2708 & 3.1379 & 3.2349 & 3.0788 & 3.1751 & \textbf{3.0078} & \textbf{3.0859} \\
 & 170 & --- & --- & --- & 3.1988 & 3.2860 & 3.1595 & 3.2539 & 3.1031 & 3.1968 & 3.0219 & 3.0986 \\
 & 206 & --- & --- & --- & 3.2044 & 3.2962 & 3.1567 & 3.2536 & 3.1020 & 3.1972 & 3.0253 & 3.1045 \\
 & 231 & \textbf{2.7396} & 3.3178 & 3.3264 & --- & --- & --- & --- & --- & --- & --- & --- \\
 & 268 & --- & --- & --- & 3.2272 & 3.3165 & 3.1870 & 3.2812 & 3.1259 & 3.2186 & 3.0456 & 3.1233 \\
 & 336 & 2.7420 & 3.3741 & 3.3765 & 3.2735 & 3.3602 & 3.2233 & 3.3187 & 3.1585 & 3.2521 & 3.0818 & 3.1587 \\
 & 472 & 2.7577 & 3.3512 & 3.3951 & --- & --- & --- & --- & --- & --- & --- & --- \\
 & 666 & 2.7845 & --- & --- & --- & --- & --- & --- & --- & --- & --- & --- \\
 & 944 & 2.8030 & --- & --- & --- & --- & --- & --- & --- & --- & --- & --- \\
\addlinespace[2pt]
$3{\times}10^{19}$ & 66 & --- & 3.1810 & 3.1794 & --- & --- & --- & --- & --- & --- & --- & --- \\
 & 85 & --- & --- & --- & 3.0139 & 3.1064 & 2.9810 & 3.0795 & 2.9349 & 3.0304 & --- & --- \\
 & 113 & --- & --- & --- & 2.9794 & 3.0741 & 2.9440 & 3.0435 & 2.8955 & 2.9913 & --- & --- \\
 & 142 & 2.6185 & 3.0317 & 3.0796 & 2.9655 & 3.0577 & 2.9234 & 3.0198 & 2.8763 & 2.9709 & 2.8156 & 2.8916 \\
 & 170 & --- & --- & --- & 2.9693 & 3.0587 & 2.9333 & 3.0275 & 2.8848 & 2.9760 & 2.8203 & 2.8950 \\
 & 206 & --- & --- & --- & 2.9542 & 3.0467 & \textbf{2.9151} & 3.0108 & 2.8660 & 2.9594 & \textbf{2.7973} & \textbf{2.8708} \\
 & 231 & 2.5735 & 3.0083 & \textbf{3.0509} & --- & --- & --- & --- & --- & --- & --- & --- \\
 & 268 & --- & --- & --- & \textbf{2.9516} & \textbf{3.0426} & 2.9151 & \textbf{3.0098} & \textbf{2.8646} & \textbf{2.9561} & 2.7974 & 2.8712 \\
 & 336 & 2.5577 & 3.0269 & 3.0630 & 2.9635 & 3.0526 & 2.9240 & 3.0186 & 2.8716 & 2.9637 & 2.8053 & 2.8777 \\
 & 472 & \textbf{2.5496} & \textbf{2.9985} & 3.0546 & 2.9915 & 3.0820 & 2.9457 & 3.0401 & 2.8910 & 2.9819 & 2.8240 & 2.8957 \\
 & 666 & 2.5646 & 3.0424 & 3.0972 & 3.0372 & 3.1251 & 2.9956 & 3.0895 & 2.9326 & 3.0234 & 2.8565 & 2.9289 \\
 & 944 & 2.5510 & 3.1436 & 3.1891 & --- & --- & --- & --- & --- & --- & --- & --- \\
 & 1233 & 2.5734 & --- & --- & --- & --- & --- & --- & --- & --- & --- & --- \\
\addlinespace[2pt]
$6{\times}10^{19}$ & 113 & --- & 2.9785 & 3.0238 & --- & --- & --- & --- & --- & --- & --- & --- \\
 & 142 & --- & --- & --- & 2.8683 & 2.9627 & 2.8299 & 2.9279 & 2.7880 & 2.8801 & --- & --- \\
 & 170 & --- & --- & --- & 2.8673 & 2.9582 & 2.8332 & 2.9271 & 2.7887 & 2.8803 & --- & --- \\
 & 206 & 2.5119 & 2.8994 & 2.9542 & 2.8436 & 2.9350 & 2.8059 & 2.9016 & 2.7648 & 2.8552 & 2.7046 & 2.7763 \\
 & 268 & --- & --- & --- & 2.8330 & 2.9248 & 2.7982 & 2.8936 & 2.7475 & 2.8380 & 2.6911 & 2.7621 \\
 & 302 & 2.4781 & 2.8813 & 2.9301 & --- & --- & --- & --- & --- & --- & --- & --- \\
 & 336 & --- & --- & --- & \textbf{2.8262} & \textbf{2.9188} & \textbf{2.7890} & \textbf{2.8847} & \textbf{2.7408} & \textbf{2.8310} & 2.6794 & 2.7497 \\
 & 472 & \textbf{2.4557} & 2.8796 & 2.9259 & 2.8309 & 2.9243 & 2.7918 & 2.8875 & 2.7448 & 2.8338 & \textbf{2.6788} & \textbf{2.7480} \\
 & 666 & 2.4560 & \textbf{2.8702} & \textbf{2.9216} & 2.8525 & 2.9433 & 2.8117 & 2.9058 & 2.7664 & 2.8553 & 2.6977 & 2.7665 \\
 & 944 & 2.4637 & 2.9083 & 2.9622 & 2.9032 & 2.9912 & 2.8542 & 2.9474 & 2.7961 & 2.8862 & 2.7234 & 2.7926 \\
 & 1233 & 2.4582 & 2.9653 & 3.0094 & --- & --- & --- & --- & --- & --- & 2.7568 & 2.8280 \\
 & 1678 & 2.4680 & --- & --- & --- & --- & --- & --- & --- & --- & --- & --- \\
\addlinespace[2pt]
$10^{20}$ & 142 & --- & 2.9131 & 2.9637 & --- & --- & --- & --- & --- & --- & --- & --- \\
 & 206 & --- & --- & --- & 2.7775 & 2.8696 & 2.7412 & 2.8363 & 2.7037 & 2.7934 & --- & --- \\
 & 268 & 2.4479 & 2.8189 & 2.8725 & 2.7607 & 2.8532 & 2.7252 & 2.8199 & 2.6823 & 2.7719 & 2.6262 & 2.6961 \\
 & 336 & --- & --- & --- & 2.7448 & 2.8363 & 2.7120 & 2.8063 & 2.6668 & 2.7557 & 2.6112 & 2.6803 \\
 & 472 & 2.4025 & 2.7997 & 2.8529 & \textbf{2.7397} & \textbf{2.8323} & \textbf{2.7019} & \textbf{2.7963} & \textbf{2.6557} & \textbf{2.7438} & 2.5973 & 2.6647 \\
 & 666 & 2.3881 & 2.8013 & 2.8565 & 2.7413 & 2.8335 & 2.7030 & 2.7979 & 2.6570 & 2.7451 & \textbf{2.5943} & \textbf{2.6619} \\
 & 944 & \textbf{2.3843} & \textbf{2.7763} & \textbf{2.8401} & 2.7699 & 2.8604 & 2.7344 & 2.8268 & 2.6842 & 2.7719 & 2.6205 & 2.6879 \\
 & 1233 & 2.4000 & 2.8177 & 2.8738 & 2.8066 & 2.8964 & 2.7625 & 2.8564 & 2.7034 & 2.7910 & 2.6372 & 2.7051 \\
 & 1678 & 2.4031 & 2.8733 & 2.9289 & --- & --- & --- & --- & --- & --- & 2.6672 & 2.7348 \\
 & 2121 & 2.3933 & --- & --- & --- & --- & --- & --- & --- & --- & --- & --- \\
\addlinespace[2pt]
$3{\times}10^{20}$ & 1678 & \textbf{2.2468} & --- & --- & --- & --- & --- & --- & --- & --- & --- & --- \\
 & 2121 & 2.2610 & --- & --- & --- & --- & --- & --- & --- & --- & --- & --- \\
 & 2718 & 2.2627 & --- & --- & --- & --- & --- & --- & --- & --- & --- & --- \\
\bottomrule
\end{tabular}
\end{table}

\begin{table}[p]
\centering
\scriptsize\setlength{\tabcolsep}{2.4pt}
\caption{Decoder-depth sweep of E2D2 $b{=}32$ (part 1 of 2): every trained model, one row per decoder depth. $N$ is the nominal model size in millions of non-embedding parameters, as named in Table~\ref{tab:runs}; DUEL and ELBO are validation NLLs in nats per token. $L_{\rm dec}/L$ is the decoder depth over the total depth. The best $L_{\rm dec}$ of each (budget, size) group is in bold. The rows at $3{\times}10^{20}$ FLOPs are the held-out budget of Section~\ref{sec:extrap}.}
\label{tab:cells3}
\begin{minipage}[t]{0.245\linewidth}\centering
\begin{tabular}{rrrr}
\toprule
$N$ & $L_{\rm dec}/L$ & DUEL & ELBO \\
\midrule
\multicolumn{4}{l}{$C={6{\times}10^{18}}$} \\
34 & 2/8 & 3.5005 & 3.5703 \\
34 & 3/8 & 3.3989 & 3.4660 \\
34 & 4/8 & 3.3500 & 3.4175 \\
34 & 5/8 & 3.3105 & 3.3794 \\
34 & 6/8 & \textbf{3.2909} & 3.3593 \\
34 & 7/8 & 3.2967 & 3.3690 \\
\addlinespace[1.5pt]
48 & 2/9 & 3.4762 & 3.5476 \\
48 & 3/9 & 3.3744 & 3.4421 \\
48 & 4/9 & 3.3219 & 3.3888 \\
48 & 5/9 & 3.2818 & 3.3501 \\
48 & 6/9 & 3.2562 & 3.3238 \\
48 & 7/9 & \textbf{3.2534} & 3.3226 \\
48 & 8/9 & 3.2626 & 3.3335 \\
\addlinespace[1.5pt]
66 & 3/10 & 3.3548 & 3.4228 \\
66 & 4/10 & 3.3043 & 3.3710 \\
66 & 5/10 & 3.2647 & 3.3304 \\
66 & 6/10 & 3.2478 & 3.3148 \\
66 & 7/10 & \textbf{3.2308} & 3.2984 \\
66 & 8/10 & 3.2336 & 3.3023 \\
66 & 9/10 & 3.2377 & 3.3105 \\
\addlinespace[1.5pt]
85 & 5/13 & 3.2978 & 3.3643 \\
85 & 6/13 & 3.2663 & 3.3340 \\
85 & 7/13 & 3.2488 & 3.3172 \\
85 & 8/13 & 3.2326 & 3.2989 \\
85 & 9/13 & 3.2232 & 3.2913 \\
85 & 10/13 & \textbf{3.2186} & 3.2869 \\
85 & 11/13 & 3.2316 & 3.2988 \\
85 & 12/13 & 3.2378 & 3.3084 \\
\addlinespace[1.5pt]
113 & 3/12 & 3.3673 & 3.4342 \\
113 & 4/12 & 3.3188 & 3.3869 \\
113 & 5/12 & 3.2841 & 3.3501 \\
113 & 6/12 & 3.2529 & 3.3204 \\
113 & 7/12 & 3.2488 & 3.3149 \\
113 & 8/12 & \textbf{3.2292} & 3.2974 \\
113 & 9/12 & 3.2338 & 3.3021 \\
113 & 10/12 & 3.2352 & 3.3019 \\
113 & 11/12 & 3.2371 & 3.3085 \\
\addlinespace[1.5pt]
142 & 5/15 & 3.3172 & 3.3820 \\
142 & 6/15 & 3.2957 & 3.3613 \\
142 & 7/15 & 3.2731 & 3.3385 \\
142 & 8/15 & 3.2565 & 3.3223 \\
142 & 9/15 & 3.2431 & 3.3104 \\
142 & 10/15 & 3.2388 & 3.3050 \\
142 & 11/15 & \textbf{3.2352} & 3.3007 \\
142 & 12/15 & 3.2443 & 3.3089 \\
142 & 13/15 & 3.2443 & 3.3105 \\
142 & 14/15 & 3.2553 & 3.3250 \\
\addlinespace[1.5pt]
170 & 8/18 & 3.2922 & 3.3582 \\
170 & 9/18 & 3.2824 & 3.3490 \\
170 & 10/18 & 3.2764 & 3.3404 \\
170 & 11/18 & 3.2595 & 3.3261 \\
170 & 12/18 & 3.2564 & 3.3208 \\
170 & 13/18 & 3.2547 & 3.3190 \\
170 & 14/18 & \textbf{3.2430} & 3.3068 \\
170 & 15/18 & 3.2489 & 3.3154 \\
170 & 16/18 & 3.2590 & 3.3253 \\
170 & 17/18 & 3.2766 & 3.3464 \\
\addlinespace[1.5pt]
231 & 7/18 & 3.3294 & 3.3938 \\
231 & 8/18 & 3.3162 & 3.3822 \\
231 & 9/18 & 3.3153 & 3.3790 \\
231 & 10/18 & 3.2933 & 3.3573 \\
231 & 11/18 & 3.2940 & 3.3569 \\
231 & 12/18 & 3.2929 & 3.3579 \\
231 & 13/18 & 3.2927 & 3.3571 \\
\bottomrule
\end{tabular}
\end{minipage}\hfill
\begin{minipage}[t]{0.245\linewidth}\centering
\begin{tabular}{rrrr}
\toprule
$N$ & $L_{\rm dec}/L$ & DUEL & ELBO \\
\midrule
\multicolumn{4}{l}{$C={6{\times}10^{18}}$ (cont.)} \\
231 & 14/18 & \textbf{3.2890} & 3.3543 \\
231 & 15/18 & 3.2972 & 3.3613 \\
231 & 16/18 & 3.2975 & 3.3618 \\
231 & 17/18 & 3.3110 & 3.3804 \\
\addlinespace[2pt]
\multicolumn{4}{l}{$C={10^{19}}$} \\
66 & 3/10 & 3.2856 & 3.3547 \\
66 & 4/10 & 3.2247 & 3.2932 \\
66 & 5/10 & 3.1850 & 3.2533 \\
66 & 6/10 & 3.1629 & 3.2330 \\
66 & 7/10 & 3.1454 & 3.2150 \\
66 & 8/10 & \textbf{3.1444} & 3.2144 \\
66 & 9/10 & 3.1517 & 3.2256 \\
\addlinespace[1.5pt]
85 & 6/13 & 3.1741 & 3.2438 \\
85 & 7/13 & 3.1486 & 3.2199 \\
85 & 8/13 & 3.1349 & 3.2035 \\
85 & 9/13 & 3.1242 & 3.1940 \\
85 & 10/13 & \textbf{3.1189} & 3.1893 \\
85 & 11/13 & 3.1270 & 3.1962 \\
85 & 12/13 & 3.1374 & 3.2090 \\
\addlinespace[1.5pt]
113 & 5/12 & 3.1728 & 3.2425 \\
113 & 6/12 & 3.1466 & 3.2149 \\
113 & 7/12 & 3.1251 & 3.1934 \\
113 & 8/12 & 3.1140 & 3.1831 \\
113 & 9/12 & \textbf{3.1134} & 3.1822 \\
113 & 10/12 & 3.1154 & 3.1861 \\
113 & 11/12 & 3.1282 & 3.1998 \\
\addlinespace[1.5pt]
142 & 8/15 & 3.1363 & 3.2061 \\
142 & 9/15 & 3.1216 & 3.1917 \\
142 & 10/15 & 3.1141 & 3.1830 \\
142 & 11/15 & \textbf{3.1073} & 3.1768 \\
142 & 12/15 & 3.1146 & 3.1825 \\
142 & 13/15 & 3.1180 & 3.1874 \\
142 & 14/15 & 3.1290 & 3.2015 \\
\addlinespace[1.5pt]
170 & 11/18 & 3.1289 & 3.1980 \\
170 & 12/18 & 3.1241 & 3.1910 \\
170 & 13/18 & 3.1155 & 3.1845 \\
170 & 14/18 & \textbf{3.1119} & 3.1797 \\
170 & 15/18 & 3.1189 & 3.1871 \\
170 & 16/18 & 3.1256 & 3.1951 \\
170 & 17/18 & 3.1404 & 3.2114 \\
\addlinespace[1.5pt]
231 & 10/18 & 3.1504 & 3.2163 \\
231 & 11/18 & 3.1435 & 3.2112 \\
231 & 12/18 & 3.1489 & 3.2176 \\
231 & 13/18 & 3.1431 & 3.2116 \\
231 & 14/18 & \textbf{3.1362} & 3.2052 \\
231 & 15/18 & 3.1427 & 3.2099 \\
231 & 16/18 & 3.1483 & 3.2149 \\
\addlinespace[1.5pt]
336 & 10/20 & 3.2085 & 3.2751 \\
336 & 11/20 & 3.1962 & 3.2628 \\
336 & 12/20 & 3.1947 & 3.2617 \\
336 & 13/20 & 3.1985 & 3.2643 \\
336 & 14/20 & 3.1951 & 3.2607 \\
336 & 15/20 & \textbf{3.1844} & 3.2501 \\
336 & 16/20 & 3.1973 & 3.2634 \\
\addlinespace[2pt]
\multicolumn{4}{l}{$C={3{\times}10^{19}}$} \\
142 & 8/15 & 2.9587 & 3.0279 \\
142 & 9/15 & 2.9414 & 3.0128 \\
142 & 10/15 & 2.9278 & 2.9984 \\
142 & 11/15 & 2.9222 & 2.9945 \\
142 & 12/15 & \textbf{2.9218} & 2.9919 \\
142 & 13/15 & 2.9299 & 3.0020 \\
142 & 14/15 & 2.9461 & 3.0189 \\
\addlinespace[1.5pt]
170 & 11/18 & 2.9314 & 3.0031 \\
170 & 12/18 & 2.9213 & 2.9923 \\
170 & 13/18 & 2.9097 & 2.9812 \\
170 & 14/18 & \textbf{2.9088} & 2.9813 \\
\bottomrule
\end{tabular}
\end{minipage}\hfill
\begin{minipage}[t]{0.245\linewidth}\centering
\begin{tabular}{rrrr}
\toprule
$N$ & $L_{\rm dec}/L$ & DUEL & ELBO \\
\midrule
\multicolumn{4}{l}{$C={3{\times}10^{19}}$ (cont.)} \\
170 & 15/18 & 2.9091 & 2.9807 \\
170 & 16/18 & 2.9223 & 2.9940 \\
170 & 17/18 & 2.9396 & 3.0125 \\
\addlinespace[1.5pt]
206 & 9/16 & 2.9238 & 2.9951 \\
206 & 10/16 & 2.9113 & 2.9829 \\
206 & 11/16 & 2.9056 & 2.9774 \\
206 & 12/16 & \textbf{2.8997} & 2.9718 \\
206 & 13/16 & 2.9010 & 2.9727 \\
206 & 14/16 & 2.9120 & 2.9847 \\
206 & 15/16 & 2.9258 & 2.9982 \\
\addlinespace[1.5pt]
268 & 9/16 & 2.9229 & 2.9937 \\
268 & 10/16 & 2.9077 & 2.9796 \\
268 & 11/16 & 2.8987 & 2.9699 \\
268 & 12/16 & \textbf{2.8975} & 2.9698 \\
268 & 13/16 & 2.8986 & 2.9710 \\
268 & 14/16 & 2.9083 & 2.9806 \\
268 & 15/16 & 2.9191 & 2.9928 \\
\addlinespace[1.5pt]
336 & 12/20 & 2.9157 & 2.9878 \\
336 & 13/20 & 2.9147 & 2.9848 \\
336 & 14/20 & 2.9062 & 2.9774 \\
336 & 15/20 & \textbf{2.9020} & 2.9724 \\
336 & 16/20 & 2.9032 & 2.9744 \\
336 & 17/20 & 2.9135 & 2.9844 \\
336 & 18/20 & 2.9173 & 2.9889 \\
\addlinespace[1.5pt]
472 & 9/18 & 2.9606 & 3.0314 \\
472 & 10/18 & 2.9472 & 3.0176 \\
472 & 11/18 & 2.9440 & 3.0151 \\
472 & 12/18 & 2.9451 & 3.0174 \\
472 & 13/18 & 2.9342 & 3.0042 \\
472 & 14/18 & 2.9368 & 3.0092 \\
472 & 15/18 & 2.9334 & 3.0046 \\
472 & 16/18 & \textbf{2.9275} & 2.9994 \\
472 & 17/18 & 2.9350 & 3.0079 \\
\addlinespace[1.5pt]
666 & 11/21 & 3.0089 & 3.0786 \\
666 & 12/21 & 3.0039 & 3.0736 \\
666 & 13/21 & 2.9899 & 3.0614 \\
666 & 14/21 & 2.9878 & 3.0579 \\
666 & 15/21 & 2.9981 & 3.0679 \\
666 & 16/21 & 2.9752 & 3.0460 \\
666 & 17/21 & 2.9738 & 3.0436 \\
666 & 18/21 & 2.9772 & 3.0466 \\
666 & 19/21 & 2.9740 & 3.0450 \\
666 & 20/21 & \textbf{2.9704} & 3.0415 \\
\addlinespace[2pt]
\multicolumn{4}{l}{$C={6{\times}10^{19}}$} \\
170 & 11/18 & 2.8514 & 2.9231 \\
170 & 12/18 & 2.8404 & 2.9111 \\
170 & 13/18 & 2.8317 & 2.9029 \\
170 & 14/18 & \textbf{2.8241} & 2.8961 \\
170 & 15/18 & 2.8250 & 2.8953 \\
170 & 16/18 & 2.8357 & 2.9063 \\
170 & 17/18 & 2.8572 & 2.9290 \\
\addlinespace[1.5pt]
206 & 9/16 & 2.8356 & 2.9080 \\
206 & 10/16 & 2.8207 & 2.8928 \\
206 & 11/16 & 2.8120 & 2.8838 \\
206 & 12/16 & 2.8056 & 2.8779 \\
206 & 13/16 & \textbf{2.8047} & 2.8773 \\
206 & 14/16 & 2.8164 & 2.8890 \\
206 & 15/16 & 2.8389 & 2.9126 \\
\addlinespace[1.5pt]
268 & 9/16 & 2.8149 & 2.8885 \\
268 & 10/16 & 2.8035 & 2.8764 \\
268 & 11/16 & 2.7944 & 2.8671 \\
268 & 12/16 & 2.7890 & 2.8612 \\
268 & 13/16 & \textbf{2.7888} & 2.8613 \\
268 & 14/16 & 2.8011 & 2.8729 \\
268 & 15/16 & 2.8151 & 2.8882 \\
\bottomrule
\end{tabular}
\end{minipage}\hfill
\begin{minipage}[t]{0.245\linewidth}\centering
\begin{tabular}{rrrr}
\toprule
$N$ & $L_{\rm dec}/L$ & DUEL & ELBO \\
\midrule
\multicolumn{4}{l}{$C={6{\times}10^{19}}$ (cont.)} \\
336 & 13/20 & 2.7882 & 2.8609 \\
336 & 14/20 & 2.7859 & 2.8587 \\
336 & 15/20 & \textbf{2.7795} & 2.8527 \\
336 & 16/20 & 2.7807 & 2.8530 \\
336 & 17/20 & 2.7846 & 2.8568 \\
336 & 18/20 & 2.7967 & 2.8690 \\
336 & 19/20 & 2.8101 & 2.8832 \\
\addlinespace[1.5pt]
472 & 10/18 & 2.8049 & 2.8772 \\
472 & 11/18 & 2.7985 & 2.8719 \\
472 & 12/18 & 2.7896 & 2.8639 \\
472 & 13/18 & \textbf{2.7886} & 2.8609 \\
472 & 14/18 & 2.7897 & 2.8630 \\
472 & 15/18 & 2.7910 & 2.8647 \\
472 & 16/18 & 2.7940 & 2.8677 \\
\addlinespace[1.5pt]
666 & 12/21 & 2.8256 & 2.8977 \\
666 & 13/21 & 2.8139 & 2.8865 \\
666 & 14/21 & 2.8086 & 2.8815 \\
666 & 15/21 & 2.8166 & 2.8901 \\
666 & 16/21 & 2.8110 & 2.8839 \\
666 & 17/21 & 2.8086 & 2.8807 \\
666 & 18/21 & \textbf{2.8076} & 2.8803 \\
666 & 19/21 & 2.8104 & 2.8829 \\
666 & 20/21 & 2.8182 & 2.8915 \\
\addlinespace[1.5pt]
944 & 15/25 & 2.8693 & 2.9421 \\
944 & 16/25 & 2.8732 & 2.9439 \\
944 & 17/25 & 2.8623 & 2.9334 \\
944 & 18/25 & 2.8570 & 2.9289 \\
944 & 19/25 & \textbf{2.8427} & 2.9156 \\
944 & 20/25 & 2.8593 & 2.9326 \\
944 & 21/25 & 2.8499 & 2.9225 \\
\addlinespace[2pt]
\multicolumn{4}{l}{$C={10^{20}}$} \\
206 & 9/16 & 2.7872 & 2.8599 \\
206 & 10/16 & 2.7715 & 2.8436 \\
206 & 11/16 & 2.7625 & 2.8349 \\
206 & 12/16 & \textbf{2.7539} & 2.8269 \\
206 & 13/16 & 2.7548 & 2.8273 \\
206 & 14/16 & 2.7644 & 2.8369 \\
206 & 15/16 & 2.7856 & 2.8591 \\
\addlinespace[1.5pt]
268 & 9/16 & 2.7588 & 2.8322 \\
268 & 10/16 & 2.7442 & 2.8158 \\
268 & 11/16 & 2.7331 & 2.8063 \\
268 & 12/16 & 2.7292 & 2.8018 \\
268 & 13/16 & \textbf{2.7275} & 2.7995 \\
268 & 14/16 & 2.7393 & 2.8122 \\
268 & 15/16 & 2.7582 & 2.8322 \\
\addlinespace[1.5pt]
336 & 13/20 & 2.7279 & 2.8002 \\
336 & 14/20 & 2.7169 & 2.7898 \\
336 & 15/20 & 2.7133 & 2.7856 \\
336 & 16/20 & \textbf{2.7095} & 2.7810 \\
336 & 17/20 & 2.7141 & 2.7866 \\
336 & 18/20 & 2.7276 & 2.7993 \\
336 & 19/20 & 2.7495 & 2.8237 \\
\addlinespace[1.5pt]
472 & 11/18 & 2.7163 & 2.7891 \\
472 & 12/18 & 2.7088 & 2.7827 \\
472 & 13/18 & 2.7058 & 2.7787 \\
472 & 14/18 & \textbf{2.7032} & 2.7761 \\
472 & 15/18 & 2.7049 & 2.7789 \\
472 & 16/18 & 2.7140 & 2.7871 \\
472 & 17/18 & 2.7284 & 2.8024 \\
\addlinespace[1.5pt]
666 & 13/21 & 2.7180 & 2.7915 \\
666 & 14/21 & 2.7132 & 2.7874 \\
666 & 15/21 & 2.7127 & 2.7861 \\
666 & 16/21 & \textbf{2.7081} & 2.7805 \\
666 & 17/21 & 2.7091 & 2.7829 \\
666 & 18/21 & 2.7144 & 2.7868 \\
\bottomrule
\end{tabular}
\end{minipage}
\end{table}

\begin{table}[p]
\centering
\scriptsize\setlength{\tabcolsep}{2.4pt}
\caption{Decoder-depth sweep of E2D2 $b{=}32$ (part 2 of 2): every trained model, one row per decoder depth.}
\label{tab:cells3_2}
\begin{minipage}[t]{0.245\linewidth}\centering
\begin{tabular}{rrrr}
\toprule
$N$ & $L_{\rm dec}/L$ & DUEL & ELBO \\
\midrule
\multicolumn{4}{l}{$C={10^{20}}$ (cont.)} \\
666 & 19/21 & 2.7170 & 2.7904 \\
\addlinespace[1.5pt]
944 & 16/25 & 2.7424 & 2.8160 \\
944 & 17/25 & 2.7359 & 2.8078 \\
944 & 18/25 & 2.7294 & 2.8019 \\
944 & 19/25 & 2.7291 & 2.8018 \\
944 & 20/25 & \textbf{2.7277} & 2.8001 \\
944 & 21/25 & 2.7293 & 2.8018 \\
944 & 22/25 & 2.7353 & 2.8089 \\
\addlinespace[1.5pt]
1233 & 14/24 & 2.7788 & 2.8515 \\
1233 & 15/24 & 2.7746 & 2.8472 \\
1233 & 16/24 & 2.7747 & 2.8472 \\
1233 & 17/24 & 2.7796 & 2.8531 \\
1233 & 18/24 & 2.7737 & 2.8456 \\
1233 & 19/24 & 2.7682 & 2.8402 \\
1233 & 20/24 & 2.7663 & 2.8392 \\
1233 & 21/24 & 2.7632 & 2.8364 \\
1233 & 22/24 & \textbf{2.7619} & 2.8352 \\
1233 & 23/24 & 2.7655 & 2.8388 \\
\addlinespace[2pt]
\multicolumn{4}{l}{$C={3{\times}10^{20}}$} \\
831 & 16/22 & 2.5473 & 2.6205 \\
831 & 17/22 & 2.5462 & 2.6192 \\
831 & 18/22 & \textbf{2.5461} & 2.6193 \\
\addlinespace[1.5pt]
944 & 18/25 & 2.5465 & 2.6188 \\
944 & 19/25 & \textbf{2.5436} & 2.6166 \\
944 & 20/25 & 2.5453 & 2.6182 \\
\addlinespace[1.5pt]
1233 & 17/24 & 2.5548 & 2.6277 \\
1233 & 18/24 & 2.5527 & 2.6258 \\
1233 & 19/24 & 2.5543 & 2.6266 \\
1233 & 20/24 & \textbf{2.5523} & 2.6261 \\
1233 & 21/24 & 2.5562 & 2.6301 \\
\bottomrule
\end{tabular}
\end{minipage}
\end{table}

\begin{table}[p]
\centering
\scriptsize\setlength{\tabcolsep}{2.4pt}
\caption{Decoder-depth sweep of E2D2 $b{=}16$ (part 1 of 2): every trained model, one row per decoder depth. $N$ is the nominal model size in millions of non-embedding parameters, as named in Table~\ref{tab:runs}; DUEL and ELBO are validation NLLs in nats per token. $L_{\rm dec}/L$ is the decoder depth over the total depth. The best $L_{\rm dec}$ of each (budget, size) group is in bold. The rows at $3{\times}10^{20}$ FLOPs are the held-out budget of Section~\ref{sec:extrap}.}
\label{tab:cells4}
\begin{minipage}[t]{0.245\linewidth}\centering
\begin{tabular}{rrrr}
\toprule
$N$ & $L_{\rm dec}/L$ & DUEL & ELBO \\
\midrule
\multicolumn{4}{l}{$C={6{\times}10^{18}}$} \\
34 & 2/8 & 3.4250 & 3.4991 \\
34 & 3/8 & 3.3382 & 3.4073 \\
34 & 4/8 & 3.2968 & 3.3658 \\
34 & 5/8 & 3.2738 & 3.3441 \\
34 & 6/8 & \textbf{3.2583} & 3.3294 \\
34 & 7/8 & 3.2755 & 3.3502 \\
\addlinespace[1.5pt]
48 & 2/9 & 3.4005 & 3.4749 \\
48 & 3/9 & 3.3136 & 3.3830 \\
48 & 4/9 & 3.2617 & 3.3317 \\
48 & 5/9 & 3.2391 & 3.3097 \\
48 & 6/9 & \textbf{3.2199} & 3.2916 \\
48 & 7/9 & 3.2239 & 3.2950 \\
48 & 8/9 & 3.2407 & 3.3146 \\
\addlinespace[1.5pt]
66 & 2/10 & 3.3867 & 3.4558 \\
66 & 3/10 & 3.2972 & 3.3679 \\
66 & 4/10 & 3.2492 & 3.3184 \\
66 & 5/10 & 3.2228 & 3.2934 \\
66 & 6/10 & 3.2085 & 3.2788 \\
66 & 7/10 & \textbf{3.1981} & 3.2685 \\
66 & 8/10 & 3.2098 & 3.2800 \\
66 & 9/10 & 3.2212 & 3.2965 \\
\addlinespace[1.5pt]
85 & 4/13 & 3.2726 & 3.3412 \\
85 & 5/13 & 3.2452 & 3.3136 \\
85 & 6/13 & 3.2211 & 3.2902 \\
85 & 7/13 & 3.2026 & 3.2727 \\
85 & 8/13 & \textbf{3.1886} & 3.2575 \\
85 & 9/13 & 3.1891 & 3.2597 \\
85 & 10/13 & 3.1921 & 3.2617 \\
85 & 11/13 & 3.2077 & 3.2783 \\
85 & 12/13 & 3.2217 & 3.2959 \\
\addlinespace[1.5pt]
113 & 3/12 & 3.3007 & 3.3700 \\
113 & 4/12 & 3.2572 & 3.3283 \\
113 & 5/12 & 3.2272 & 3.2967 \\
113 & 6/12 & 3.2065 & 3.2759 \\
113 & 7/12 & 3.1931 & 3.2626 \\
113 & 8/12 & \textbf{3.1857} & 3.2549 \\
113 & 9/12 & 3.1949 & 3.2657 \\
113 & 10/12 & 3.2068 & 3.2778 \\
113 & 11/12 & 3.2170 & 3.2914 \\
\addlinespace[1.5pt]
142 & 6/15 & 3.2401 & 3.3078 \\
142 & 7/15 & 3.2216 & 3.2906 \\
142 & 8/15 & 3.2104 & 3.2791 \\
142 & 9/15 & 3.2002 & 3.2702 \\
142 & 10/15 & \textbf{3.1976} & 3.2686 \\
142 & 11/15 & 3.1982 & 3.2673 \\
142 & 12/15 & 3.2105 & 3.2819 \\
142 & 13/15 & 3.2207 & 3.2915 \\
142 & 14/15 & 3.2385 & 3.3136 \\
\addlinespace[1.5pt]
170 & 8/18 & 3.2417 & 3.3105 \\
170 & 9/18 & 3.2294 & 3.2986 \\
170 & 10/18 & 3.2244 & 3.2947 \\
170 & 11/18 & 3.2176 & 3.2878 \\
170 & 12/18 & 3.2104 & 3.2787 \\
170 & 13/18 & \textbf{3.2039} & 3.2739 \\
170 & 14/18 & 3.2087 & 3.2780 \\
170 & 15/18 & 3.2196 & 3.2914 \\
170 & 16/18 & 3.2356 & 3.3060 \\
170 & 17/18 & 3.2613 & 3.3362 \\
\addlinespace[1.5pt]
231 & 8/18 & 3.2696 & 3.3381 \\
231 & 9/18 & 3.2542 & 3.3229 \\
231 & 10/18 & \textbf{3.2462} & 3.3155 \\
231 & 11/18 & 3.2469 & 3.3167 \\
231 & 12/18 & 3.2532 & 3.3213 \\
231 & 13/18 & 3.2479 & 3.3174 \\
\bottomrule
\end{tabular}
\end{minipage}\hfill
\begin{minipage}[t]{0.245\linewidth}\centering
\begin{tabular}{rrrr}
\toprule
$N$ & $L_{\rm dec}/L$ & DUEL & ELBO \\
\midrule
\multicolumn{4}{l}{$C={6{\times}10^{18}}$ (cont.)} \\
231 & 14/18 & 3.2507 & 3.3210 \\
231 & 15/18 & 3.2537 & 3.3226 \\
231 & 16/18 & 3.2710 & 3.3424 \\
231 & 17/18 & 3.2883 & 3.3614 \\
\addlinespace[2pt]
\multicolumn{4}{l}{$C={10^{19}}$} \\
66 & 2/10 & 3.3111 & 3.3815 \\
66 & 3/10 & 3.2211 & 3.2923 \\
66 & 4/10 & 3.1700 & 3.2394 \\
66 & 5/10 & 3.1397 & 3.2091 \\
66 & 6/10 & 3.1244 & 3.1951 \\
66 & 7/10 & \textbf{3.1138} & 3.1849 \\
66 & 8/10 & 3.1156 & 3.1880 \\
66 & 9/10 & 3.1351 & 3.2110 \\
\addlinespace[1.5pt]
85 & 4/13 & 3.1851 & 3.2561 \\
85 & 5/13 & 3.1503 & 3.2202 \\
85 & 6/13 & 3.1275 & 3.1989 \\
85 & 7/13 & 3.1055 & 3.1782 \\
85 & 8/13 & 3.0957 & 3.1664 \\
85 & 9/13 & 3.0924 & 3.1648 \\
85 & 10/13 & \textbf{3.0923} & 3.1647 \\
85 & 11/13 & 3.0991 & 3.1707 \\
\addlinespace[1.5pt]
113 & 3/12 & 3.2018 & 3.2715 \\
113 & 4/12 & 3.1557 & 3.2266 \\
113 & 5/12 & 3.1234 & 3.1945 \\
113 & 6/12 & 3.1065 & 3.1757 \\
113 & 7/12 & 3.0861 & 3.1570 \\
113 & 8/12 & \textbf{3.0753} & 3.1475 \\
113 & 9/12 & 3.0781 & 3.1499 \\
113 & 10/12 & 3.0941 & 3.1679 \\
113 & 11/12 & 3.1057 & 3.1820 \\
\addlinespace[1.5pt]
142 & 6/15 & 3.1211 & 3.1919 \\
142 & 7/15 & 3.1070 & 3.1779 \\
142 & 8/15 & 3.0910 & 3.1630 \\
142 & 9/15 & 3.0824 & 3.1528 \\
142 & 10/15 & 3.0790 & 3.1496 \\
142 & 11/15 & \textbf{3.0739} & 3.1461 \\
142 & 12/15 & 3.0783 & 3.1502 \\
142 & 13/15 & 3.0993 & 3.1723 \\
142 & 14/15 & 3.1125 & 3.1881 \\
\addlinespace[1.5pt]
170 & 8/18 & 3.1151 & 3.1865 \\
170 & 9/18 & 3.1040 & 3.1738 \\
170 & 10/18 & 3.0968 & 3.1679 \\
170 & 11/18 & 3.0885 & 3.1608 \\
170 & 12/18 & 3.0829 & 3.1545 \\
170 & 13/18 & 3.0791 & 3.1495 \\
170 & 14/18 & \textbf{3.0777} & 3.1490 \\
170 & 15/18 & 3.0858 & 3.1583 \\
170 & 16/18 & 3.1018 & 3.1741 \\
170 & 17/18 & 3.1267 & 3.2009 \\
\addlinespace[1.5pt]
231 & 6/18 & 3.1535 & 3.2230 \\
231 & 7/18 & 3.1344 & 3.2038 \\
231 & 8/18 & 3.1261 & 3.1975 \\
231 & 9/18 & 3.1167 & 3.1863 \\
231 & 10/18 & 3.1238 & 3.1949 \\
231 & 11/18 & \textbf{3.1016} & 3.1733 \\
231 & 12/18 & 3.1036 & 3.1749 \\
231 & 13/18 & 3.1055 & 3.1774 \\
231 & 14/18 & 3.1087 & 3.1805 \\
231 & 15/18 & 3.1115 & 3.1833 \\
231 & 16/18 & 3.1245 & 3.1969 \\
\addlinespace[1.5pt]
336 & 8/20 & 3.2038 & 3.2751 \\
336 & 9/20 & 3.1698 & 3.2396 \\
336 & 10/20 & 3.1660 & 3.2364 \\
336 & 11/20 & 3.1764 & 3.2456 \\
336 & 12/20 & 3.1707 & 3.2424 \\
\bottomrule
\end{tabular}
\end{minipage}\hfill
\begin{minipage}[t]{0.245\linewidth}\centering
\begin{tabular}{rrrr}
\toprule
$N$ & $L_{\rm dec}/L$ & DUEL & ELBO \\
\midrule
\multicolumn{4}{l}{$C={10^{19}}$ (cont.)} \\
336 & 13/20 & \textbf{3.1495} & 3.2191 \\
336 & 14/20 & 3.1626 & 3.2336 \\
336 & 15/20 & 3.1575 & 3.2273 \\
336 & 16/20 & 3.1611 & 3.2298 \\
336 & 17/20 & 3.1530 & 3.2224 \\
336 & 18/20 & 3.1663 & 3.2377 \\
\addlinespace[2pt]
\multicolumn{4}{l}{$C={3{\times}10^{19}}$} \\
142 & 6/15 & 2.9565 & 3.0278 \\
142 & 7/15 & 2.9341 & 3.0062 \\
142 & 8/15 & 2.9180 & 2.9904 \\
142 & 9/15 & 2.9029 & 2.9746 \\
142 & 10/15 & 2.8941 & 2.9674 \\
142 & 11/15 & 2.8885 & 2.9623 \\
142 & 12/15 & \textbf{2.8884} & 2.9622 \\
142 & 13/15 & 2.9006 & 2.9754 \\
142 & 14/15 & 2.9290 & 3.0048 \\
\addlinespace[1.5pt]
170 & 8/18 & 2.9330 & 3.0043 \\
170 & 9/18 & 2.9134 & 2.9856 \\
170 & 10/18 & 2.9040 & 2.9768 \\
170 & 11/18 & 2.8883 & 2.9611 \\
170 & 12/18 & 2.8867 & 2.9590 \\
170 & 13/18 & 2.8776 & 2.9512 \\
170 & 14/18 & \textbf{2.8768} & 2.9506 \\
170 & 15/18 & 2.8825 & 2.9559 \\
170 & 16/18 & 2.8966 & 2.9718 \\
170 & 17/18 & 2.9208 & 2.9975 \\
\addlinespace[1.5pt]
206 & 7/16 & 2.9094 & 2.9820 \\
206 & 8/16 & 2.8945 & 2.9677 \\
206 & 9/16 & 2.8827 & 2.9558 \\
206 & 10/16 & 2.8715 & 2.9454 \\
206 & 11/16 & 2.8674 & 2.9412 \\
206 & 12/16 & \textbf{2.8645} & 2.9384 \\
206 & 13/16 & 2.8704 & 2.9444 \\
206 & 14/16 & 2.8857 & 2.9607 \\
206 & 15/16 & 2.9085 & 2.9850 \\
\addlinespace[1.5pt]
268 & 7/16 & 2.9018 & 2.9755 \\
268 & 8/16 & 2.8880 & 2.9610 \\
268 & 9/16 & 2.8785 & 2.9525 \\
268 & 10/16 & 2.8684 & 2.9425 \\
268 & 11/16 & 2.8636 & 2.9379 \\
268 & 12/16 & \textbf{2.8609} & 2.9362 \\
268 & 13/16 & 2.8694 & 2.9436 \\
268 & 14/16 & 2.8816 & 2.9567 \\
268 & 15/16 & 2.9009 & 2.9774 \\
\addlinespace[1.5pt]
336 & 9/20 & 2.9007 & 2.9746 \\
336 & 10/20 & 2.8880 & 2.9624 \\
336 & 11/20 & 2.8849 & 2.9588 \\
336 & 12/20 & 2.8794 & 2.9528 \\
336 & 13/20 & 2.8714 & 2.9457 \\
336 & 14/20 & 2.8682 & 2.9424 \\
336 & 15/20 & \textbf{2.8642} & 2.9382 \\
336 & 16/20 & 2.8649 & 2.9396 \\
336 & 17/20 & 2.8812 & 2.9559 \\
336 & 18/20 & 2.8946 & 2.9697 \\
336 & 19/20 & 2.9115 & 2.9881 \\
\addlinespace[1.5pt]
472 & 8/18 & 2.9181 & 2.9915 \\
472 & 9/18 & 2.9081 & 2.9829 \\
472 & 10/18 & 2.9095 & 2.9828 \\
472 & 11/18 & 2.9035 & 2.9773 \\
472 & 12/18 & 2.8931 & 2.9684 \\
472 & 13/18 & 2.8959 & 2.9699 \\
472 & 14/18 & \textbf{2.8923} & 2.9662 \\
472 & 15/18 & 2.8961 & 2.9701 \\
472 & 16/18 & 2.9023 & 2.9772 \\
472 & 17/18 & 2.9165 & 2.9923 \\
\bottomrule
\end{tabular}
\end{minipage}\hfill
\begin{minipage}[t]{0.245\linewidth}\centering
\begin{tabular}{rrrr}
\toprule
$N$ & $L_{\rm dec}/L$ & DUEL & ELBO \\
\midrule
\multicolumn{4}{l}{$C={3{\times}10^{19}}$ (cont.)} \\
666 & 10/21 & 2.9782 & 3.0506 \\
666 & 11/21 & 2.9580 & 3.0307 \\
666 & 12/21 & 2.9602 & 3.0322 \\
666 & 13/21 & 2.9491 & 3.0217 \\
666 & 14/21 & 2.9481 & 3.0205 \\
666 & 15/21 & 2.9588 & 3.0310 \\
666 & 16/21 & 2.9437 & 3.0175 \\
666 & 17/21 & \textbf{2.9373} & 3.0097 \\
666 & 18/21 & 2.9458 & 3.0184 \\
666 & 19/21 & 2.9460 & 3.0193 \\
666 & 20/21 & 2.9511 & 3.0255 \\
\addlinespace[2pt]
\multicolumn{4}{l}{$C={6{\times}10^{19}}$} \\
170 & 7/18 & 2.8697 & 2.9412 \\
170 & 8/18 & 2.8483 & 2.9210 \\
170 & 9/18 & 2.8349 & 2.9086 \\
170 & 10/18 & 2.8211 & 2.8938 \\
170 & 11/18 & 2.8102 & 2.8828 \\
170 & 12/18 & 2.8011 & 2.8745 \\
170 & 13/18 & 2.7938 & 2.8681 \\
170 & 14/18 & \textbf{2.7905} & 2.8646 \\
170 & 15/18 & 2.7941 & 2.8689 \\
170 & 16/18 & 2.8064 & 2.8813 \\
170 & 17/18 & 2.8363 & 2.9135 \\
\addlinespace[1.5pt]
206 & 7/16 & 2.8312 & 2.9036 \\
206 & 8/16 & 2.8105 & 2.8830 \\
206 & 9/16 & 2.7945 & 2.8673 \\
206 & 10/16 & 2.7829 & 2.8565 \\
206 & 11/16 & 2.7781 & 2.8517 \\
206 & 12/16 & \textbf{2.7724} & 2.8463 \\
206 & 13/16 & 2.7754 & 2.8493 \\
206 & 14/16 & 2.7891 & 2.8648 \\
206 & 15/16 & 2.8152 & 2.8917 \\
\addlinespace[1.5pt]
268 & 6/16 & 2.8319 & 2.9042 \\
268 & 7/16 & 2.8050 & 2.8790 \\
268 & 8/16 & 2.7890 & 2.8623 \\
268 & 9/16 & 2.7732 & 2.8471 \\
268 & 10/16 & 2.7640 & 2.8378 \\
268 & 11/16 & 2.7574 & 2.8315 \\
268 & 12/16 & \textbf{2.7546} & 2.8283 \\
268 & 13/16 & 2.7584 & 2.8329 \\
268 & 14/16 & 2.7743 & 2.8496 \\
268 & 15/16 & 2.7980 & 2.8746 \\
\addlinespace[1.5pt]
336 & 10/20 & 2.7782 & 2.8520 \\
336 & 11/20 & 2.7665 & 2.8401 \\
336 & 12/20 & 2.7625 & 2.8362 \\
336 & 13/20 & 2.7531 & 2.8276 \\
336 & 14/20 & 2.7484 & 2.8216 \\
336 & 15/20 & \textbf{2.7454} & 2.8200 \\
336 & 16/20 & 2.7476 & 2.8215 \\
336 & 17/20 & 2.7525 & 2.8261 \\
336 & 18/20 & 2.7723 & 2.8472 \\
336 & 19/20 & 2.7957 & 2.8720 \\
\addlinespace[1.5pt]
472 & 8/18 & 2.7802 & 2.8549 \\
472 & 9/18 & 2.7722 & 2.8477 \\
472 & 10/18 & 2.7620 & 2.8357 \\
472 & 11/18 & 2.7540 & 2.8289 \\
472 & 12/18 & 2.7515 & 2.8256 \\
472 & 13/18 & \textbf{2.7461} & 2.8218 \\
472 & 14/18 & 2.7479 & 2.8238 \\
472 & 15/18 & 2.7556 & 2.8310 \\
472 & 16/18 & 2.7655 & 2.8410 \\
472 & 17/18 & 2.7836 & 2.8612 \\
\addlinespace[1.5pt]
666 & 10/21 & 2.7918 & 2.8656 \\
666 & 11/21 & 2.7846 & 2.8589 \\
666 & 12/21 & 2.7810 & 2.8556 \\
\bottomrule
\end{tabular}
\end{minipage}
\end{table}

\begin{table}[p]
\centering
\scriptsize\setlength{\tabcolsep}{2.4pt}
\caption{Decoder-depth sweep of E2D2 $b{=}16$ (part 2 of 2): every trained model, one row per decoder depth.}
\label{tab:cells4_2}
\begin{minipage}[t]{0.245\linewidth}\centering
\begin{tabular}{rrrr}
\toprule
$N$ & $L_{\rm dec}/L$ & DUEL & ELBO \\
\midrule
\multicolumn{4}{l}{$C={6{\times}10^{19}}$ (cont.)} \\
666 & 13/21 & 2.7724 & 2.8466 \\
666 & 14/21 & 2.7700 & 2.8447 \\
666 & 15/21 & \textbf{2.7675} & 2.8432 \\
666 & 16/21 & 2.7751 & 2.8503 \\
666 & 17/21 & 2.7710 & 2.8465 \\
666 & 18/21 & 2.7764 & 2.8519 \\
666 & 19/21 & 2.7817 & 2.8579 \\
666 & 20/21 & 2.8005 & 2.8772 \\
\addlinespace[1.5pt]
944 & 14/25 & 2.8181 & 2.8915 \\
944 & 15/25 & 2.8161 & 2.8914 \\
944 & 16/25 & 2.8265 & 2.9005 \\
944 & 17/25 & \textbf{2.8104} & 2.8834 \\
944 & 18/25 & 2.8149 & 2.8889 \\
944 & 19/25 & 2.8134 & 2.8879 \\
944 & 20/25 & 2.8107 & 2.8862 \\
944 & 21/25 & 2.8131 & 2.8866 \\
944 & 22/25 & 2.8142 & 2.8881 \\
944 & 23/25 & 2.8184 & 2.8938 \\
944 & 24/25 & 2.8282 & 2.9038 \\
\addlinespace[2pt]
\multicolumn{4}{l}{$C={10^{20}}$} \\
206 & 7/16 & 2.7813 & 2.8535 \\
206 & 8/16 & 2.7601 & 2.8330 \\
206 & 9/16 & 2.7437 & 2.8164 \\
206 & 10/16 & 2.7314 & 2.8049 \\
206 & 11/16 & 2.7242 & 2.7980 \\
206 & 12/16 & \textbf{2.7196} & 2.7942 \\
206 & 13/16 & 2.7239 & 2.7984 \\
206 & 14/16 & 2.7369 & 2.8125 \\
206 & 15/16 & 2.7635 & 2.8412 \\
\addlinespace[1.5pt]
268 & 7/16 & 2.7530 & 2.8257 \\
268 & 8/16 & 2.7323 & 2.8054 \\
268 & 9/16 & 2.7170 & 2.7899 \\
268 & 10/16 & 2.7059 & 2.7791 \\
268 & 11/16 & 2.6967 & 2.7707 \\
268 & 12/16 & 2.6967 & 2.7710 \\
268 & 13/16 & \textbf{2.6963} & 2.7715 \\
268 & 14/16 & 2.7103 & 2.7861 \\
268 & 15/16 & 2.7374 & 2.8139 \\
\addlinespace[1.5pt]
336 & 11/20 & 2.7018 & 2.7748 \\
336 & 12/20 & 2.6964 & 2.7695 \\
336 & 13/20 & 2.6897 & 2.7632 \\
336 & 14/20 & 2.6812 & 2.7551 \\
336 & 15/20 & \textbf{2.6786} & 2.7521 \\
336 & 16/20 & 2.6788 & 2.7528 \\
336 & 17/20 & 2.6840 & 2.7582 \\
336 & 18/20 & 2.7013 & 2.7771 \\
336 & 19/20 & 2.7217 & 2.8012 \\
\addlinespace[1.5pt]
472 & 8/18 & 2.7056 & 2.7798 \\
472 & 9/18 & 2.6917 & 2.7663 \\
472 & 10/18 & 2.6821 & 2.7563 \\
472 & 11/18 & 2.6755 & 2.7492 \\
472 & 12/18 & 2.6705 & 2.7455 \\
472 & 13/18 & \textbf{2.6678} & 2.7422 \\
472 & 14/18 & 2.6689 & 2.7430 \\
472 & 15/18 & 2.6709 & 2.7465 \\
472 & 16/18 & 2.6895 & 2.7653 \\
472 & 17/18 & 2.7068 & 2.7839 \\
\addlinespace[1.5pt]
666 & 11/21 & 2.6899 & 2.7644 \\
666 & 12/21 & 2.6825 & 2.7573 \\
666 & 13/21 & 2.6794 & 2.7534 \\
666 & 14/21 & 2.6712 & 2.7462 \\
666 & 15/21 & \textbf{2.6698} & 2.7448 \\
666 & 16/21 & 2.6702 & 2.7444 \\
666 & 17/21 & 2.6700 & 2.7444 \\
666 & 18/21 & 2.6800 & 2.7551 \\
\bottomrule
\end{tabular}
\end{minipage}\hfill
\begin{minipage}[t]{0.245\linewidth}\centering
\begin{tabular}{rrrr}
\toprule
$N$ & $L_{\rm dec}/L$ & DUEL & ELBO \\
\midrule
\multicolumn{4}{l}{$C={10^{20}}$ (cont.)} \\
666 & 19/21 & 2.6891 & 2.7651 \\
666 & 20/21 & 2.7072 & 2.7864 \\
\addlinespace[1.5pt]
944 & 14/25 & 2.7049 & 2.7792 \\
944 & 15/25 & 2.6998 & 2.7757 \\
944 & 16/25 & 2.6951 & 2.7698 \\
944 & 17/25 & 2.6902 & 2.7635 \\
944 & 18/25 & 2.6903 & 2.7651 \\
944 & 19/25 & 2.6918 & 2.7668 \\
944 & 20/25 & \textbf{2.6901} & 2.7645 \\
944 & 21/25 & 2.6979 & 2.7731 \\
944 & 22/25 & 2.6994 & 2.7749 \\
944 & 23/25 & 2.7136 & 2.7899 \\
944 & 24/25 & 2.7256 & 2.8027 \\
\addlinespace[1.5pt]
1233 & 14/24 & 2.7341 & 2.8092 \\
1233 & 15/24 & 2.7299 & 2.8049 \\
1233 & 16/24 & 2.7303 & 2.8054 \\
1233 & 17/24 & 2.7345 & 2.8089 \\
1233 & 19/24 & 2.7302 & 2.8051 \\
1233 & 20/24 & \textbf{2.7258} & 2.8012 \\
1233 & 21/24 & 2.7311 & 2.8071 \\
1233 & 22/24 & 2.7358 & 2.8117 \\
1233 & 23/24 & 2.7430 & 2.8203 \\
\addlinespace[2pt]
\multicolumn{4}{l}{$C={3{\times}10^{20}}$} \\
831 & 15/22 & 2.5136 & 2.5862 \\
831 & 16/22 & \textbf{2.5095} & 2.5824 \\
831 & 17/22 & 2.5097 & 2.5834 \\
\addlinespace[1.5pt]
944 & 16/25 & 2.5127 & 2.5858 \\
944 & 17/25 & 2.5123 & 2.5852 \\
944 & 18/25 & 2.5106 & 2.5841 \\
944 & 19/25 & \textbf{2.5084} & 2.5816 \\
944 & 20/25 & 2.5112 & 2.5838 \\
\addlinespace[1.5pt]
1233 & 16/24 & 2.5199 & 2.5930 \\
1233 & 17/24 & 2.5165 & 2.5897 \\
1233 & 18/24 & \textbf{2.5129} & 2.5866 \\
\bottomrule
\end{tabular}
\end{minipage}
\end{table}

\begin{table}[p]
\centering
\scriptsize\setlength{\tabcolsep}{2.4pt}
\caption{Decoder-depth sweep of E2D2 $b{=}8$ (part 1 of 2): every trained model, one row per decoder depth. $N$ is the nominal model size in millions of non-embedding parameters, as named in Table~\ref{tab:runs}; DUEL and ELBO are validation NLLs in nats per token. $L_{\rm dec}/L$ is the decoder depth over the total depth. The best $L_{\rm dec}$ of each (budget, size) group is in bold.}
\label{tab:cells5}
\begin{minipage}[t]{0.245\linewidth}\centering
\begin{tabular}{rrrr}
\toprule
$N$ & $L_{\rm dec}/L$ & DUEL & ELBO \\
\midrule
\multicolumn{4}{l}{$C={6{\times}10^{18}}$} \\
34 & 2/8 & 3.3391 & 3.4108 \\
34 & 3/8 & 3.2682 & 3.3356 \\
34 & 4/8 & 3.2355 & 3.3051 \\
34 & 5/8 & 3.2167 & 3.2879 \\
34 & 6/8 & \textbf{3.2092} & 3.2822 \\
34 & 7/8 & 3.2386 & 3.3156 \\
\addlinespace[1.5pt]
48 & 2/9 & 3.3077 & 3.3766 \\
48 & 3/9 & 3.2392 & 3.3078 \\
48 & 4/9 & 3.1997 & 3.2707 \\
48 & 5/9 & 3.1757 & 3.2461 \\
48 & 6/9 & \textbf{3.1698} & 3.2405 \\
48 & 7/9 & 3.1739 & 3.2472 \\
48 & 8/9 & 3.2048 & 3.2808 \\
\addlinespace[1.5pt]
66 & 2/10 & 3.2886 & 3.3568 \\
66 & 3/10 & 3.2121 & 3.2812 \\
66 & 4/10 & 3.1750 & 3.2441 \\
66 & 5/10 & 3.1579 & 3.2267 \\
66 & 6/10 & 3.1460 & 3.2171 \\
66 & 7/10 & \textbf{3.1412} & 3.2141 \\
66 & 8/10 & 3.1604 & 3.2346 \\
66 & 9/10 & 3.1849 & 3.2623 \\
\addlinespace[1.5pt]
85 & 4/13 & 3.1946 & 3.2630 \\
85 & 5/13 & 3.1726 & 3.2414 \\
85 & 6/13 & 3.1515 & 3.2232 \\
85 & 7/13 & 3.1352 & 3.2068 \\
85 & 8/13 & \textbf{3.1308} & 3.2015 \\
85 & 9/13 & 3.1310 & 3.2033 \\
85 & 10/13 & 3.1368 & 3.2112 \\
85 & 11/13 & 3.1533 & 3.2274 \\
85 & 12/13 & 3.1849 & 3.2621 \\
\addlinespace[1.5pt]
113 & 4/12 & 3.1760 & 3.2457 \\
113 & 5/12 & 3.1503 & 3.2205 \\
113 & 6/12 & 3.1394 & 3.2100 \\
113 & 7/12 & 3.1355 & 3.2064 \\
113 & 8/12 & \textbf{3.1259} & 3.1987 \\
113 & 9/12 & 3.1424 & 3.2152 \\
113 & 10/12 & 3.1649 & 3.2403 \\
113 & 11/12 & 3.1820 & 3.2602 \\
\addlinespace[1.5pt]
170 & 8/18 & 3.1673 & 3.2381 \\
170 & 9/18 & 3.1587 & 3.2310 \\
170 & 10/18 & 3.1561 & 3.2286 \\
170 & 11/18 & 3.1505 & 3.2231 \\
170 & 12/18 & 3.1467 & 3.2190 \\
170 & 13/18 & \textbf{3.1457} & 3.2187 \\
170 & 14/18 & 3.1504 & 3.2236 \\
170 & 15/18 & 3.1657 & 3.2402 \\
170 & 16/18 & 3.1883 & 3.2649 \\
\addlinespace[1.5pt]
231 & 7/18 & 3.2318 & 3.3039 \\
231 & 8/18 & 3.1947 & 3.2657 \\
231 & 9/18 & 3.1851 & 3.2572 \\
231 & 10/18 & 3.2050 & 3.2760 \\
231 & 11/18 & 3.1760 & 3.2481 \\
231 & 12/18 & 3.2084 & 3.2811 \\
231 & 13/18 & \textbf{3.1670} & 3.2395 \\
231 & 14/18 & 3.1914 & 3.2634 \\
231 & 15/18 & 3.1962 & 3.2710 \\
\addlinespace[2pt]
\multicolumn{4}{l}{$C={10^{19}}$} \\
66 & 3/10 & 3.1454 & 3.2138 \\
66 & 4/10 & 3.0988 & 3.1682 \\
66 & 5/10 & 3.0779 & 3.1472 \\
66 & 6/10 & 3.0635 & 3.1340 \\
66 & 7/10 & \textbf{3.0599} & 3.1317 \\
66 & 8/10 & 3.0652 & 3.1388 \\
66 & 9/10 & 3.0973 & 3.1735 \\
\addlinespace[1.5pt]
85 & 5/13 & 3.0832 & 3.1518 \\
\bottomrule
\end{tabular}
\end{minipage}\hfill
\begin{minipage}[t]{0.245\linewidth}\centering
\begin{tabular}{rrrr}
\toprule
$N$ & $L_{\rm dec}/L$ & DUEL & ELBO \\
\midrule
\multicolumn{4}{l}{$C={10^{19}}$ (cont.)} \\
85 & 6/13 & 3.0641 & 3.1339 \\
85 & 7/13 & 3.0438 & 3.1147 \\
85 & 8/13 & 3.0353 & 3.1070 \\
85 & 9/13 & \textbf{3.0327} & 3.1051 \\
85 & 10/13 & 3.0409 & 3.1141 \\
85 & 11/13 & 3.0553 & 3.1286 \\
85 & 12/13 & 3.0869 & 3.1637 \\
\addlinespace[1.5pt]
113 & 4/12 & 3.0764 & 3.1453 \\
113 & 5/12 & 3.0560 & 3.1259 \\
113 & 6/12 & 3.0334 & 3.1041 \\
113 & 7/12 & 3.0233 & 3.0947 \\
113 & 8/12 & \textbf{3.0175} & 3.0900 \\
113 & 9/12 & 3.0266 & 3.1011 \\
113 & 10/12 & 3.0477 & 3.1232 \\
113 & 11/12 & 3.0709 & 3.1476 \\
\addlinespace[1.5pt]
142 & 6/15 & 3.0496 & 3.1206 \\
142 & 7/15 & 3.0376 & 3.1081 \\
142 & 8/15 & 3.0256 & 3.0972 \\
142 & 9/15 & 3.0230 & 3.0955 \\
142 & 10/15 & 3.0159 & 3.0883 \\
142 & 11/15 & \textbf{3.0156} & 3.0892 \\
142 & 12/15 & 3.0244 & 3.0975 \\
142 & 13/15 & 3.0530 & 3.1292 \\
142 & 14/15 & 3.0762 & 3.1545 \\
\addlinespace[1.5pt]
170 & 9/18 & 3.0391 & 3.1105 \\
170 & 10/18 & 3.0349 & 3.1074 \\
170 & 11/18 & 3.0246 & 3.0965 \\
170 & 12/18 & 3.0239 & 3.0970 \\
170 & 13/18 & \textbf{3.0192} & 3.0934 \\
170 & 14/18 & 3.0231 & 3.0968 \\
170 & 15/18 & 3.0341 & 3.1097 \\
170 & 16/18 & 3.0595 & 3.1353 \\
170 & 17/18 & 3.0878 & 3.1659 \\
\addlinespace[1.5pt]
231 & 8/18 & 3.0534 & 3.1251 \\
231 & 9/18 & 3.0469 & 3.1189 \\
231 & 10/18 & 3.0599 & 3.1326 \\
231 & 11/18 & 3.0374 & 3.1105 \\
231 & 12/18 & 3.0564 & 3.1292 \\
231 & 13/18 & \textbf{3.0323} & 3.1063 \\
231 & 14/18 & 3.0554 & 3.1295 \\
231 & 15/18 & 3.0544 & 3.1291 \\
231 & 16/18 & 3.0803 & 3.1553 \\
\addlinespace[1.5pt]
336 & 8/20 & 3.1034 & 3.1759 \\
336 & 9/20 & 3.1241 & 3.1958 \\
336 & 10/20 & 3.1047 & 3.1780 \\
336 & 11/20 & 3.1223 & 3.1948 \\
336 & 12/20 & 3.1077 & 3.1805 \\
336 & 13/20 & 3.1074 & 3.1810 \\
336 & 14/20 & 3.1072 & 3.1809 \\
336 & 15/20 & 3.1021 & 3.1754 \\
336 & 16/20 & \textbf{3.0884} & 3.1624 \\
\addlinespace[2pt]
\multicolumn{4}{l}{$C={3{\times}10^{19}}$} \\
142 & 8/15 & 2.8637 & 2.9332 \\
142 & 9/15 & 2.8489 & 2.9200 \\
142 & 10/15 & 2.8438 & 2.9155 \\
142 & 11/15 & \textbf{2.8403} & 2.9125 \\
142 & 12/15 & 2.8448 & 2.9179 \\
142 & 13/15 & 2.8587 & 2.9340 \\
142 & 14/15 & 2.8905 & 2.9664 \\
\addlinespace[1.5pt]
206 & 8/16 & 2.8349 & 2.9046 \\
206 & 9/16 & 2.8247 & 2.8958 \\
206 & 10/16 & 2.8155 & 2.8867 \\
206 & 11/16 & \textbf{2.8150} & 2.8880 \\
206 & 12/16 & 2.8161 & 2.8882 \\
206 & 13/16 & 2.8233 & 2.8969 \\
\bottomrule
\end{tabular}
\end{minipage}\hfill
\begin{minipage}[t]{0.245\linewidth}\centering
\begin{tabular}{rrrr}
\toprule
$N$ & $L_{\rm dec}/L$ & DUEL & ELBO \\
\midrule
\multicolumn{4}{l}{$C={3{\times}10^{19}}$ (cont.)} \\
206 & 14/16 & 2.8432 & 2.9188 \\
206 & 15/16 & 2.8737 & 2.9502 \\
\addlinespace[1.5pt]
268 & 8/16 & 2.8263 & 2.8980 \\
268 & 9/16 & 2.8153 & 2.8880 \\
268 & 10/16 & 2.8107 & 2.8831 \\
268 & 11/16 & \textbf{2.8074} & 2.8800 \\
268 & 12/16 & 2.8076 & 2.8812 \\
268 & 13/16 & 2.8168 & 2.8911 \\
268 & 14/16 & 2.8390 & 2.9148 \\
268 & 15/16 & 2.8663 & 2.9437 \\
\addlinespace[1.5pt]
336 & 10/20 & 2.8274 & 2.8999 \\
336 & 11/20 & 2.8230 & 2.8957 \\
336 & 12/20 & 2.8162 & 2.8900 \\
336 & 14/20 & 2.8119 & 2.8847 \\
336 & 15/20 & \textbf{2.8088} & 2.8823 \\
336 & 16/20 & 2.8138 & 2.8881 \\
336 & 17/20 & 2.8354 & 2.9105 \\
336 & 18/20 & 2.8474 & 2.9229 \\
\addlinespace[1.5pt]
472 & 8/18 & 2.8496 & 2.9215 \\
472 & 9/18 & 2.8450 & 2.9175 \\
472 & 10/18 & 2.8488 & 2.9204 \\
472 & 11/18 & 2.8430 & 2.9152 \\
472 & 12/18 & 2.8424 & 2.9155 \\
472 & 13/18 & \textbf{2.8368} & 2.9100 \\
472 & 14/18 & 2.8398 & 2.9137 \\
472 & 15/18 & 2.8455 & 2.9193 \\
472 & 16/18 & 2.8568 & 2.9323 \\
\addlinespace[1.5pt]
666 & 11/21 & 2.8856 & 2.9580 \\
666 & 12/21 & 2.8856 & 2.9578 \\
666 & 13/21 & 2.8854 & 2.9581 \\
666 & 14/21 & 2.8898 & 2.9636 \\
666 & 15/21 & 2.8914 & 2.9639 \\
666 & 16/21 & \textbf{2.8771} & 2.9501 \\
666 & 17/21 & 2.8802 & 2.9533 \\
666 & 18/21 & 2.8856 & 2.9595 \\
666 & 19/21 & 2.8983 & 2.9728 \\
\addlinespace[2pt]
\multicolumn{4}{l}{$C={6{\times}10^{19}}$} \\
206 & 8/16 & 2.7552 & 2.8238 \\
206 & 9/16 & 2.7430 & 2.8130 \\
206 & 10/16 & 2.7341 & 2.8043 \\
206 & 11/16 & 2.7287 & 2.7993 \\
206 & 12/16 & \textbf{2.7271} & 2.7978 \\
206 & 13/16 & 2.7305 & 2.8030 \\
206 & 14/16 & 2.7488 & 2.8223 \\
206 & 15/16 & 2.7773 & 2.8535 \\
\addlinespace[1.5pt]
268 & 8/16 & 2.7310 & 2.8010 \\
268 & 9/16 & 2.7200 & 2.7905 \\
268 & 10/16 & 2.7131 & 2.7838 \\
268 & 11/16 & \textbf{2.7076} & 2.7785 \\
268 & 12/16 & 2.7096 & 2.7817 \\
268 & 13/16 & 2.7145 & 2.7867 \\
268 & 14/16 & 2.7316 & 2.8056 \\
268 & 15/16 & 2.7632 & 2.8391 \\
\addlinespace[1.5pt]
336 & 10/20 & 2.7247 & 2.7952 \\
336 & 11/20 & 2.7152 & 2.7864 \\
336 & 12/20 & 2.7047 & 2.7760 \\
336 & 13/20 & 2.7038 & 2.7756 \\
336 & 14/20 & 2.7007 & 2.7721 \\
336 & 15/20 & \textbf{2.6986} & 2.7710 \\
336 & 16/20 & 2.7011 & 2.7736 \\
336 & 17/20 & 2.7102 & 2.7844 \\
336 & 18/20 & 2.7297 & 2.8034 \\
\addlinespace[1.5pt]
472 & 8/18 & 2.7219 & 2.7924 \\
472 & 9/18 & 2.7141 & 2.7850 \\
472 & 10/18 & 2.7066 & 2.7771 \\
\bottomrule
\end{tabular}
\end{minipage}\hfill
\begin{minipage}[t]{0.245\linewidth}\centering
\begin{tabular}{rrrr}
\toprule
$N$ & $L_{\rm dec}/L$ & DUEL & ELBO \\
\midrule
\multicolumn{4}{l}{$C={6{\times}10^{19}}$ (cont.)} \\
472 & 11/18 & 2.6989 & 2.7701 \\
472 & 12/18 & 2.6987 & 2.7704 \\
472 & 13/18 & 2.6969 & 2.7691 \\
472 & 14/18 & \textbf{2.6968} & 2.7694 \\
472 & 15/18 & 2.7079 & 2.7810 \\
472 & 16/18 & 2.7273 & 2.8019 \\
\addlinespace[1.5pt]
666 & 10/21 & 2.7268 & 2.7985 \\
666 & 11/21 & 2.7237 & 2.7955 \\
666 & 12/21 & 2.7173 & 2.7893 \\
666 & 13/21 & 2.7128 & 2.7848 \\
666 & 14/21 & 2.7133 & 2.7860 \\
666 & 15/21 & \textbf{2.7093} & 2.7812 \\
666 & 16/21 & 2.7159 & 2.7885 \\
666 & 17/21 & 2.7181 & 2.7915 \\
666 & 18/21 & 2.7265 & 2.8007 \\
\addlinespace[1.5pt]
944 & 14/25 & 2.7562 & 2.8282 \\
944 & 15/25 & 2.7547 & 2.8281 \\
944 & 16/25 & 2.7626 & 2.8352 \\
944 & 17/25 & 2.7568 & 2.8289 \\
944 & 18/25 & 2.7596 & 2.8330 \\
944 & 19/25 & \textbf{2.7524} & 2.8256 \\
944 & 20/25 & 2.7576 & 2.8302 \\
944 & 21/25 & 2.7526 & 2.8254 \\
944 & 22/25 & 2.7660 & 2.8395 \\
\addlinespace[2pt]
\multicolumn{4}{l}{$C={10^{20}}$} \\
268 & 8/16 & 2.6807 & 2.7504 \\
268 & 9/16 & 2.6647 & 2.7343 \\
268 & 10/16 & 2.6562 & 2.7263 \\
268 & 11/16 & \textbf{2.6501} & 2.7208 \\
268 & 12/16 & 2.6507 & 2.7220 \\
268 & 13/16 & 2.6535 & 2.7258 \\
268 & 14/16 & 2.6693 & 2.7427 \\
268 & 15/16 & 2.6986 & 2.7742 \\
\addlinespace[1.5pt]
336 & 11/20 & 2.6546 & 2.7243 \\
336 & 12/20 & 2.6476 & 2.7173 \\
336 & 13/20 & 2.6383 & 2.7088 \\
336 & 14/20 & 2.6344 & 2.7046 \\
336 & 15/20 & \textbf{2.6314} & 2.7019 \\
336 & 16/20 & 2.6364 & 2.7077 \\
336 & 17/20 & 2.6425 & 2.7146 \\
336 & 18/20 & 2.6615 & 2.7347 \\
336 & 19/20 & 2.6919 & 2.7687 \\
\addlinespace[1.5pt]
472 & 8/18 & 2.6443 & 2.7125 \\
472 & 9/18 & 2.6355 & 2.7043 \\
472 & 10/18 & 2.6285 & 2.6968 \\
472 & 11/18 & 2.6225 & 2.6924 \\
472 & 12/18 & \textbf{2.6168} & 2.6868 \\
472 & 13/18 & 2.6173 & 2.6874 \\
472 & 14/18 & 2.6215 & 2.6924 \\
472 & 15/18 & 2.6283 & 2.6999 \\
\addlinespace[1.5pt]
666 & 10/21 & 2.6385 & 2.7078 \\
666 & 11/21 & 2.6330 & 2.7033 \\
666 & 12/21 & 2.6277 & 2.6985 \\
666 & 13/21 & 2.6231 & 2.6938 \\
666 & 14/21 & 2.6219 & 2.6927 \\
666 & 15/21 & 2.6207 & 2.6913 \\
666 & 16/21 & \textbf{2.6181} & 2.6896 \\
666 & 17/21 & 2.6222 & 2.6933 \\
666 & 18/21 & 2.6310 & 2.7030 \\
\addlinespace[1.5pt]
944 & 14/25 & 2.6424 & 2.7135 \\
944 & 15/25 & 2.6393 & 2.7104 \\
944 & 16/25 & 2.6385 & 2.7104 \\
944 & 17/25 & 2.6362 & 2.7074 \\
944 & 18/25 & \textbf{2.6357} & 2.7074 \\
944 & 19/25 & 2.6357 & 2.7082 \\
\bottomrule
\end{tabular}
\end{minipage}
\end{table}

\begin{table}[p]
\centering
\scriptsize\setlength{\tabcolsep}{2.4pt}
\caption{Decoder-depth sweep of E2D2 $b{=}8$ (part 2 of 2): every trained model, one row per decoder depth.}
\label{tab:cells5_2}
\begin{minipage}[t]{0.245\linewidth}\centering
\begin{tabular}{rrrr}
\toprule
$N$ & $L_{\rm dec}/L$ & DUEL & ELBO \\
\midrule
\multicolumn{4}{l}{$C={10^{20}}$ (cont.)} \\
944 & 20/25 & 2.6426 & 2.7151 \\
944 & 21/25 & 2.6398 & 2.7117 \\
944 & 22/25 & 2.6566 & 2.7302 \\
\addlinespace[1.5pt]
1233 & 12/24 & 2.6782 & 2.7507 \\
1233 & 13/24 & 2.6743 & 2.7464 \\
1233 & 14/24 & 2.6712 & 2.7427 \\
1233 & 15/24 & 2.6751 & 2.7469 \\
1233 & 16/24 & \textbf{2.6657} & 2.7384 \\
1233 & 17/24 & 2.6696 & 2.7422 \\
1233 & 18/24 & 2.6692 & 2.7415 \\
1233 & 19/24 & 2.6730 & 2.7449 \\
1233 & 20/24 & 2.6758 & 2.7491 \\
\bottomrule
\end{tabular}
\end{minipage}
\end{table}

\begin{table}[p]
\centering
\scriptsize\setlength{\tabcolsep}{2.4pt}
\caption{Decoder-depth sweep of E2D2 $b{=}4$ (part 1 of 3): every trained model, one row per decoder depth. $N$ is the nominal model size in millions of non-embedding parameters, as named in Table~\ref{tab:runs}; DUEL and ELBO are validation NLLs in nats per token. $L_{\rm dec}/L$ is the decoder depth over the total depth. The best $L_{\rm dec}$ of each (budget, size) group is in bold.}
\label{tab:cells6}
\begin{minipage}[t]{0.245\linewidth}\centering
\begin{tabular}{rrrr}
\toprule
$N$ & $L_{\rm dec}/L$ & DUEL & ELBO \\
\midrule
\multicolumn{4}{l}{$C={6{\times}10^{18}}$} \\
34 & 2/8 & 3.2262 & 3.2826 \\
34 & 3/8 & 3.1761 & 3.2335 \\
34 & 4/8 & 3.1530 & 3.2133 \\
34 & 5/8 & 3.1417 & 3.2042 \\
34 & 6/8 & \textbf{3.1368} & 3.2019 \\
34 & 7/8 & 3.1803 & 3.2489 \\
\addlinespace[1.5pt]
48 & 2/9 & 3.1894 & 3.2444 \\
48 & 3/9 & 3.1356 & 3.1932 \\
48 & 4/9 & 3.1095 & 3.1675 \\
48 & 5/9 & \textbf{3.0923} & 3.1527 \\
48 & 6/9 & 3.0941 & 3.1554 \\
48 & 7/9 & 3.0999 & 3.1653 \\
48 & 8/9 & 3.1420 & 3.2100 \\
\addlinespace[1.5pt]
66 & 2/10 & 3.1607 & 3.2162 \\
66 & 3/10 & 3.1083 & 3.1649 \\
66 & 4/10 & 3.0793 & 3.1374 \\
66 & 5/10 & 3.0672 & 3.1266 \\
66 & 6/10 & \textbf{3.0664} & 3.1270 \\
66 & 7/10 & 3.0702 & 3.1322 \\
66 & 8/10 & 3.0798 & 3.1442 \\
66 & 9/10 & 3.1249 & 3.1930 \\
\addlinespace[1.5pt]
85 & 2/13 & 3.1711 & 3.2241 \\
85 & 3/13 & 3.1202 & 3.1756 \\
85 & 4/13 & 3.0906 & 3.1462 \\
85 & 5/13 & 3.0702 & 3.1279 \\
85 & 6/13 & 3.0552 & 3.1155 \\
85 & 7/13 & 3.0463 & 3.1065 \\
85 & 8/13 & \textbf{3.0448} & 3.1057 \\
85 & 9/13 & 3.0495 & 3.1125 \\
85 & 10/13 & 3.0595 & 3.1235 \\
85 & 11/13 & 3.0826 & 3.1488 \\
85 & 12/13 & 3.1251 & 3.1932 \\
\addlinespace[1.5pt]
113 & 2/12 & 3.1474 & 3.2003 \\
113 & 3/12 & 3.0970 & 3.1530 \\
113 & 4/12 & 3.0689 & 3.1266 \\
113 & 5/12 & 3.0507 & 3.1095 \\
113 & 6/12 & 3.0410 & 3.1007 \\
113 & 7/12 & \textbf{3.0360} & 3.0984 \\
113 & 8/12 & 3.0362 & 3.0979 \\
113 & 9/12 & 3.0549 & 3.1200 \\
113 & 10/12 & 3.0934 & 3.1587 \\
113 & 11/12 & 3.1255 & 3.1943 \\
\addlinespace[1.5pt]
170 & 2/18 & 3.2066 & 3.2599 \\
170 & 3/18 & 3.1560 & 3.2116 \\
170 & 4/18 & 3.1256 & 3.1829 \\
170 & 5/18 & 3.1012 & 3.1596 \\
170 & 6/18 & 3.0895 & 3.1490 \\
170 & 7/18 & 3.0757 & 3.1357 \\
170 & 8/18 & 3.0676 & 3.1289 \\
170 & 9/18 & 3.0624 & 3.1246 \\
170 & 10/18 & 3.0621 & 3.1243 \\
170 & 11/18 & \textbf{3.0536} & 3.1162 \\
170 & 12/18 & 3.0548 & 3.1177 \\
170 & 13/18 & 3.0553 & 3.1181 \\
170 & 14/18 & 3.0590 & 3.1224 \\
170 & 15/18 & 3.0733 & 3.1380 \\
170 & 16/18 & 3.1153 & 3.1829 \\
170 & 17/18 & 3.1604 & 3.2310 \\
\addlinespace[1.5pt]
231 & 2/18 & 3.2208 & 3.2750 \\
231 & 3/18 & 3.1737 & 3.2307 \\
231 & 4/18 & 3.1396 & 3.1971 \\
231 & 5/18 & 3.1229 & 3.1814 \\
231 & 6/18 & 3.1165 & 3.1768 \\
231 & 7/18 & 3.0962 & 3.1571 \\
\bottomrule
\end{tabular}
\end{minipage}\hfill
\begin{minipage}[t]{0.245\linewidth}\centering
\begin{tabular}{rrrr}
\toprule
$N$ & $L_{\rm dec}/L$ & DUEL & ELBO \\
\midrule
\multicolumn{4}{l}{$C={6{\times}10^{18}}$ (cont.)} \\
231 & 8/18 & 3.0854 & 3.1464 \\
231 & 9/18 & 3.0870 & 3.1500 \\
231 & 10/18 & 3.0777 & 3.1400 \\
231 & 11/18 & 3.0780 & 3.1398 \\
231 & 12/18 & \textbf{3.0774} & 3.1416 \\
231 & 13/18 & 3.0898 & 3.1523 \\
231 & 14/18 & 3.0908 & 3.1542 \\
231 & 15/18 & 3.1165 & 3.1812 \\
231 & 16/18 & 3.1464 & 3.2133 \\
231 & 17/18 & 3.2015 & 3.2731 \\
\addlinespace[1.5pt]
336 & 2/20 & 3.2797 & 3.3339 \\
336 & 3/20 & 3.2703 & 3.3282 \\
336 & 4/20 & 3.2812 & 3.3417 \\
336 & 5/20 & 3.2417 & 3.3034 \\
336 & 6/20 & 3.2278 & 3.2902 \\
336 & 7/20 & 3.2010 & 3.2630 \\
336 & 8/20 & 3.1415 & 3.2038 \\
336 & 9/20 & 3.2137 & 3.2766 \\
336 & 10/20 & 3.1313 & 3.1947 \\
336 & 11/20 & 3.1955 & 3.2597 \\
336 & 12/20 & 3.1408 & 3.2052 \\
336 & 13/20 & 3.1485 & 3.2123 \\
336 & 14/20 & \textbf{3.1275} & 3.1904 \\
336 & 15/20 & 3.1393 & 3.2036 \\
336 & 16/20 & 3.1700 & 3.2349 \\
336 & 17/20 & 3.1777 & 3.2436 \\
336 & 18/20 & 3.1915 & 3.2589 \\
336 & 19/20 & 3.2588 & 3.3340 \\
\addlinespace[1.5pt]
472 & 2/18 & 3.2975 & 3.3505 \\
472 & 3/18 & 3.2595 & 3.3149 \\
472 & 4/18 & 3.2478 & 3.3063 \\
472 & 5/18 & 3.2865 & 3.3474 \\
472 & 6/18 & 3.2411 & 3.3019 \\
472 & 7/18 & 3.2493 & 3.3107 \\
472 & 8/18 & 3.2467 & 3.3073 \\
472 & 9/18 & 3.2641 & 3.3274 \\
472 & 10/18 & 3.2269 & 3.2898 \\
472 & 11/18 & \textbf{3.2243} & 3.2877 \\
472 & 12/18 & 3.2456 & 3.3094 \\
472 & 13/18 & 3.2295 & 3.2934 \\
472 & 14/18 & 3.2243 & 3.2883 \\
472 & 15/18 & 3.2244 & 3.2891 \\
472 & 16/18 & 3.2368 & 3.3033 \\
472 & 17/18 & 3.2727 & 3.3419 \\
\addlinespace[2pt]
\multicolumn{4}{l}{$C={10^{19}}$} \\
66 & 4/10 & 3.0106 & 3.0676 \\
66 & 5/10 & 2.9942 & 3.0531 \\
66 & 6/10 & 2.9908 & 3.0510 \\
66 & 7/10 & \textbf{2.9894} & 3.0517 \\
66 & 8/10 & 2.9954 & 3.0599 \\
66 & 9/10 & 3.0359 & 3.1031 \\
\addlinespace[1.5pt]
85 & 4/13 & 3.0153 & 3.0710 \\
85 & 5/13 & 2.9904 & 3.0473 \\
85 & 6/13 & 2.9753 & 3.0333 \\
85 & 7/13 & 2.9639 & 3.0233 \\
85 & 8/13 & \textbf{2.9567} & 3.0169 \\
85 & 9/13 & 2.9661 & 3.0271 \\
85 & 10/13 & 2.9705 & 3.0337 \\
85 & 11/13 & 2.9839 & 3.0494 \\
85 & 12/13 & 3.0265 & 3.0932 \\
\addlinespace[1.5pt]
113 & 4/12 & 2.9733 & 3.0294 \\
113 & 5/12 & 2.9566 & 3.0139 \\
113 & 6/12 & 2.9422 & 3.0005 \\
113 & 7/12 & 2.9395 & 2.9993 \\
113 & 8/12 & \textbf{2.9363} & 2.9970 \\
\bottomrule
\end{tabular}
\end{minipage}\hfill
\begin{minipage}[t]{0.245\linewidth}\centering
\begin{tabular}{rrrr}
\toprule
$N$ & $L_{\rm dec}/L$ & DUEL & ELBO \\
\midrule
\multicolumn{4}{l}{$C={10^{19}}$ (cont.)} \\
113 & 9/12 & 2.9466 & 3.0098 \\
113 & 10/12 & 2.9768 & 3.0424 \\
113 & 11/12 & 3.0116 & 3.0795 \\
\addlinespace[1.5pt]
142 & 3/15 & 3.0189 & 3.0731 \\
142 & 4/15 & 2.9882 & 3.0437 \\
142 & 5/15 & 2.9670 & 3.0239 \\
142 & 6/15 & 2.9550 & 3.0121 \\
142 & 7/15 & 2.9458 & 3.0045 \\
142 & 8/15 & 2.9367 & 2.9971 \\
142 & 9/15 & \textbf{2.9310} & 2.9918 \\
142 & 10/15 & 2.9333 & 2.9950 \\
142 & 11/15 & 2.9362 & 2.9989 \\
142 & 12/15 & 2.9478 & 3.0116 \\
142 & 13/15 & 2.9769 & 3.0414 \\
142 & 14/15 & 3.0164 & 3.0850 \\
\addlinespace[1.5pt]
170 & 4/18 & 3.0089 & 3.0648 \\
170 & 5/18 & 2.9824 & 3.0404 \\
170 & 6/18 & 2.9722 & 3.0308 \\
170 & 7/18 & 2.9595 & 3.0183 \\
170 & 8/18 & 2.9469 & 3.0065 \\
170 & 9/18 & 2.9431 & 3.0036 \\
170 & 10/18 & 2.9403 & 3.0018 \\
170 & 11/18 & 2.9358 & 2.9971 \\
170 & 12/18 & \textbf{2.9349} & 2.9963 \\
170 & 13/18 & 2.9362 & 2.9978 \\
170 & 14/18 & 2.9442 & 3.0074 \\
170 & 15/18 & 2.9531 & 3.0177 \\
170 & 16/18 & 2.9886 & 3.0553 \\
170 & 17/18 & 3.0301 & 3.0994 \\
\addlinespace[1.5pt]
231 & 4/18 & 3.0219 & 3.0788 \\
231 & 5/18 & 2.9889 & 3.0464 \\
231 & 6/18 & 2.9789 & 3.0384 \\
231 & 7/18 & 2.9686 & 3.0279 \\
231 & 8/18 & 2.9557 & 3.0164 \\
231 & 9/18 & 2.9529 & 3.0139 \\
231 & 10/18 & 2.9477 & 3.0086 \\
231 & 11/18 & \textbf{2.9426} & 3.0043 \\
231 & 12/18 & 2.9434 & 3.0063 \\
231 & 13/18 & 2.9521 & 3.0148 \\
231 & 14/18 & 2.9487 & 3.0114 \\
231 & 15/18 & 2.9779 & 3.0417 \\
231 & 16/18 & 3.0055 & 3.0704 \\
231 & 17/18 & 3.0416 & 3.1102 \\
\addlinespace[1.5pt]
336 & 4/20 & 3.0964 & 3.1552 \\
336 & 5/20 & 3.0726 & 3.1328 \\
336 & 6/20 & 3.0382 & 3.0987 \\
336 & 7/20 & 3.0004 & 3.0602 \\
336 & 8/20 & 2.9888 & 3.0497 \\
336 & 9/20 & 2.9890 & 3.0501 \\
336 & 10/20 & \textbf{2.9805} & 3.0415 \\
336 & 11/20 & 2.9809 & 3.0426 \\
336 & 12/20 & 2.9914 & 3.0548 \\
336 & 13/20 & 2.9982 & 3.0606 \\
336 & 14/20 & 2.9899 & 3.0530 \\
336 & 15/20 & 2.9965 & 3.0603 \\
336 & 16/20 & 3.0084 & 3.0718 \\
336 & 17/20 & 3.0198 & 3.0847 \\
336 & 18/20 & 3.0396 & 3.1058 \\
336 & 19/20 & 3.0807 & 3.1524 \\
\addlinespace[1.5pt]
472 & 4/18 & 3.0902 & 3.1475 \\
472 & 5/18 & 3.0874 & 3.1465 \\
472 & 6/18 & \textbf{3.0325} & 3.0915 \\
472 & 7/18 & 3.0649 & 3.1255 \\
472 & 8/18 & 3.0558 & 3.1163 \\
\bottomrule
\end{tabular}
\end{minipage}\hfill
\begin{minipage}[t]{0.245\linewidth}\centering
\begin{tabular}{rrrr}
\toprule
$N$ & $L_{\rm dec}/L$ & DUEL & ELBO \\
\midrule
\multicolumn{4}{l}{$C={10^{19}}$ (cont.)} \\
472 & 9/18 & 3.0480 & 3.1097 \\
472 & 10/18 & 3.0556 & 3.1181 \\
472 & 11/18 & 3.0455 & 3.1082 \\
472 & 12/18 & 3.0566 & 3.1194 \\
472 & 13/18 & 3.0434 & 3.1064 \\
472 & 14/18 & 3.0454 & 3.1091 \\
472 & 15/18 & 3.0544 & 3.1185 \\
472 & 16/18 & 3.0602 & 3.1252 \\
472 & 17/18 & 3.0848 & 3.1521 \\
\addlinespace[2pt]
\multicolumn{4}{l}{$C={3{\times}10^{19}}$} \\
142 & 4/15 & 2.8474 & 2.9014 \\
142 & 5/15 & 2.8215 & 2.8772 \\
142 & 6/15 & 2.8036 & 2.8593 \\
142 & 7/15 & 2.7955 & 2.8525 \\
142 & 8/15 & 2.7849 & 2.8422 \\
142 & 9/15 & 2.7756 & 2.8333 \\
142 & 10/15 & 2.7756 & 2.8343 \\
142 & 11/15 & \textbf{2.7737} & 2.8335 \\
142 & 12/15 & 2.7809 & 2.8425 \\
142 & 13/15 & 2.7978 & 2.8602 \\
142 & 14/15 & 2.8355 & 2.9001 \\
\addlinespace[1.5pt]
206 & 4/16 & 2.8121 & 2.8658 \\
206 & 5/16 & 2.7933 & 2.8480 \\
206 & 6/16 & 2.7719 & 2.8277 \\
206 & 8/16 & 2.7535 & 2.8107 \\
206 & 9/16 & 2.7487 & 2.8065 \\
206 & 10/16 & 2.7475 & 2.8059 \\
206 & 11/16 & 2.7463 & 2.8057 \\
206 & 12/16 & \textbf{2.7454} & 2.8046 \\
206 & 13/16 & 2.7563 & 2.8177 \\
206 & 14/16 & 2.7814 & 2.8446 \\
206 & 15/16 & 2.8161 & 2.8806 \\
\addlinespace[1.5pt]
268 & 3/16 & 2.8260 & 2.8783 \\
268 & 4/16 & 2.7929 & 2.8464 \\
268 & 5/16 & 2.7751 & 2.8296 \\
268 & 6/16 & 2.7559 & 2.8117 \\
268 & 7/16 & 2.7492 & 2.8060 \\
268 & 8/16 & 2.7430 & 2.8008 \\
268 & 9/16 & 2.7340 & 2.7923 \\
268 & 10/16 & 2.7338 & 2.7927 \\
268 & 11/16 & \textbf{2.7317} & 2.7905 \\
268 & 12/16 & 2.7392 & 2.7993 \\
268 & 13/16 & 2.7479 & 2.8095 \\
268 & 14/16 & 2.7724 & 2.8356 \\
268 & 15/16 & 2.8097 & 2.8743 \\
\addlinespace[1.5pt]
336 & 4/20 & 2.8114 & 2.8660 \\
336 & 5/20 & 2.7910 & 2.8457 \\
336 & 6/20 & 2.7713 & 2.8258 \\
336 & 7/20 & 2.7625 & 2.8192 \\
336 & 8/20 & 2.7518 & 2.8086 \\
336 & 9/20 & 2.7465 & 2.8044 \\
336 & 10/20 & 2.7380 & 2.7957 \\
336 & 11/20 & 2.7376 & 2.7961 \\
336 & 12/20 & 2.7352 & 2.7946 \\
336 & 13/20 & 2.7353 & 2.7950 \\
336 & 14/20 & \textbf{2.7338} & 2.7938 \\
336 & 15/20 & 2.7378 & 2.7979 \\
336 & 16/20 & 2.7419 & 2.8029 \\
336 & 17/20 & 2.7562 & 2.8188 \\
336 & 18/20 & 2.7837 & 2.8482 \\
336 & 19/20 & 2.8215 & 2.8893 \\
\addlinespace[1.5pt]
472 & 4/18 & 2.8030 & 2.8573 \\
472 & 5/18 & 2.7847 & 2.8403 \\
472 & 6/18 & 2.7723 & 2.8282 \\
472 & 7/18 & 2.7641 & 2.8215 \\
\bottomrule
\end{tabular}
\end{minipage}
\end{table}

\begin{table}[p]
\centering
\scriptsize\setlength{\tabcolsep}{2.4pt}
\caption{Decoder-depth sweep of E2D2 $b{=}4$ (part 2 of 3): every trained model, one row per decoder depth.}
\label{tab:cells6_2}
\begin{minipage}[t]{0.245\linewidth}\centering
\begin{tabular}{rrrr}
\toprule
$N$ & $L_{\rm dec}/L$ & DUEL & ELBO \\
\midrule
\multicolumn{4}{l}{$C={3{\times}10^{19}}$ (cont.)} \\
472 & 8/18 & 2.7557 & 2.8136 \\
472 & 9/18 & 2.7522 & 2.8107 \\
472 & 10/18 & 2.7473 & 2.8059 \\
472 & 11/18 & \textbf{2.7401} & 2.7998 \\
472 & 12/18 & 2.7457 & 2.8048 \\
472 & 13/18 & 2.7531 & 2.8128 \\
472 & 14/18 & 2.7558 & 2.8160 \\
472 & 15/18 & 2.7699 & 2.8313 \\
472 & 16/18 & 2.7879 & 2.8507 \\
472 & 17/18 & 2.8122 & 2.8768 \\
\addlinespace[1.5pt]
666 & 4/21 & 2.8589 & 2.9140 \\
666 & 5/21 & 2.8463 & 2.9021 \\
666 & 6/21 & 2.8193 & 2.8759 \\
666 & 7/21 & 2.8202 & 2.8785 \\
666 & 8/21 & 2.8150 & 2.8733 \\
666 & 9/21 & 2.7979 & 2.8566 \\
666 & 10/21 & 2.7952 & 2.8535 \\
666 & 11/21 & 2.7941 & 2.8533 \\
666 & 12/21 & 2.7961 & 2.8562 \\
666 & 13/21 & \textbf{2.7839} & 2.8438 \\
666 & 14/21 & 2.7901 & 2.8500 \\
666 & 15/21 & 2.7911 & 2.8510 \\
666 & 16/21 & 2.7971 & 2.8580 \\
666 & 17/21 & 2.7984 & 2.8591 \\
666 & 18/21 & 2.8061 & 2.8683 \\
666 & 19/21 & 2.8174 & 2.8795 \\
666 & 20/21 & 2.8400 & 2.9027 \\
\addlinespace[1.5pt]
944 & 4/25 & 2.9194 & 2.9758 \\
944 & 5/25 & 2.9191 & 2.9768 \\
944 & 6/25 & 2.9078 & 2.9666 \\
944 & 7/25 & 2.8925 & 2.9508 \\
944 & 8/25 & 2.8742 & 2.9344 \\
944 & 9/25 & 2.8795 & 2.9398 \\
944 & 10/25 & 2.8697 & 2.9289 \\
944 & 11/25 & 2.9058 & 2.9659 \\
944 & 12/25 & 2.8689 & 2.9293 \\
944 & 13/25 & 2.8468 & 2.9071 \\
944 & 14/25 & 2.8795 & 2.9402 \\
944 & 15/25 & 2.8439 & 2.9051 \\
944 & 16/25 & 2.8612 & 2.9219 \\
944 & 17/25 & 2.8485 & 2.9108 \\
944 & 18/25 & \textbf{2.8409} & 2.9036 \\
944 & 19/25 & 2.8564 & 2.9181 \\
944 & 20/25 & 2.8542 & 2.9160 \\
944 & 21/25 & 2.8547 & 2.9168 \\
944 & 22/25 & 2.8533 & 2.9162 \\
944 & 23/25 & 2.8615 & 2.9240 \\
944 & 24/25 & 2.8812 & 2.9459 \\
\addlinespace[2pt]
\multicolumn{4}{l}{$C={6{\times}10^{19}}$} \\
206 & 4/16 & 2.7523 & 2.8055 \\
206 & 5/16 & 2.7298 & 2.7842 \\
206 & 6/16 & 2.7063 & 2.7607 \\
206 & 7/16 & 2.6884 & 2.7432 \\
206 & 8/16 & 2.6833 & 2.7385 \\
206 & 9/16 & 2.6707 & 2.7269 \\
206 & 10/16 & 2.6645 & 2.7214 \\
206 & 11/16 & 2.6620 & 2.7193 \\
206 & 12/16 & \textbf{2.6618} & 2.7200 \\
206 & 13/16 & 2.6696 & 2.7285 \\
206 & 14/16 & 2.6917 & 2.7526 \\
\addlinespace[1.5pt]
268 & 4/16 & 2.7199 & 2.7725 \\
268 & 5/16 & 2.6993 & 2.7534 \\
268 & 6/16 & 2.6735 & 2.7265 \\
268 & 7/16 & 2.6604 & 2.7144 \\
268 & 8/16 & 2.6531 & 2.7081 \\
\bottomrule
\end{tabular}
\end{minipage}\hfill
\begin{minipage}[t]{0.245\linewidth}\centering
\begin{tabular}{rrrr}
\toprule
$N$ & $L_{\rm dec}/L$ & DUEL & ELBO \\
\midrule
\multicolumn{4}{l}{$C={6{\times}10^{19}}$ (cont.)} \\
268 & 9/16 & 2.6497 & 2.7047 \\
268 & 10/16 & 2.6419 & 2.6981 \\
268 & 11/16 & \textbf{2.6379} & 2.6944 \\
268 & 12/16 & 2.6453 & 2.7028 \\
268 & 13/16 & 2.6531 & 2.7117 \\
268 & 14/16 & 2.6728 & 2.7333 \\
\addlinespace[1.5pt]
336 & 4/20 & 2.7227 & 2.7742 \\
336 & 5/20 & 2.7075 & 2.7609 \\
336 & 6/20 & 2.6809 & 2.7342 \\
336 & 7/20 & 2.6661 & 2.7203 \\
336 & 8/20 & 2.6553 & 2.7095 \\
336 & 9/20 & 2.6556 & 2.7107 \\
336 & 10/20 & 2.6363 & 2.6920 \\
336 & 11/20 & 2.6368 & 2.6930 \\
336 & 12/20 & 2.6355 & 2.6918 \\
336 & 13/20 & 2.6365 & 2.6933 \\
336 & 14/20 & \textbf{2.6336} & 2.6904 \\
336 & 15/20 & 2.6367 & 2.6945 \\
336 & 16/20 & 2.6379 & 2.6964 \\
336 & 17/20 & 2.6494 & 2.7091 \\
336 & 18/20 & 2.6719 & 2.7326 \\
336 & 19/20 & 2.7075 & 2.7724 \\
\addlinespace[1.5pt]
472 & 3/18 & 2.7143 & 2.7648 \\
472 & 4/18 & 2.6872 & 2.7388 \\
472 & 5/18 & 2.6659 & 2.7180 \\
472 & 6/18 & 2.6506 & 2.7037 \\
472 & 7/18 & 2.6413 & 2.6953 \\
472 & 8/18 & 2.6340 & 2.6891 \\
472 & 9/18 & 2.6270 & 2.6821 \\
472 & 10/18 & 2.6227 & 2.6780 \\
472 & 11/18 & \textbf{2.6206} & 2.6773 \\
472 & 12/18 & 2.6223 & 2.6796 \\
472 & 13/18 & 2.6214 & 2.6791 \\
472 & 14/18 & 2.6276 & 2.6855 \\
472 & 15/18 & 2.6405 & 2.6996 \\
472 & 16/18 & 2.6639 & 2.7245 \\
472 & 17/18 & 2.6940 & 2.7561 \\
\addlinespace[1.5pt]
666 & 4/21 & 2.7031 & 2.7556 \\
666 & 5/21 & 2.6807 & 2.7336 \\
666 & 6/21 & 2.6704 & 2.7245 \\
666 & 7/21 & 2.6612 & 2.7163 \\
666 & 8/21 & 2.6494 & 2.7043 \\
666 & 9/21 & 2.6435 & 2.6997 \\
666 & 10/21 & 2.6377 & 2.6934 \\
666 & 11/21 & 2.6380 & 2.6945 \\
666 & 12/21 & 2.6407 & 2.6984 \\
666 & 13/21 & 2.6332 & 2.6901 \\
666 & 14/21 & 2.6346 & 2.6925 \\
666 & 15/21 & \textbf{2.6313} & 2.6887 \\
666 & 16/21 & 2.6321 & 2.6904 \\
666 & 17/21 & 2.6384 & 2.6965 \\
666 & 18/21 & 2.6578 & 2.7169 \\
666 & 19/21 & 2.6749 & 2.7363 \\
666 & 20/21 & 2.7062 & 2.7701 \\
\addlinespace[1.5pt]
944 & 4/25 & 2.7380 & 2.7908 \\
944 & 5/25 & 2.7249 & 2.7800 \\
944 & 6/25 & 2.7171 & 2.7731 \\
944 & 7/25 & 2.6976 & 2.7531 \\
944 & 8/25 & 2.6959 & 2.7531 \\
944 & 9/25 & 2.6984 & 2.7552 \\
944 & 10/25 & 2.6978 & 2.7553 \\
944 & 11/25 & 2.6845 & 2.7424 \\
944 & 12/25 & 2.6793 & 2.7376 \\
944 & 13/25 & 2.6772 & 2.7355 \\
\bottomrule
\end{tabular}
\end{minipage}\hfill
\begin{minipage}[t]{0.245\linewidth}\centering
\begin{tabular}{rrrr}
\toprule
$N$ & $L_{\rm dec}/L$ & DUEL & ELBO \\
\midrule
\multicolumn{4}{l}{$C={6{\times}10^{19}}$ (cont.)} \\
944 & 14/25 & 2.6712 & 2.7289 \\
944 & 15/25 & 2.6678 & 2.7253 \\
944 & 16/25 & 2.6762 & 2.7350 \\
944 & 17/25 & 2.6729 & 2.7318 \\
944 & 18/25 & 2.6690 & 2.7277 \\
944 & 19/25 & \textbf{2.6664} & 2.7255 \\
944 & 20/25 & 2.6770 & 2.7357 \\
944 & 21/25 & 2.6759 & 2.7341 \\
944 & 22/25 & 2.6891 & 2.7493 \\
944 & 23/25 & 2.7030 & 2.7643 \\
944 & 24/25 & 2.7255 & 2.7886 \\
\addlinespace[1.5pt]
1233 & 4/24 & 2.7630 & 2.8180 \\
1233 & 5/24 & 2.7853 & 2.8420 \\
1233 & 6/24 & 2.7545 & 2.8118 \\
1233 & 7/24 & 2.7437 & 2.8015 \\
1233 & 8/24 & 2.7343 & 2.7917 \\
1233 & 9/24 & 2.7298 & 2.7875 \\
1233 & 10/24 & 2.7349 & 2.7935 \\
1233 & 11/24 & 2.7293 & 2.7880 \\
1233 & 12/24 & 2.7199 & 2.7780 \\
1233 & 13/24 & 2.7287 & 2.7877 \\
1233 & 14/24 & 2.7186 & 2.7777 \\
1233 & 15/24 & \textbf{2.7141} & 2.7731 \\
1233 & 16/24 & 2.7417 & 2.8025 \\
1233 & 17/24 & 2.7216 & 2.7817 \\
1233 & 18/24 & 2.7215 & 2.7806 \\
1233 & 19/24 & 2.7165 & 2.7751 \\
1233 & 20/24 & 2.7208 & 2.7804 \\
1233 & 21/24 & 2.7187 & 2.7782 \\
1233 & 22/24 & 2.7286 & 2.7885 \\
1233 & 23/24 & 2.7476 & 2.8087 \\
\addlinespace[2pt]
\multicolumn{4}{l}{$C={10^{20}}$} \\
268 & 5/16 & 2.6477 & 2.7001 \\
268 & 6/16 & 2.6346 & 2.6875 \\
268 & 7/16 & 2.6178 & 2.6726 \\
268 & 8/16 & 2.6048 & 2.6591 \\
268 & 9/16 & 2.5947 & 2.6495 \\
268 & 10/16 & 2.5910 & 2.6467 \\
268 & 11/16 & \textbf{2.5853} & 2.6405 \\
268 & 12/16 & 2.5890 & 2.6454 \\
268 & 13/16 & 2.5966 & 2.6543 \\
268 & 14/16 & 2.6139 & 2.6734 \\
\addlinespace[1.5pt]
336 & 4/20 & 2.6765 & 2.7279 \\
336 & 5/20 & 2.6583 & 2.7109 \\
336 & 6/20 & 2.6362 & 2.6898 \\
336 & 7/20 & 2.6168 & 2.6706 \\
336 & 8/20 & 2.6108 & 2.6651 \\
336 & 9/20 & 2.6024 & 2.6564 \\
336 & 10/20 & 2.5878 & 2.6427 \\
336 & 11/20 & 2.5856 & 2.6406 \\
336 & 12/20 & 2.5774 & 2.6323 \\
336 & 13/20 & 2.5744 & 2.6300 \\
336 & 14/20 & 2.5728 & 2.6287 \\
336 & 15/20 & \textbf{2.5724} & 2.6292 \\
336 & 16/20 & 2.5765 & 2.6338 \\
336 & 17/20 & 2.5849 & 2.6432 \\
336 & 18/20 & 2.6046 & 2.6640 \\
336 & 19/20 & 2.6394 & 2.7017 \\
\addlinespace[1.5pt]
472 & 4/18 & 2.6306 & 2.6810 \\
472 & 5/18 & 2.5997 & 2.6506 \\
472 & 6/18 & 2.5820 & 2.6330 \\
472 & 7/18 & 2.5704 & 2.6227 \\
472 & 8/18 & 2.5614 & 2.6141 \\
472 & 9/18 & 2.5581 & 2.6111 \\
472 & 10/18 & 2.5538 & 2.6079 \\
\bottomrule
\end{tabular}
\end{minipage}\hfill
\begin{minipage}[t]{0.245\linewidth}\centering
\begin{tabular}{rrrr}
\toprule
$N$ & $L_{\rm dec}/L$ & DUEL & ELBO \\
\midrule
\multicolumn{4}{l}{$C={10^{20}}$ (cont.)} \\
472 & 11/18 & \textbf{2.5524} & 2.6063 \\
472 & 12/18 & 2.5555 & 2.6107 \\
472 & 13/18 & 2.5541 & 2.6096 \\
472 & 14/18 & 2.5601 & 2.6164 \\
472 & 15/18 & 2.5693 & 2.6266 \\
472 & 16/18 & 2.5909 & 2.6503 \\
472 & 17/18 & 2.6229 & 2.6840 \\
\addlinespace[1.5pt]
666 & 3/21 & 2.6577 & 2.7067 \\
666 & 4/21 & 2.6229 & 2.6728 \\
666 & 5/21 & 2.6037 & 2.6549 \\
666 & 6/21 & 2.5892 & 2.6415 \\
666 & 7/21 & 2.5738 & 2.6265 \\
666 & 8/21 & 2.5668 & 2.6199 \\
666 & 9/21 & 2.5576 & 2.6114 \\
666 & 10/21 & 2.5535 & 2.6074 \\
666 & 11/21 & 2.5515 & 2.6056 \\
666 & 12/21 & 2.5506 & 2.6055 \\
666 & 13/21 & 2.5488 & 2.6039 \\
666 & 14/21 & \textbf{2.5484} & 2.6042 \\
666 & 15/21 & 2.5492 & 2.6049 \\
666 & 16/21 & 2.5535 & 2.6098 \\
666 & 17/21 & 2.5562 & 2.6130 \\
666 & 18/21 & 2.5697 & 2.6275 \\
666 & 19/21 & 2.5902 & 2.6495 \\
666 & 20/21 & 2.6251 & 2.6870 \\
\addlinespace[1.5pt]
944 & 4/25 & 2.6421 & 2.6934 \\
944 & 5/25 & 2.6204 & 2.6729 \\
944 & 6/25 & 2.6076 & 2.6604 \\
944 & 7/25 & 2.5954 & 2.6487 \\
944 & 8/25 & 2.5867 & 2.6406 \\
944 & 9/25 & 2.5767 & 2.6317 \\
944 & 10/25 & 2.5714 & 2.6258 \\
944 & 11/25 & 2.5679 & 2.6234 \\
944 & 12/25 & 2.5690 & 2.6247 \\
944 & 13/25 & 2.5633 & 2.6185 \\
944 & 14/25 & 2.5622 & 2.6181 \\
944 & 15/25 & 2.5610 & 2.6171 \\
944 & 16/25 & 2.5630 & 2.6200 \\
944 & 17/25 & \textbf{2.5609} & 2.6169 \\
944 & 18/25 & 2.5644 & 2.6204 \\
944 & 19/25 & 2.5634 & 2.6199 \\
944 & 20/25 & 2.5651 & 2.6215 \\
944 & 21/25 & 2.5707 & 2.6278 \\
944 & 22/25 & 2.5842 & 2.6421 \\
944 & 23/25 & 2.6057 & 2.6642 \\
944 & 24/25 & 2.6330 & 2.6955 \\
\addlinespace[1.5pt]
1233 & 4/24 & 2.6590 & 2.7112 \\
1233 & 5/24 & 2.6434 & 2.6964 \\
1233 & 6/24 & 2.6366 & 2.6912 \\
1233 & 7/24 & 2.6208 & 2.6751 \\
1233 & 8/24 & 2.6089 & 2.6640 \\
1233 & 9/24 & 2.6036 & 2.6589 \\
1233 & 10/24 & 2.6059 & 2.6619 \\
1233 & 11/24 & 2.5904 & 2.6469 \\
1233 & 12/24 & 2.5864 & 2.6419 \\
1233 & 13/24 & \textbf{2.5858} & 2.6416 \\
1233 & 14/24 & 2.5971 & 2.6543 \\
1233 & 15/24 & 2.5936 & 2.6502 \\
1233 & 16/24 & 2.5918 & 2.6488 \\
1233 & 17/24 & 2.5972 & 2.6541 \\
1233 & 18/24 & 2.5922 & 2.6492 \\
1233 & 19/24 & 2.5931 & 2.6502 \\
1233 & 20/24 & 2.5928 & 2.6498 \\
1233 & 21/24 & 2.6052 & 2.6632 \\
\bottomrule
\end{tabular}
\end{minipage}
\end{table}

\begin{table}[p]
\centering
\scriptsize\setlength{\tabcolsep}{2.4pt}
\caption{Decoder-depth sweep of E2D2 $b{=}4$ (part 3 of 3): every trained model, one row per decoder depth.}
\label{tab:cells6_3}
\begin{minipage}[t]{0.245\linewidth}\centering
\begin{tabular}{rrrr}
\toprule
$N$ & $L_{\rm dec}/L$ & DUEL & ELBO \\
\midrule
\multicolumn{4}{l}{$C={10^{20}}$ (cont.)} \\
1233 & 22/24 & 2.6232 & 2.6821 \\
1233 & 23/24 & 2.6418 & 2.7035 \\
\addlinespace[1.5pt]
1678 & 4/25 & 2.6946 & 2.7491 \\
1678 & 5/25 & 2.6936 & 2.7485 \\
1678 & 6/25 & 2.6715 & 2.7271 \\
1678 & 7/25 & 2.6638 & 2.7204 \\
1678 & 8/25 & 2.6658 & 2.7217 \\
1678 & 9/25 & 2.6566 & 2.7136 \\
1678 & 10/25 & 2.6458 & 2.7022 \\
1678 & 11/25 & 2.6635 & 2.7211 \\
1678 & 12/25 & 2.6448 & 2.7016 \\
1678 & 13/25 & \textbf{2.6374} & 2.6945 \\
1678 & 14/25 & 2.6472 & 2.7049 \\
1678 & 15/25 & 2.6409 & 2.6976 \\
1678 & 16/25 & 2.6394 & 2.6973 \\
1678 & 17/25 & 2.6477 & 2.7055 \\
1678 & 18/25 & 2.6475 & 2.7057 \\
1678 & 19/25 & 2.6486 & 2.7068 \\
1678 & 20/25 & 2.6484 & 2.7068 \\
1678 & 21/25 & 2.6397 & 2.6976 \\
1678 & 22/25 & 2.6442 & 2.7034 \\
1678 & 23/25 & 2.6571 & 2.7165 \\
1678 & 24/25 & 2.6741 & 2.7344 \\
\bottomrule
\end{tabular}
\end{minipage}
\end{table}

\end{document}